\documentclass[review]{elsarticle}

\usepackage[a4paper,portrait, margin=1in]{geometry}
\usepackage{calc}
\usepackage{amssymb}
\usepackage{amstext}
\usepackage{amsmath}
\usepackage{amsthm}
\usepackage{caption}
\usepackage{subcaption}
\usepackage{xfrac}
\usepackage{array}
\usepackage{multicol}
\usepackage{hhline}
\usepackage{gensymb}
\usepackage{times}
\usepackage{enumerate}
\usepackage{pslatex}
\usepackage{multirow}
\usepackage{rotating}
\usepackage{setspace}	
\usepackage{subfloat}
\usepackage{float}
\usepackage{makecell}
\usepackage{epstopdf}

\usepackage[linesnumbered,ruled,vlined,noend]{algorithm2e}
\usepackage{algorithmic,algorithm2e,float}

\usepackage{hyperref}
\newcommand*\rfrac[2]{{}^{#1}\!/_{#2}}
  
\graphicspath{{./Figs/}}

\newcommand{\algor}[1]{%
\begin{minipage}[t]{0.52\linewidth}
  \vspace{0pt}  
\begin{algorithm}[H] 
\SetKw{KwGoTo}{go to}
\algsetup{linenosize=\small}
\small
\setstretch{1}
\DontPrintSemicolon
\KwIn{Training set: $\mathbf{S}=\{(\mathbf{x}_i,y_i);i=1,...,M\}$,$y_i \in \{-1,1\}$ \newline
$\#$~of iterations: $E$ \newline
Test input : $\mathbf{X}$}
\KwOut{Prediction Function: $H(\cdot)$\\ }
Initialize $\textbf{W}_1(i)=\frac{1}{M}$ for $i=1,...,M$.

\For {$e=1,..,E$}{
\begin{enumerate} [i]
\item  Create new training set $\textbf{S}_{e}^{'}$ with weight distribution $\textbf{W}_{e}^{'}$. {\label{err} }
 \item   Train classifier $C_e$ on $\textbf{S}_{e}^{'}$ with $\textbf{W}_{e}^{'}$.  
 \item   Test $C_e$ on $\textbf{S}$ and get back a label set $\{Y_i, i=1,...,M\}$.  
 \item   Calculate the pseudo-loss for $\textbf{S}$ and $\textbf{W}_{e}$:
 
  $\epsilon_e=\underset{(i,Y_i):y_i \neq Y_i}{\sum}W_e(i) \ $.  
  \item If $\epsilon_e>0.5$ \KwGoTo step \ref{err}
 \item   Calculate the weight update parameter: 
  $\alpha_e=\frac{\epsilon_e}{1-\epsilon_e}$  
 \item   Update $\textbf{W}_{\rm e+1}(i)=\textbf{W}_{\rm e}(i)\alpha_e^{\rm \lvert y_i-Y_i \rvert/2} \ $  
 \item   Normalize  $\textbf{W}_{\rm e+1}$ such that: $\sum \textbf{W}_{\rm e+1}=1$.
 \end{enumerate}
  }  
 Output the final hypothesis:  $H(\cdot)= \sum_{e=1}^{E}h_e(\cdot)\log \frac{1}{\alpha_e}$	
 
\caption{AdaBoost.M1 Algorithm}
\label{Boosting}		
\end{algorithm}
\end{minipage}%
}
\newcommand{\algo}[1]{%
\begin{minipage}[t]{0.375\linewidth}
  \vspace{0pt}  
\begin{algorithm}[H]
\SetKw{KwGoTo}{go to}
\algsetup{linenosize=\small}
\small
\setstretch{1}
\DontPrintSemicolon
\KwIn{Training set: $\mathbf{S}=\{(\mathbf{x}_i,y_i);i=1,...,M\}$,$y_i \in \{-1,1\}$ \newline
$\#$~of iterations: $E$ \newline
Bootstrap size: $N_B$\newline
Test input : $\mathbf{X}$}
\KwOut{Prediction Function: $H(\cdot)$\\ }
\For {$e=1,..,E$}{
\begin{enumerate} [i]
\item  Select $ N_B $ samples to create training 

subset $\textbf{S}_{e}^{'}$ using RUS. 
 \item   Train classifier $C_e$ on $\textbf{S}_{e}^{'}$.
 \end{enumerate}
  }  
 Output the final hypothesis:  $H(\cdot)= \sum_{e=1}^{E}h_e(\cdot)$	
 
\caption{Bagging}
\label{Bagging}
\end{algorithm}
\end{minipage}%
}

\journal{Journal of Information Fusion}

\begin{document}

\begin{frontmatter}

\title{Progressive Boosting for Class Imbalance}

\author[rvt]{Roghayeh Soleymani\corref{cor1}}
\ead{rSoleymani@livia.etsmtl.ca}
\author[rvt]{Eric Granger}
\ead{Eric.Granger@etsmtl.ca}
\author[els]{Giorgio Fumera}
\ead{Fumera@diee.unica.it}

\cortext[cor1]{Corresponding author}
\address[rvt]{Laboratoire d'imagerie, de vision et d'intelligence artificielle, \'Ecole de technologie sup\'erieure\\ Universit\'e du Qu\'ebec, Montreal, Canada}
\address[els]{Pattern Recognition and Applications Group, Dept. of Electrical and Electronic Engineering\\ University of Cagliari, Cagliari, Italy}

\begin{abstract}

In practice, pattern recognition applications often suffer from imbalanced data distributions between classes, which may vary during operations w.r.t. the design data.
Two-class classification systems designed using imbalanced data tend to recognize the majority (negative) class better, while the class of interest (positive class) often has the smaller number of samples.
Several data-level techniques have been proposed to alleviate this issue, where classifier ensembles are designed with balanced data subsets by up-sampling positive samples or under-sampling negative samples. However, some informative samples may be neglected by random under-sampling and adding synthetic positive samples through up-sampling adds to training complexity.
In this paper, a new ensemble learning algorithm called Progressive Boosting (PBoost) is proposed that progressively inserts uncorrelated groups of samples into a Boosting procedure to avoid loosing information while generating a diverse pool of classifiers.
Base classifiers in this ensemble are generated from one iteration to the next, using subsets from a validation set that grows gradually in size and imbalance. Consequently, PBoost is more robust when the operational data may have unknown and variable levels of skew. In addition, the computation complexity of PBoost is lower than Boosting ensembles in literature that use under-sampling for learning from imbalanced data because not all of the base classifiers are validated on all negative samples.
In PBoost algorithm, a new loss factor is proposed to avoid bias of performance towards the negative class. Using this loss factor, the weight update of samples and classifier contribution in final predictions are set based on the ability to recognize both classes. Using the proposed loss factor instead of standard accuracy can avoid biasing performance in any Boosting ensemble.
The proposed approach was validated and compared using synthetic data, videos from the Faces In Action dataset that emulates face re-identification applications, and KEEL collection of datasets. Results show that PBoost can outperform state of the art techniques in terms of both accuracy and complexity over different levels of imbalance and overlap between classes. 
\end{abstract}


\end{frontmatter}

%\linenumbers

\section{Introduction}

Class imbalance is a fundamental issue in many real-world pattern recognition applications found in, e.g., automated video surveillance, fraud detection, intrusion detection in computer and network security, risk management, and medical diagnosis. Imbalance appears in binary classification problems and binarization of multi-class classification problems using one-vs-all strategy when samples from one class are compared against all samples from all other classes~\cite{galar2011overview,wang2012multiclass}.
In practice, the level of imbalance observed during operations in unknown a priori and varies over time. This level of skew may differ from what is seen in the design data. Classification algorithms designed using imbalanced data are often biased towards the majority (negative) class, even though the minority class is the (positive) class of interest. The main reason is that learning algorithms are typically designed to optimize the performance in terms of standard accuracy. Consequently, correct classification of negative class becomes their priority due to the abundance of samples for this class. 

Several approaches have been proposed in literature to design ensembles of classifiers using imbalanced data \cite{he2009learning,galar2012review}. In this paper, these approaches are divided into data-level and algorithm-level approaches. Data-level approaches either up-sample the positive class, under-sample the negative class or combine up-sampling and under-sampling to re-balance data for learning an ensemble of classifiers. Algorithm-level methods create or modify learning algorithms to counter the bias towards the negative class through cost-free techniques or by introducing uneven misclassification costs for the samples from different classes in cost-sensitive approaches.

Ensembles can be designed according to a static or dynamic approach.  
Static ensembles generate a diverse set of base classifiers a priori, often by re-balancing the training data. 
The ensembles selection or fusion may be set off-line using validation data, but typically assume a fixed level of imbalance during operations.
Dynamic ensembles allow to adapt the selection and fusion of base classifiers during operations based on the estimated level of skew~\cite{radtke2014skew, de2015adaptive}. 
For example, in~\cite{radtke2014skew, de2015adaptive} authors design base classifiers for a range of different levels of imbalance. Then, they estimate skew level of input data stream and select a suitable fusion function based on that level.

However, the level of imbalance may be difficult to estimate accurately during operations and the diverging selection and fusion function can decrease performance.
In contrast, using a static approach, the range of possible imbalance levels can be accounted for during design by training base classifiers on data subsets with different imbalance levels \cite{icpram2016TUS}.

Most of the ensemble learning methods to handle imbalance in literature are static approaches. Boosting~\cite{freund1995desicion,freund1996experiments} is a common static ensemble method that has been modified  in several ways to learn from imbalanced data (see a review by Galar et al.~\cite{galar2012review}). In data-level Boosting approaches, training data is rebalanced by up-sampling positive class, under-sampling negative class, or using both up-sampling and under-sampling\cite{chawla2003smoteboost,hu2009msmote,mease2007cost,guo2004learning,seiffert2010rusboost,galar2013eusboost,diez2015random}.  Up-sampling methods like SMOTEBoost~\cite{chawla2003smoteboost} are often more accurate, but they are computationally complex. In contrast, random under-sampling (RUS)~\cite{seiffert2010rusboost} is more computationally efficient, but suffers from information loss. 
 
Boosting ensembles may suffer from the bias of performance towards negative class because the loss factor, which guides their learning process, is obtained based on weighted accuracy. In cases of imbalance, weighted accuracy reflects the ability for correct classification of negative samples more than positive ones. This issue can be avoided by adopting a cost-sensitive approach~\cite{fan1999adacost,ting2000comparative,sun2007cost}, that defines different misclassification costs for different classes and integrates these cost factors into Boosting learning process. The drawback of these cost-sensitive techniques is that they rely on the suitable selection of cost factors which is often estimated by searching a range of possible values. In contrast, cost-free techniques modify learning algorithms by enhancing loss factor calculation without considering cost factors~\cite{joshi2001evaluating,kim2015geometric}. 

In this paper, the Progressive Boosting (PBoost) algorithm is proposed to design static classifier ensembles that can maintain a high level of performance over a range of possible levels of imbalance and complexity in the data encountered during operations. In this algorithm, samples from the negative class are regrouped into disjoint partitions, and over iterations, these partitions are gradually accumulated into a temporary design subset. During each Boosting iteration, a new base classifier is trained using negative samples selected randomly from this subset. However, samples from the newly added partition and the important samples from previous iterations have an equally higher probability of being selected. The base classifier is then validated on the whole temporary subset. 
As with traditional Boosting ensembles, the samples that are misclassified are considered as the most important samples and their weights increase.
With the sample selection scheme proposed in this paper, loss of information is considerably reduced, correlation among subsets of negative class is low, and only important samples tend to appear in more than one training subset. Therefore, the diversity and accuracy of Boosting ensembles tend to increase. In addition, to avoid biasing the performance towards the negative class, the proposed Boosting algorithm employs a new loss factor based on the F$_\beta$-measure that is applicable in any Boosting ensemble. 

 The diverse pool of classifiers generated with PBoost allows to globally model a range of different levels of imbalance and decision bound complexities for the data. Therefore, the static ensembles produced using PBoost are robust to possible variations in data processed during operations because base classifiers are validated on a growing number of negative samples (imbalance level). In addition, the number of samples used per iteration to design (train and validate) a classifier in this ensemble is smaller than Boosting methods in the literature, which translates to a lower computational complexity for design.
 
 The contributions of the proposed PBoost algorithm is summarized as follows:
 \begin{itemize}
 \item A sample selection process for design where negative class samples are regrouped into disjoint partitions for training diverse base classifiers to avoid loss of information and bias of performance;
 \item A procedure to validate base classifiers on growing number of negative samples to increase robustness to imbalance and decrease computation complexity;
 \item A new general loss factor based on the F$_\beta$-measure that is applicable in any Boosting ensemble, to avoid bias of performance towards the negative class.
  \end{itemize} 
The PBoost algorithm has been compared to state of the art Boosting ensembles on synthetic, video and KEEL collection datasets in terms of both accuracy and computational complexity.  

The rest of the paper is structured as follows. Section 2 contains a review of literature on ensemble learning for class imbalance.
In Section 3, the proposed PBoost algorithm is described. The experimental methodology and results are presented in Sections 4 and 5, respectively.

\section{Boosting Ensemble Learning for Class Imbalance}
Learning from imbalanced data has been addressed in literature through data-level, algorithm-level, and cost-sensitive techniques. Ensemble learning methods exploit one or a combination of aforementioned techniques \cite{galar2012review} to handle imbalance.
Classifier ensembles can provide higher accuracy and robustness than a single classifier system by combining diverse classifiers~\cite{rokach2010ensemble}. 
Boosting is a common static ensemble learning algorithm initiated with AdaBoost~\cite{freund1995desicion} and improved in AdaBoost.M1 (for 2-class problems) and AdaBoost.M2 (for multiple-class problems)~\cite{freund1996experiments} to effectively promote a weak learner that performs slightly better than random guessing into a stronger ensemble.
In AdaBoost.M1 (Algo.\ref{Boosting}) samples are assigned weights that indicate their importance. These weights guide the learning process such that base classifiers in the ensemble focus on correct classification of more important samples as the learning iterations proceed.
Samples that are misclassified in each iteration gain more importance for the next iteration and more accurate base classifiers gain higher contribution in final decision.
These weights are used directly or for re-sampling training data, depending on the type of the base classifier being used. When the base classifier is from a type that is not designed to incorporate sample weights in its learning process (like SVMs), training data is re-sampled according to the weights of the samples. This case is considered here to explain the Boosting procedure.

Let's consider a two-class problem with $M$ labelled training samples $\mathbf{S}=\{(\mathbf{x}_i,y_i);i=1,...,M\}$ where $y_i \in \{-1,1\}$ that contains $M^+$ positive samples and $M^-$ negative samples. All samples in the dataset are initially associated with the same weight $\mathbf{W_1}(i)=1/M$, $i=1,...,M$. Then, a new training subset is re-sampled into $\mathbf{S'}$ with $\mathbf{W'}$ to trained classifier $C_e$. This classifier is tested on all training samples ($\mathbf{S}$) and a loss factor ($\epsilon_e$) is calculated as the sum of the weights of misclassified samples: 
%\vspace{-0.6cm}
\begin{eqnarray}\label{ep}
\epsilon_e=\underset{(i,Y_i):y_i \neq Y_i}{\sum}\mathbf{W_e(i)}
\end{eqnarray}
where $Y_i$ is the label associated with $\mathbf{x}_i$ by $C_e$.
If the classifier is too weak ($\epsilon_e>0.5$), the classifier is discarded and training set is re-sampled to train another classifier.
The loss factor is then used to define a weight update factor:
%\vspace{-0.6cm}
\begin{eqnarray}\label{alp}
\alpha_e=\frac{\epsilon_e}{1-\epsilon_e}.
\end{eqnarray}
The weights of the samples are then updated as:
%\vspace{-0.6cm}
\begin{eqnarray}\label{W}
 \textbf{W}_{\rm e+1}(i)=\textbf{W}_{\rm e}(i)\alpha_e^{\rm \frac{1}{2}\lvert y_i-Y_i \rvert} \ ,  
\end{eqnarray}
Weight vector is normalized such that the weights of the misclassified samples (more important samples) increase exponentially while the weights of the correctly classified samples decrease. $\alpha_e$ is also used to determine the contribution of the classifier in final predictions (Equation \ref{H}) so that more accurate classifiers play more important role in identifying the class of the input sample. This process is repeated for a predefined number of times to design $E$ classifiers. 
Considering $h_e(\textbf{x})$ as the output of $C_e$ (either a classification score or a label) for an input sample $\mathbf{x}$, final prediction of the ensemble is obtained from:
%\vspace{-0.6cm}
\begin{eqnarray}\label{H}
 H(\mathbf{x})= \sum_{e=1}^{E}h_e(\mathbf{x})\log \frac{1}{\alpha_e}
\end{eqnarray}

\begin{algorithm}[!htb] 
\SetKw{KwGoTo}{go to}
\algsetup{linenosize=\small}
\small
\setstretch{1}
\DontPrintSemicolon
\KwIn{Training set: $\mathbf{S}=\{(\mathbf{x}_i,y_i);i=1,...,M\}$,$y_i \in \{-1,1\}$ \newline
$\#$~of iterations: $E$ \newline
Test input : $\mathbf{X}$}
\KwOut{Prediction Function: $H(\cdot)$\\ }
%
%\smallskip
%\nonl // \textit{\textbf{Design Phase}} //
%\smallskip

Initialize $\textbf{W}_1(i)=\frac{1}{M}$ for $i=1,...,M$.

\For {$e=1,..,E$}{
\begin{enumerate} [i]
\item  Create new training set $\textbf{S}_{e}^{'}$ with weight distribution $\textbf{W}_{e}^{'}$. {\label{err} }
 \item   Train classifier $C_e$ on $\textbf{S}_{e}^{'}$ with $\textbf{W}_{e}^{'}$.  
 \item   Test $C_e$ on $\textbf{S}$ and get back a label set $\{Y_i, i=1,...,M\}$.  
 \item   Calculate the pseudo-loss for $\textbf{S}$ and $\textbf{W}_{e}$:
 
  $\epsilon_e=\underset{(i,Y_i):y_i \neq Y_i}{\sum}W_e(i) \ $.  
  \item If $\epsilon_e>0.5$ \KwGoTo step \ref{err}
 \item   Calculate the weight update parameter: 
  $\alpha_e=\frac{\epsilon_e}{1-\epsilon_e}$  
 \item   Update $\textbf{W}_{\rm e+1}(i)=\textbf{W}_{\rm e}(i)\alpha_e^{\rm \lvert y_i-Y_i \rvert/2} \ $  
 \item   Normalize  $\textbf{W}_{\rm e+1}$ such that: $\sum \textbf{W}_{\rm e+1}=1$.
 \end{enumerate}
  }  
%\smallskip
%\nonl // \textit{\textbf{Test Phase}} //
%\smallskip
%
%\For {$e=1,..,E$}{
%%\begin{enumerate} [i]
%\nonl Test $C_e$ classifier on $\textbf{x}$ and get back $h_e(\mathbf{x})$.
% %\end{enumerate}
%  }
 Output the final hypothesis:  $H(\cdot)= \sum_{e=1}^{E}h_e(\cdot)\log \frac{1}{\alpha_e}$	
 
\caption{AdaBoost.M1 ensemble learning method.}
\label{Boosting}		
\end{algorithm}

Analogous to most learning algorithms, AdaBoost is not effective to learn from imbalanced data for two reasons. Negative samples are the majority and when training data is re-sampled in line 2.i of AdaBoost (see Algo. \ref{Boosting}), they contribute more in $\mathbf{S'}$. Therefore, $C_e$ is trained biased to correct classification of this class. After that, when $C_e$ is tested on $\mathbf{S}$, loss factor in line 2.iv is calculated as a weighted error rate of classification. Again, negative samples contribute more in loss factor calculation and the weight update formula and classifiers contribution in final prediction become biased such that weight of negative samples increases for the next iteration and classifiers that mostly classify negative samples correctly get higher importance in final prediction of the ensemble. 
A taxonomy of methods in literature that modify AdaBoost to handle imbalance is presented in Figure~\ref{txn}. Based on the issue these approaches address, they are divided to two categories, data-level and algorithm-level methods that are presented in subsections 2.1 and 2.2, respectively.

\begin{figure}[!tb]
\centering
   \includegraphics[width=0.8\textwidth]{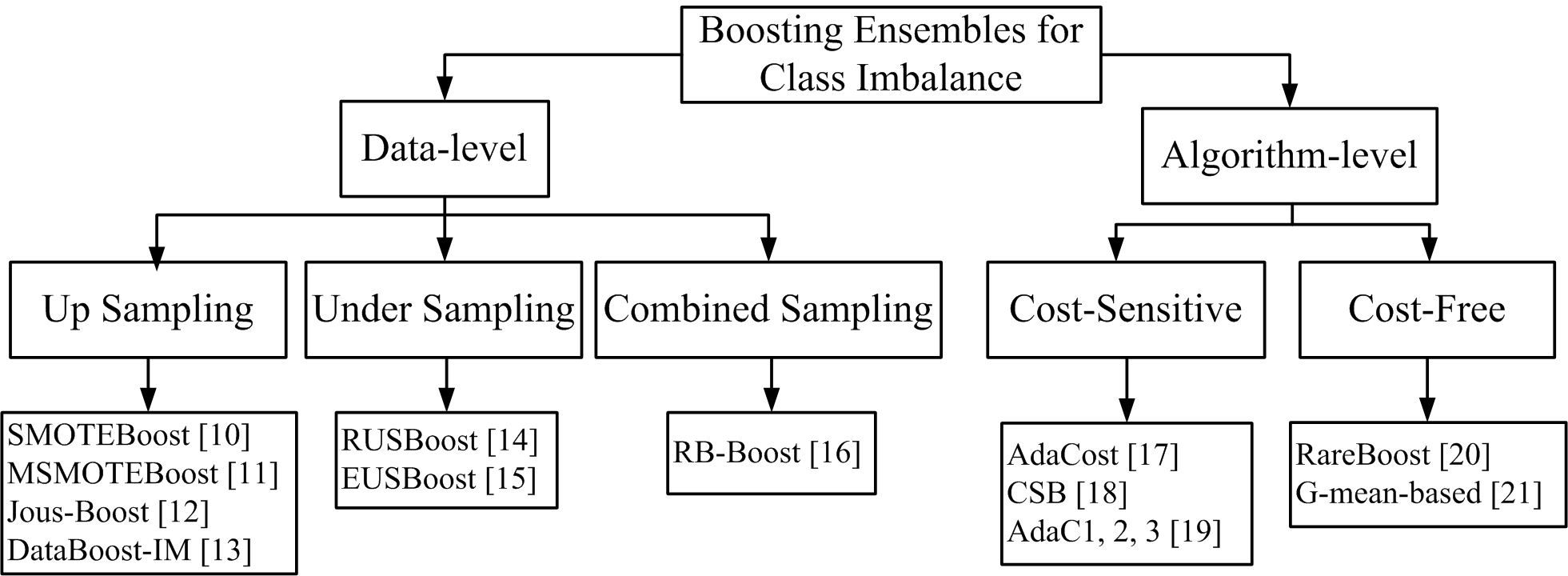}
   \caption{A taxonomy of Boosting ensembles learning methods specialized for imbalanced data.}
\label{txn}
\label{txn}
\end{figure} 

\subsection{Data-Level Methods:}
Class imbalance can be handled in Boosting ensembles through up-sampling the positive class, under-sampling the negative class or combination of them. A popular up-sampling Boosting approach is SMOTEBoost~\cite{chawla2003smoteboost} that integrates  Synthetic Minority Over-sampling Technique (SMOTE) into AdaBoost.M2. SMOTE creates synthetic samples by interpolating each positive sample with its k-nearest neighbours. MSMOTEBoost~\cite{hu2009msmote} use modified SMOTE (MSMOTE) by eliminating noisy samples and oversampling only safe samples.
Jous-Boost~\cite{mease2007cost} oversample the positive class by duplicating it, instead of creating new samples, and introduce perturbation (jittering) to this data in order to avoid overfitting. DataBoost-IM~\cite{guo2004learning} oversample difficult samples from both classes and integrates it into AdaBoost.M1 . 

Up-sampling techniques address the bias of performance in classifiers through balancing class distribution without loss of information. However,
up-sampling, in general, increase the number of samples and consequently increase the complexity of learning algorithms, and SMOTE involves additional computations due to interpolating each sample with its k-nearest neighbours to generate synthetic samples.

In under-sampling Boosting category, RUSBoost~\cite{seiffert2010rusboost} integrates random under-sampling (RUS) into AdaBoost.M1. RUSBoost is similar to AdaBoost presented in Algo.~\ref{Boosting} where in line 2.i of this algorithm, $\mathbf{S'}$ contains all positive samples and a randomly selected subset of negative class, often with a size equal to the positive class. The subsets of negative class selected randomly over iterations of RUSBoost could be highly correlated and the classifiers trained on them can lack in diversity, especially when the skew level of training data is high.
 The sample selection paradigm in RUSBoost is managed in EUSBoost~\cite{galar2013eusboost} to create less correlated subsets using evolutionary prototype selection~\cite{garcia2009evolutionary}. 

Some researchers combine SMOTE and RUS in AdaBoost to achieve greater diversity and avoid loss of information as in Random Balance Boosting (RB-Boost)~\cite{diez2015random}. RB-Boost combines SMOTE and RUS to create training subsets with random and different skew levels in AdaBoost.M1. 

Repetition of sampling in Boosting ensembles increase the chance of low correlation between subsets of data that are used for designing base classifiers and therefore maintain diversity among them. However, some potentially informative samples may be overlooked from these subsets in under-sampling process. In partitional approaches~\cite{icpram2016TUS,yan2003predicting,li2013constructing} bootstraps are selected without replacement either randomly~\cite{yan2003predicting}, by clustering~\cite{li2013constructing} or based on a prior knowledge from the application (like trajectories in video surveillance applications such as face re-identification~\cite{icpram2016TUS}). In these ensemble bootstraps are drawn from a set of negative samples that reduces size in each iteration. In other words, after selection of a bootstrap in each iteration, its samples are eliminated from the main set. In random partitioning of negative samples by Yan et al.~\cite{yan2003predicting} the negative data is randomly decomposed into a number of subsets and each subset, combined with the positive samples, is used to train a classifier. 
Li et al.~\cite{li2013constructing} partition negative data by clustering it using k-means in the feature space and then create an ensemble from the classifiers trained on each negative cluster and the positive samples. The contribution of the classifiers in the ensemble are then weighted based on the distance between the corresponding negative cluster and positive class. In~\cite{icpram2016TUS}, partitioning negative class is done by selecting samples from a set of trajectories that are formed based on the tracking information, as found in several video surveillance applications like face re-identification. In this approach, data from the trajectories are accumulated as the training iteration proceeds and therefore, base classifiers in the ensemble are trained on different imbalance levels to increase robustness of the ensemble to the possible variations in the skew level and complexity of operational data. 

In contrast to RUSBoost, these partitional approaches use all negative samples from partitions to design ensembles and avoid loss of information. However, not all samples are informative and using all samples for training may result in unnecessary time and memory complexity. Therefore, enhancing partitional methods with more intelligent sample selection and ensemble learning algorithm (like RUSBoost) can avoid information loss and excessive time complexity at the same time.

\subsection{Algorithm-Level Methods:}
Using standard error in Boosting ensemble learning algorithms biases their performance towards negative class. In literature this issue is avoided at the algorithm level using two types of techniques; those that employ two different misclassification cost factors, one for positive and another for negative classes and those that handle this issue without the use of cost factors.
Cost-sensitive Boosting methods including AdaCost~\cite{fan1999adacost}, CSB~\cite{ting2000comparative} and AdaC~\cite{sun2007cost}, embed different misclassification cost factors into loss function or weight update formula of AdaBoost.M2. 

Given $\mu_i$ as the cost factor of sample $\textbf{x}_i$, in AdaCost~\cite{fan1999adacost}, two cost adjustment functions are defined for each sample as $\phi_+=-0.5\mu_i+0.5$ and $\phi_-=0.5\mu_i+0.5$ and weight update formula is changed to:
\begin{eqnarray}
\textbf{W}_{\rm e+1}(i)=
\begin{cases}
\textbf{W}_{\rm e}(i)\exp\{-\alpha_e \phi_+{\rm \lvert y_i-Y_i \rvert}/2\}  & \text{for } Y_i=1\\
\textbf{W}_{\rm e}(i)\exp\{-\alpha_e \phi_-{\rm \lvert y_i-Y_i \rvert}/2\}  & \text{for } Y_i=-1 \\
\end{cases} 
\end{eqnarray} 

CSB~\cite{ting2000comparative} introduce two different cost factors for positive and negative classes as $\mu_+=1$ and $\mu_-\geq1$, respectively.
\begin{eqnarray}
\textbf{W}_{\rm e+1}(i)=
\begin{cases}
\textbf{W}_{\rm e}(i) \mu_+ \exp\{-\alpha_e {\rm \lvert y_i-Y_i \rvert}/2\}  & \text{for } Y_i=1\\
\textbf{W}_{\rm e}(i) \mu_- exp (-\alpha_e {\rm \lvert y_i-Y_i \rvert}/2\}  & \text{for } Y_i=-1 \\
\end{cases} 
\end{eqnarray} 
In AdaC1, 2, 3~\cite{sun2007cost} cost factors are embedded into the weight update formula in three different ways. Given $\mu_i \in [0,+\infty)$, 
in AdaC1:
\begin{eqnarray}
\alpha_e & = & \frac{1}{2} \ln \frac{1+\underset{i,y_i=Y_i}{\sum }\mu_i \textbf{W}_e(i)-\underset{i,y_i\neq Y_i}{\sum}\mu_i \textbf{W}_e(i)}{1-\underset{i,y_i=Y_i}{\sum }\mu_i \textbf{W}_e(i)+\underset{i,y_i\neq Y_i}{\sum}\mu_i \textbf{W}_e(i)}, \\\
\textbf{W}_{\rm e+1}(i) & = & \textbf{W}_{\rm e}(i)\exp\{-\alpha_e \mu_i Y_iy_i)
\end{eqnarray}
In AdaC2:
\begin{eqnarray}
\alpha_e & = & \frac{1}{2} \ln \frac{\underset{i,y_i=Y_i}{\sum }\mu_i \textbf{W}_e(i)}{\underset{i,y_i\neq Y_i}{\sum}\mu_i \textbf{W}_e(i)}, \\\
\textbf{W}_{\rm e+1}(i) & = & \mu_i \textbf{W}_{\rm e}(i)\exp\{-\alpha_e Y_i y_i\}
\end{eqnarray}
In AdaC3:
\begin{eqnarray}
\alpha_e & = & \frac{1}{2} \ln \frac{\sum_i \mu_i \textbf{W}_e(i)+\underset{i,y_i=Y_i}{\sum }\mu_i^2 \textbf{W}_e(i)-\underset{i,y_i\neq Y_i}{\sum}\mu_i^2 \textbf{W}_e(i)}{\sum_i \mu_i \textbf{W}_e(i)-\underset{i,y_i=Y_i}{\sum }\mu_i^2 \textbf{W}_e(i)+\underset{i,y_i\neq Y_i}{\sum}\mu_i^2 \textbf{W}_e(i)}, \\\
\textbf{W}_{\rm e+1}(i) & = & \mu_i \textbf{W}_{\rm e}(i)\exp\{-\alpha_e \mu_i Y_i y_i\}
\end{eqnarray}
In these cost-sensitive approaches by setting $\mu_+$ greater than $\mu_-$ the weights of misclassified samples from positive class increase more than that of the misclassified samples from negative class. In addition, the weights of the classifiers that correctly classify positive class better than the negative class is higher in final decision. Therefore,
these cost-sensitive approaches can make up for the usage of standard error rate in Boosting ensembles and allow adapting the performance by selecting proper cost factors based on the application. 
The drawback of these cost-sensitive approaches is that they require known $\mu_i$s that are usually set ad-hoc or by conducting a search in the space of possible costs for a dataset. 

Some cost-free approaches have been proposed to deal with the bias of performance caused by using standard error in Boosting ensembles.
In RareBoost~\cite{joshi2001evaluating}, two different $\alpha$s are defined for positive and negative classes as:
\begin{eqnarray}
 \alpha_e^+=\frac{1}{2}ln(\frac{TP_e}{FP_e})
\end{eqnarray}
\begin{eqnarray}
  \alpha_e^-=\frac{1}{2}ln(\frac{TN_e}{FN_e})
\end{eqnarray}
where $TP_e$ and $TN_e$ are the true positive and true negative counts, respectively. Then the weight update formula and final classification prediction are modified as:
\begin{eqnarray}
\textbf{W}_{\rm e+1}(i)=
\begin{cases}
\textbf{W}_{\rm e}(i) \exp\{-\alpha_e^+{\rm \lvert y_i-Y_i \rvert}/2\}  & \text{for } Y_i=1\\
\textbf{W}_{\rm e}(i) \exp\{-\alpha_e^-{\rm \lvert y_i-Y_i \rvert}/2\}  & \text{for } Y_i=-1 \\
\end{cases} 
\end{eqnarray}
\begin{eqnarray}
 H(\mathbf{x})= sign(\underset{e:h_e(\mathbf{x})\geq 0}{\sum}\alpha_e^+h_e(\mathbf{x})+\underset{e:h_e(\mathbf{x})< 0}{\sum}\alpha_e^-h_e(\mathbf{x})))
\end{eqnarray} 
Kim et al.~\cite{kim2015geometric} also define two different $\alpha_e$s for positive and negative classes as:
\begin{eqnarray}
\alpha_e^+=\frac{1-l^+}{l^+}, l^+=\frac{\underset{i;y_i=+1}{\sum} W_e(i) \lvert y_i-Y_i \rvert /2}{\underset{i;y_i=+1}{\sum}W_e(i)}
\end{eqnarray}
\begin{eqnarray}
\alpha_e^-=\frac{1-l^-}{l^-},l^-=\frac{\underset{i;y_i=-1}{\sum} W_e(i) \lvert y_i-Y_i \rvert /2}{\underset{i;y_i=-1}{\sum}W_e(i)}
\end{eqnarray}
where $l^+$ and $l^-$ are pseudo errors of classifier in classifying each class.
Finally:
\begin{eqnarray}\label{GG}
\alpha_e=\ln(\sqrt{\mu_i \alpha_e^+ \alpha_e^-}) \ ,
\end{eqnarray}%
$\mu_i$ is a multiplier to control the weight of each sample. The problem with this loss factor is that, if there are no misclassified samples in one class or in both classes, $\alpha_e$ is undefined.

Cost-free methods enhance the performance of Boosting ensembles without setting any cost factors and guide the learning process using a more suitable loss factor calculation since the use of weighted standard accuracy, as in original Boosting algorithm, biases the learning process towards correct classification of the negative class. These approaches inspire us to use a more suitable error calculation method in the proposed Boosting algorithm.
Therefore, it is relevant to review some of performance measures for imbalanced data classification in the following subsection.
\subsection{Performance Measures for Imbalanced Data Classification}
The trade-off between true positive rate (\mbox{TPR}) and false positive rate (\mbox{FPR}) for different operating settings can be traced with a Receiver-Operating Characteristic (ROC) curve. ROC curves are widely used to compare classifiers performance. This curve can also be summarized into a global scalar metric; area under the ROC curve (\mbox{AUC}). In addition, for a specific operating setting, G-mean performance measure is defined as the geometrical mean of \mbox{TPR} and \mbox{TNR} (or 1-\mbox{FPR}).
\begin{eqnarray} \label{G}
\mbox{G-mean} = \sqrt{\mbox{TNR} \cdot \mbox{TPR}} \ .
\end{eqnarray}
G-mean gives an equal weight to the efficiency of classifier in correct classification of both classes. 
ROC space does not adequately reflect the impact of imbalance~\cite{fawcett2006introduction} on performance because big variations in the number of misclassified negative class (\mbox{FP}) results in a small change in \mbox{FPR}, especially if a small increase in \mbox{TPR} can mask it. A suitable alternative for \mbox{TPR} is precision (\textit{Pr}) that compares the number of misclassified negative samples (\mbox{FP}) to the number of correctly classified positive samples (\mbox{TP}).
\begin{eqnarray}
\label{eq:Pr-skew}
\mbox{Pr}=\frac{\mbox{TPR}}{\mbox{TPR}+\lambda \mbox{FPR}} \ .
\end{eqnarray}
where $\lambda=M^-/M^+$.
It is evident that, as imbalance ($\lambda$) increases, any given decrease in \mbox{FPR} results in a higher reduction in \mbox{Pr}.
Therefore, precision drops severely when correct classification of positive class is in expense of high misclassification of negative class.
For different operating settings, precision-recall (PR) curve  depicts the trade off between precision and recall (\mbox{Re}=\mbox{TPR}) when data is imbalanced. Inspired from \mbox{AUC}, area under PR curve (\mbox{AUPR}) can also be used to compare classifiers globally over all operating settings.

For a specific operating setting, precision and recall can be weighted and combined into a local performance metric, the F$_\beta$-measure:
\begin{eqnarray}
F_\beta=\frac{(1+\beta^2)\mbox{Pr}.\mbox{Re}}{\beta^2\mbox{Pr}+\mbox{Re}}=\frac{(1+\beta^2)\mbox{TP}}{(1+\beta^2)\mbox{TP}+\mbox{FP}+\beta^2 \mbox{FN}}
\end{eqnarray}
Although \mbox{Pr} is very sensitive to misclassification of negative samples due to their abundance, with selection of suitable $\beta$ in F$_\beta$-measure, this sensitivity can be controlled. With $\beta\geq1$, this sensitivity reduces and F$_\beta$ gives a higher importance to correct classification of positive class. With $\beta<1$, this sensitivity increases and correct classification of negative samples becomes the priority.

Another metric that takes imbalance into account is expected cost which is calculated as:
\begin{eqnarray} \label{NEC}
\mbox{EC} & = & \pi \cdot \mbox{FNR} \cdot C_{\mbox{FN}}+(1-\pi) \cdot \mbox{FPR} \cdot C_{\mbox{FP}} 
\end{eqnarray}
where $\pi=\rfrac{M^+}{M}$ is the proportion of positive samples, and $C_{\rm FN}$ and $C_{\rm FP}$ are the misclassification cost of positive and negative classes, respectively.
For higher values of $\lambda$ (lower values of $\pi<0.5$) and $C_{\rm FN}=C_{\rm FP}$, \mbox{EC} is more sensitive to increase of \mbox{FPR} rather than \mbox{FNR}. For a specific value of $\pi$, the sensitivity of \mbox{EC} to each of \mbox{FNR} and \mbox{FPR} can be controlled by tuning $C_{\rm FN}$ and $C_{\rm FP}$.

Some authors define variants of the existing performance metrics by accounting for $\pi$~\cite{garcia2010theoretical,hernandez2012unified,flach2015precision}.
For example, in ~\cite{hernandez2012unified}, the authors define
a measure of expected accuracy
in terms of\mbox{ AUC } as $\pi(1-\pi)(2AUC-1)+1/2$, and
precision-recall gain (PRG) curve~\cite{flach2015precision} normalize precision and recall in terms of $\pi$ as:
\begin{eqnarray}
\mbox{precG}=\frac{\mbox{Pr}-\pi}{(1-\pi)\mbox{Pr}},\\
\mbox{recG}=\frac{\mbox{Re}-\pi}{(1-\pi)\mbox{Re}}.
\end{eqnarray} 
ROC, PR, and PRG curves are usually produced by varying decision threshold over real valued scores output by classifiers.
Consequently, using areas under the curves to generate global metrics or their variants in Boosting ensembles may increase computational cost of these algorithms. Masking the effect of imbalance in performance metrics as done in \cite{garcia2010theoretical,hernandez2012unified,flach2015precision} can misguide the learning process of Boosting ensembles and bias the performance towards negative class.
The $F_\beta$-measure and expected cost are more suitable performance metrics that take into account correct classification of both classes considering the level of imbalance. However, adjusting the sensitivity of the $F_\beta$-measure to the correct classification of the positive and the negative class is easier than expected cost. The reason is that adjusting the $F_\beta$-measure involves tuning $\beta$ rather than 
tuning two factors ($C_{\rm FN}$ and $C_{\rm FP}$) for expected cost. Nevertheless, EC is a more suitable performance metric for cost-sensitive learning algorithms.

In this paper, the $F_\beta$-measure, the most frequently
used measures for performance evaluation in class imbalance
learning, is employed to modify the loss factor calculation in Boosting ensembles, to avoid biasing the performance towards negative class.

\section{Progressive Boosting for Learning Ensembles from Imbalanced Data:}

The Progressive Boosting (PBoost) learning method is proposed to sustain a high level of performance over a range of imbalance and complexity levels in the data seen during operations.
This method follows a static approach, and learns ensembles based on a combination of under-sampling and cost-free adjustment of Boosting ensemble learning. 

With the PBoost algorithm, negative class is partitioned into disjoint subsets. These partitions are accumulated into a temporary design set progressively as learning iterations proceed. In each iteration, a subset of this temporary set is used for training a classifier such that the most important samples plus samples from the new partition are given an equally high opportunity to be used in training a base classifier. Loss of information is therefore avoided and ensemble diversity is increased.
The trained classifier is then validated on the temporary set that contains all positive samples and only those negative partitions that have already been used in previous training iterations. As the temporary set grows, its imbalance level increases and therefore, the ensemble's robustness to diverse levels of skew and decision bound complexities during operations is increased. In PBoost, the error of the classifier is determined based on its ability to correctly classify both positive and negative classes. This loss factor plays an important role in determining the contribution of classifiers in final prediction, and in selection criteria of samples for designing the next classifiers.

\begin{figure}[!htb]
\centering
\includegraphics[width=0.94\textwidth]{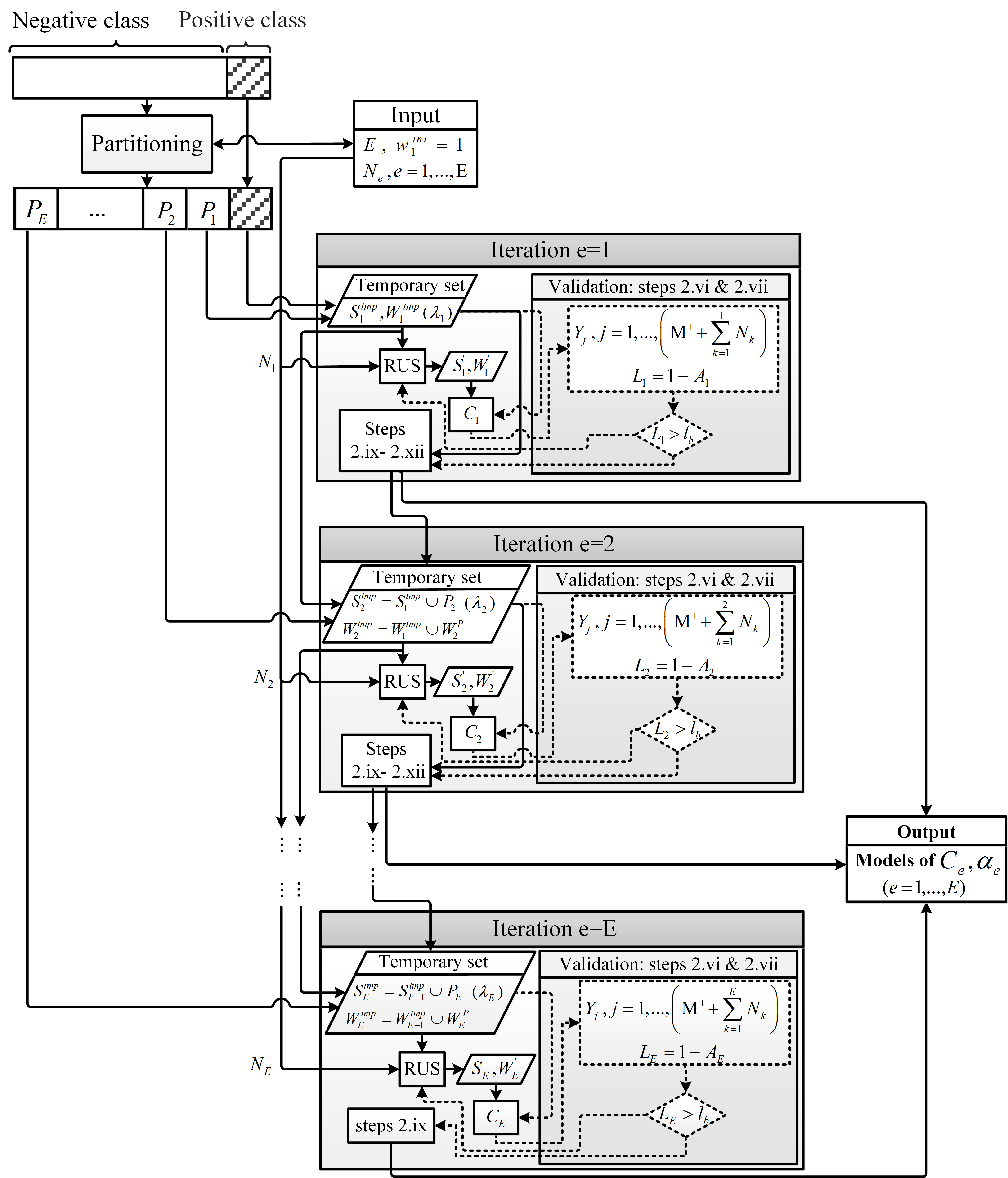}
\caption{Block diagram representation of PBoost learning method.}
\label{PB}
\end{figure}

The progressive Boosting method is presented in Algo.~\ref{PBoost} and Figure \ref{PB}. Its main steps are explained in the following.
In the proposed algorithm, the negative samples are first regrouped into $E$ disjoint partitions $\textbf{P}_e$, where $ e=1,..,E$, one per classifier in the ensemble (line 1). $E$ and the number of negative samples in each partition $N_e$ varies and depends on the partitioning method and the data distribution. There are several possible ways to partition the negative samples into disjoint subsets in literature~\cite{xu2005survey} e.g., prototype-based methods like k-means and GMM algorithms, affinity-based methods like spectral, normalized-cut and sub-space algorithms to represent the negatives, and thus define partitions (number of clusters and association of data to clusters).
Two partitioning techniques have been used in literature to partition data to learn ensembles from imbalanced data: Random Under-Sampling without replacement (we call RUSwR in this paper)~\cite{yan2003predicting} , and Cluster Under-Sampling (CUS)~\cite{li2013constructing}.
In some applications the data is already partitioned, like binarization of multi-class classification problems using one-vs-all strategy.
In some applications the data may be grouped based on some contextual or application-based knowledge of data. For example, Trajectory Under-Sampling (TUS) is applicable in video surveillance applications where samples captured for a same individual are regrouped into a trajectory~\cite{icpram2016TUS}.
In the case of random under-sampling without replacement $E$ is preselected and $N_e$ takes a fixed random value $N_e \in [M^+/2, 2M^+]$ such that $\sum^E_{e=1}N_e=M^-$. In the case of CUS and TUS, $E$ and $N_e$ depend on the number of samples that are assigned to each partition by the clustering algorithm and the tracker, respectively.

Given a training data set $\textbf{S}$, one partition $\textbf{P}_e$ is selected in each iteration and added to a temporary set $\textbf{S}_e^{\rm tmp}$ (line 5.ii) which initially contains the positive samples. The same initial weight $w_{\rm ini}$ is assigned to the samples in the new partition creating a weight vector $\textbf{W}_e^p$ (line 5.i) which is also added to a temporary weight set $\textbf{W}_e^{\rm tmp}$ (line 5.ii).
In the next step (line 5.iv), $N_e$ samples from the temporary set $\textbf{S}_e^{\rm tmp}$ are selected through random under-sampling to create a new subset $\textbf{S}_e^{'}$ with the weight distribution of $\textbf{W}_e^{'}$.
A classifier $C_e$ is trained on $\textbf{S}_e^{'}$ (line 5.v). Then it is tested on the whole temporary set $\textbf{S}_e^{\rm tmp}$ that has an imbalance level of \begin{math}\lambda_e=1:\sum^e_{f=1}\rfrac{N_f}{M^+}\end{math} (line 5.vi).
 Therefore, the classifiers in this ensemble are in fact validated on data subsets with a growing level of imbalance and complexity.

After that, a new loss factor is calculated that adapt Boosting algorithm for classifying imbalanced data based on F$_\beta$-measure (line 5.vii) because F$_\beta$-measure is more sensitive to imbalance and at the same time allows us to give more importance to one class than the other. 

To calculate the loss factor, we first split the temporary weight vector $\textbf{W}_e^{\rm tmp}$ to two weight matrices for positive $\textbf{W}_e^{\rm tmp,+}$ and negative $\textbf{W}_e^{\rm tmp,-}$ classes. The size of $\textbf{W}_e^{\rm tmp,+}$ is $M^+$ and the size of $\textbf{W}_e^{\rm tmp,-}$ is $\sum^e_{f=1}N_f$, and:
\begin{eqnarray}
\textbf{W}_e^{\rm tmp,+} & = & \{\textbf{W}^{\rm tmp}_e(j), j=1,...,(M^++\sum^e_{f=1}N_f) | y_j=1\} \label{W+} \ , \\
\textbf{W}_e^{\rm tmp,-} & = & \{\textbf{W}^{\rm tmp}_e(j), j=1,...,(M^++\sum^e_{f=1}N_f) | y_j=-1\} \label{W-} \ .
\end{eqnarray}
Then, weighted versions of true positive, false positive, true negative and false negative %
counts are defined as:
\begin{eqnarray}
{\rm TP}_e & = & \underset{k: Y_k=1}{\sum}\textbf{W}_e^{\rm tmp,+}(k),k=1,...,M^+ \label{TP} \\
{\rm FP}_e & = & \underset{k: Y_k=1}{\sum}\textbf{W}_e^{\rm tmp,-}(k),k=1,...,\sum^e_{f=1}N_f \label{FP} \\
{\rm TN}_e & = & \underset{k: Y_k=-1}{\sum}\textbf{W}_e^{\rm tmp,-}(k),k=1,...,\sum^e_{f=1}N_f \label{TN} \\
{\rm FN}_e & = & \underset{k: Y_k=-1}{\sum}\textbf{W}_e^{\rm tmp,+}(k),k=1,...,M^+ \label{FN}
\end{eqnarray}
Based on these values, the accuracy of a classifier is computed in terms of $F_\beta$-measure as:
\begin{eqnarray}
A_F & = & \frac{(1+\beta^2){\rm TP}_e}{(1+\beta^2){\rm TP}_e+{\rm FP}_e+\beta^2{\rm FN}_e} \label{af} \ ,
\end{eqnarray}
To measure the error of the classifiers, the corresponding loss factor is defined as:
\begin{eqnarray}
L_e & =& 1-A_F = \frac{{\rm FP}_e+\beta^2{\rm FN}_e}{(1+\beta^2){\rm TP}_e+{\rm FP}_e+\beta^2{\rm FN}_e} \ . \label{lf} 
\end{eqnarray}
The condition $\epsilon_e>0.5$
in line (v) of AdaBoost.M1 (Algo. ~\ref{Boosting}) means
that classifiers in a Boosting ensemble should perform better than random guessing.
When F$_\beta$-measure is used as the evaluation metric, the base classifier to beat is the one that predicts everything as positive~\cite{flach2015precision}. % in what sense is such a classifier "random"?
Therefore, when the loss factor is calculated using Eq.~\eqref{lf}, the accuracy criterion of 0.5 in AdaBoost.M1 
should be replaced by $l_b=\frac{M^-}{(1+\beta^2)M^++M^-}$ (line 5.vii). 

After calculation of $\alpha_e$ (line 5.ix) from: 
\vspace{-0.5cm}
\begin{eqnarray}
\alpha_e & = & \frac{L_e}{1-L_e} , \label{alph}\,
\end{eqnarray}
the weights in the temporary set $\textbf{W}_e^{\rm tmp}$ are updated (line 5.x) as: 
\begin{eqnarray}
\textbf{W}^{\rm tmp}_{\rm e+1}(j) & = & \textbf{W}^{\rm tmp}_{\rm e}(j)\ \alpha_e^{\rm \lvert y_j-Y_j \rvert /2} \  . \label{Wtemp}
\end{eqnarray}

\begin{algorithm} [!htb]
\SetKw{KwGoTo}{go to}
\algsetup{linenosize=\small}
\small
%\linespread{1}
\setstretch{1}
\DontPrintSemicolon
\KwIn{Training set: $\mathbf{S}=\{(\mathbf{x}_i,y_i);i=1,...,M\},y_i \in \{-1,1\}, M=M^-+M^+ $%\newline
%Ensemble size: $E$ 
%Test input : $\mathbf{X}$
}
\KwOut{Predicted score or label: $H(\mathbf{\cdot})$\\ }
%\smallskip
%\nonl // \textit{\textbf{Design Phase}} //
%\smallskip

Partition non-target samples from $\textbf{S}$ into $E$ clusters $\{\textbf{P}_e;e=1,...,E\}$.

%Include tmporary training set $\textbf{S}_1^{\rm tmp}$ with only target samples and corresponding tmporary weight distribution $\textbf{W}_{e}^{\rm tmp}$ as ones.
Create a temporary training set and weight vector: $\textbf{S}_1^{\rm tmp} \leftarrow \{(\mathbf{x}_i,y_i) \in \mathbf{S}|y_i=1\}$ and $\textbf{W}_{1}^{\rm tmp}(k)=1, k=1,...,M^+$.

Initialize $w_1^{\rm ini}=1$.

Set $l_b=\frac{M^-}{(1+\beta^2)M^++M^-}$ 

\For {$e=1,..,E$}{\small
\begin{enumerate} [i]
\item  Initialize weight distribution of $\textbf{P}_e$ as $ \textbf{W}_e^p(k)=w_e^{\rm ini},k=1,...,N_e$.\hspace{-5cm}\tcp*{$N_e$ is the size of $\textbf{P}_e$.}
% \item  Concatenate $\textbf{P}_e$ into $\textbf{S}_e^{\rm tmp}$ and $\textbf{W}_e^p$ into $\textbf{W}_{e}^{\rm tmp}$.  
\item 
$ \mathbf S_e^{\rm tmp} \leftarrow \mathbf S_e^{\rm tmp} \bigcup \mathbf P_e$ , $ \mathbf W_e^{\rm tmp} \leftarrow \mathbf W_e^{\rm tmp} \bigcup \mathbf W_e^p$
 \item  Normalize  $\textbf{W}_{e}^{\rm tmp}$ such that: $\sum \textbf{W}_{e}^{\rm tmp}=1$.  
% \item  Based on $ N_e $, create training subset $\textbf{S}_e^{'}$ from $\textbf{S}_e^{\rm tmp}$ with weight distribution $\textbf{W}_{e}^{'}$ using RUS.  {\label{err1}}
\item  Randomly select $ N_e $ non-target samples from $\textbf{S}_e^{\rm tmp}$ based on $\textbf{W}_{e}^{\rm tmp}$, to create a training subset $\textbf{S}_e^{'}$ with $\textbf{W}_{e}^{'}$.  {\label{errl}}
 \item  Train $C_e$ on $\textbf{S}_e^{'}$ with $\textbf{W}_{e}^{'}$.  
 \item  Test $C_e$ on $\textbf{S}_e^{\rm tmp}$ and get back labels $Y_j$,$j=1,...,(M^++\sum^e_{f=1}N_f)$.
 \item  Calculate the pseudo-loss for $\textbf{S}_e^{\rm tmp}$ from $\textbf{W}_{e}^{\rm tmp}$ (using Equations \ref{W+} to \ref{FN}): 
 $\textbf{W}_e^{\rm tmp,+}  =  \{\textbf{W}^{\rm tmp}_e(j), j=1,...,(M^++\sum^e_{f=1}N_f) | y_j=1\} , $ \newline
 $\textbf{W}_e^{\rm tmp,-}  =  \{\textbf{W}^{\rm tmp}_e(j), j=1,...,(M^++\sum^e_{f=1}N_f) | y_j=-1\}  , $ \newline
 ${\rm TP}_e  =  \underset{(k,Y_k): Y_k=1}{\sum}\textbf{W}_e^{\rm tmp,+}(k),k=1,...,M^+ , $ \newline
 ${\rm FP}_e  =  \underset{(k,Y_k): Y_k=1}{\sum}\textbf{W}_e^{\rm tmp,-}(k),k=1,...,\sum^e_{f=1}N_f , $ \newline
 ${\rm TN}_e  =  \underset{(k,Y_k): Y_k=-1}{\sum}\textbf{W}_e^{\rm tmp,-}(k),k=1,...,\sum^e_{f=1}N_f , $ \newline
 ${\rm FN}_e  =  \underset{(k,Y_k): Y_k=-1}{\sum}\textbf{W}_e^{\rm tmp,+}(k),k=1,...,M^+ , $  \newline
 $L_e  = 1-A_F = \frac{{\rm FP}_e+\beta^2{\rm FN}_e}{(1+\beta^2){\rm TP}_e+{\rm FP}_e+\beta^2{\rm FN}_e}. $
\item If $L_e>l_b$ \KwGoTo step \ref{errl}
  %\item Calculate variation in the complexity of decision bound of $C_e$ as $\gamma_e$.
  %\item If $L_e>0.5$ or $\gamma_e<?$ \KwGoTo \ref{err1}\
%  \item If $\epsilon_e>0.5$ or $\gamma_e<?$ \KwGoTo \ref{err1}\
 %\item  $G_e$ $\gets$ G-mean attained by $C_e$ on $S_e^{\rm tmp}$.   
 \item  Calculate the weight update parameter: 
  $\alpha_e=\frac{L_e}{1-L_e}$% \times \frac{1-G_e}{G_e} $  
 \item  Update $\textbf{W}^{\rm tmp}_{\rm e+1}(j) = \textbf{W}^{\rm tmp}_{\rm e}(j)\alpha_e^{\rm \lvert y_j-Y_j \rvert/2}$
 \item  Normalize  $\textbf{W}_{\rm e+1}^{\rm tmp}$ such that: $\sum \textbf{W}_{\rm e+1}^{\rm tmp}=1$. 
  \item   Set $w_{\rm e+1}^{\rm ini}=\max(\textbf{W}_{\rm e}^{\rm tmp}),y_j=-1$
   \end{enumerate}
	}
%\smallskip
%\nonl // \textit{\textbf{Test Phase}} //
%\smallskip

%  \For {$e=1,..,E$}{
%%\begin{enumerate} [i]
% \nonl Test $C_e$ classifier on $\textbf{x}$ and get back $h_e(\mathbf{x})$.
%  %\end{enumerate}
%  }
 Output the final hypothesis:  $H(\cdot)= \sum_{e=1}^{E}h_e(\cdot)\log \frac{1}{\alpha_e}$ \tcp*{$h_e(\cdot)$ is the output of $C_e$.}
  
\caption{Progressive Boosting ensemble learning method.}
\label{PBoost}
\end{algorithm}

Even though it is desirable to limit the loss of information during under-sampling of data, some samples (like borderline samples) are of more interest than others for training classifiers in the ensemble. 
In Boosting ensembles, these samples are often detected as misclassified samples because borderline samples play more important role in defining the decision bound and they are more likely to be misclassified. More importance is given to these samples by assigning higher weights to them, so that they have a higher chance to be included in training subset(s). In the proposed PBoost ensemble, after normalization of $\textbf{W}_e^{\rm tmp}$, its maximum value among negative samples is selected as the initial weight for the next iteration (line 5.xii):
\begin{eqnarray}
w^{\rm ini}_{\rm e+1} & = & \underset{y_j=-1}{\max}\{\textbf{W}^{\rm tmp}_{\rm e}(j)\},\ j=1,...,M^++\sum^e_{f=1}N_f.
\end{eqnarray}
This value corresponds to the weight of more important misclassified negative samples. 
Therefore, in each iteration, new samples and misclassified samples from previous iterations have more chance to be included in the training subset. 
Finally, $\alpha_e$ is used to obtain the final class prediction of the ensemble from \eqref{H} (line 6).

PBoost is somewhat inspired from RUSBoost, but differs in three main respects. First, during each iteration, instead of random under-sampling with replacement, most of training negative samples are selected from disjoint partitions. Consequently, repeatedly selection of the same samples over all iterations and information loss is avoided while the diversity increases.
Second, instead of validating the classifiers on all samples, the classifiers are validated only on a subset of training set that grows in size and imbalance over iterations. Therefore, robustness to different levels of data imbalance and complexity increases, and the computations complexity of validation step decreases significantly. 
Third, instead of weighted accuracy, F-measure, an imbalance-compatible performance metric, avoids biasing performance towards negative class.

\section{Experimental Methodology}
In our experiments, the proposed PBoost ensemble learning method is assessed and compared with AdaBoost.M1~\cite{freund1996experiments}, and one state of the art method from each family of the data-level approaches reviewed in Section 2 including SMOTEBoost~\cite{chawla2003smoteboost}, RUSBoost~\cite{seiffert2010rusboost}, and RB-Boost~\cite{diez2015random}. 
The datasets that are used for the experiments include: (1) A set of synthetic 2D data sets in which the level of skew and overlap between classes are controllable, (2) the Face In Action (FIA) video database~\cite{goh2005cmu} that emulates a passport checking scenario in face re-identification application, (3) a set of 21 real-world
problems from the KEEL dataset repository~\cite{alcala2010keel}. %\footnote{(http://www.KEEL.es/dataset.php)}.
The rest of this section presents the datasets used in the experiments followed by the experimental and evaluation protocols. 

\subsection{Datasets}
\subsubsection{Synthetic Dataset}
The performance of classification systems may vary on different levels of overlap and skew between classes in both training and test data. Therefore, in our experiments on synthetic data, different synthetic datasets with different overlap and skew levels are generated and used to compare classification systems.

The data is generated to emulate both binarization of a multi-class classification problem when the classification strategy is one versus all and binary classification problems where there is no prior knowledge of optimal partitions.
The samples of both positive and negative classes are generated in groups of samples distributed in a normal distribution. The samples from one normal distribution are considered as positive class and all other samples are considered as negative class.

To generate the 2D synthetic data, $M^+=100$ positive class samples are generated with a normal distribution as $N(m_+,\sigma_+)$, where $m_+=(0,0)$ and $\sigma_+=[\begin{smallmatrix} 1&0\\ 0&1 \end{smallmatrix}]$ indicate the mean and covariance matrix of this distribution, respectively. Then, $T^-=100$ points are selected randomly in a uniform distribution around $m_+$. These points ($m_{-,j}$, $j=1,...,T^-$) are generated as the mean of $T^-$ Normal distributions ($N(m_{-,j},\sigma_-)$, $j=1,\dotsc,T^-$) for negative class where $\sigma_-=\sigma_+$. Each normal distribution contains $M^+=100$ samples and is considered as an ideal cluster of negative class (used for PCUS$_i$).

The mean of these clusters ($m_{-,j}$s) keep a margin distance $\delta$ from $m_+$. This margin is used to control the level of overlap between positive and negative classes. 

For the experiments, we selected the parameter $\delta$ as 0.1 (maximum overlap) and 0.2 (medium overlap). 
For each overlap level, each normal distribution is randomly divided into two subsets for design and testing. Then the design subsets are divided into 5 folds considering one fold for validation and 4 folds for training. Five replications is possible by alternating the validation fold in each iteration and by reversing the role of design and testing subsets, a total of 10 replications is achieved.

Two settings are considered for the experiments that differ based on the skew level of training data where $\lambda_{\rm train}=1:\rfrac{M^-}{M^+}$ is set to 1:50 in one setting and to 1:20 in the other.
When $\lambda_{\rm train}=1:50$, only 50 clusters from the negative class are used for training. The objective is to compare different classification algorithms when they are designed on different levels of imbalance.
Properties of training data generated with these settings are summarized in Table~\ref{dt} and examples of training data generated with these settings are presented in Figure~\ref{synthOV}.
% and the corresponding clusters generated by k-means and considering Dunn index to find the optimal number of clusters are shown in Figure ~\ref{synthOVC}.

In a similar way, four settings are considered for testing under different imbalance levels ($\lambda_{\rm test}=\{1:1, 1:20, 1:50, 1:100\}$) to evaluate the robustness of the classification algorithms over varying skew levels of data during operation. Examples of synthetic test data corresponding to setting $D_1$ is presented in Figure \ref{test}.
\begin{table}[!b]
\centering
\renewcommand\multirowsetup{\centering}
\caption{Settings used for data generation.}
\label{dt}
\begin{tabular}{lccc}\hline\hline 
                & $D_1$   & $D_2$   & $D_3$    \\\hline 
                \\[-0.8em]
$\lambda_{tr}$ & 1:50 & 1:50 & 1:20  \\
$\delta$        & 0.2  & 0.1  & 0.2   \\\hline\hline 
\end{tabular}
\end{table}
%, 

\begin{figure}[!b]
\begin{center}
        \begin{subfigure}[b]{0.3\textwidth}
                \includegraphics[width=\textwidth]{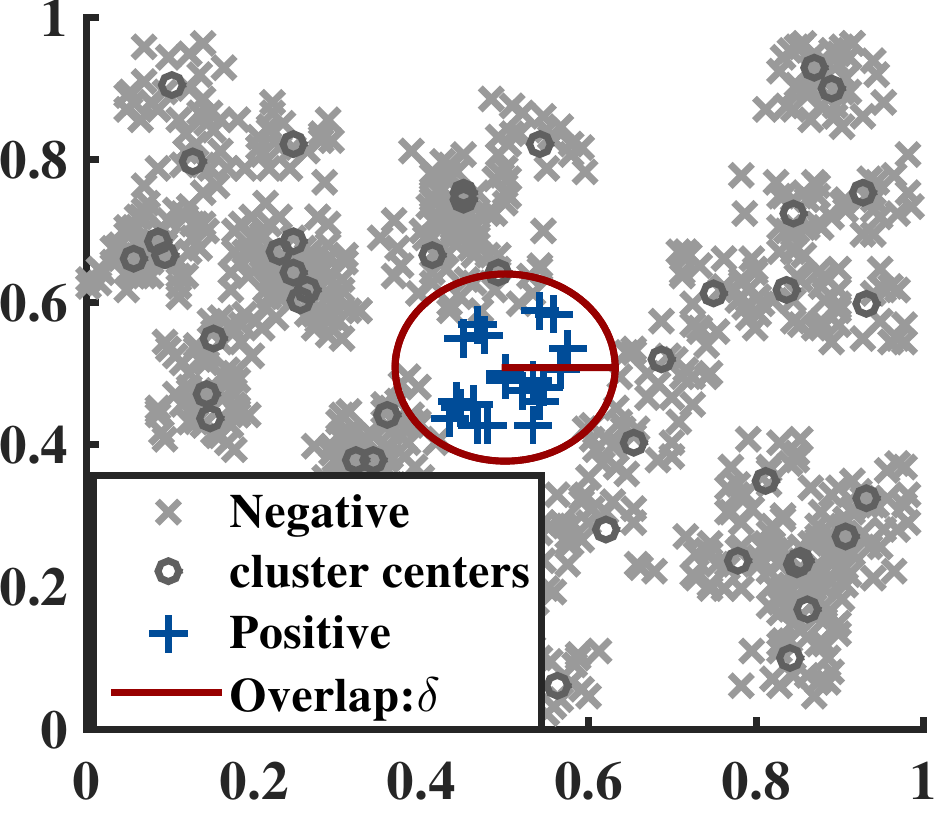}
                \caption{$D_1$.}
        \end{subfigure}
        \begin{subfigure}[b]{0.3\textwidth}
                \includegraphics[width=\textwidth]{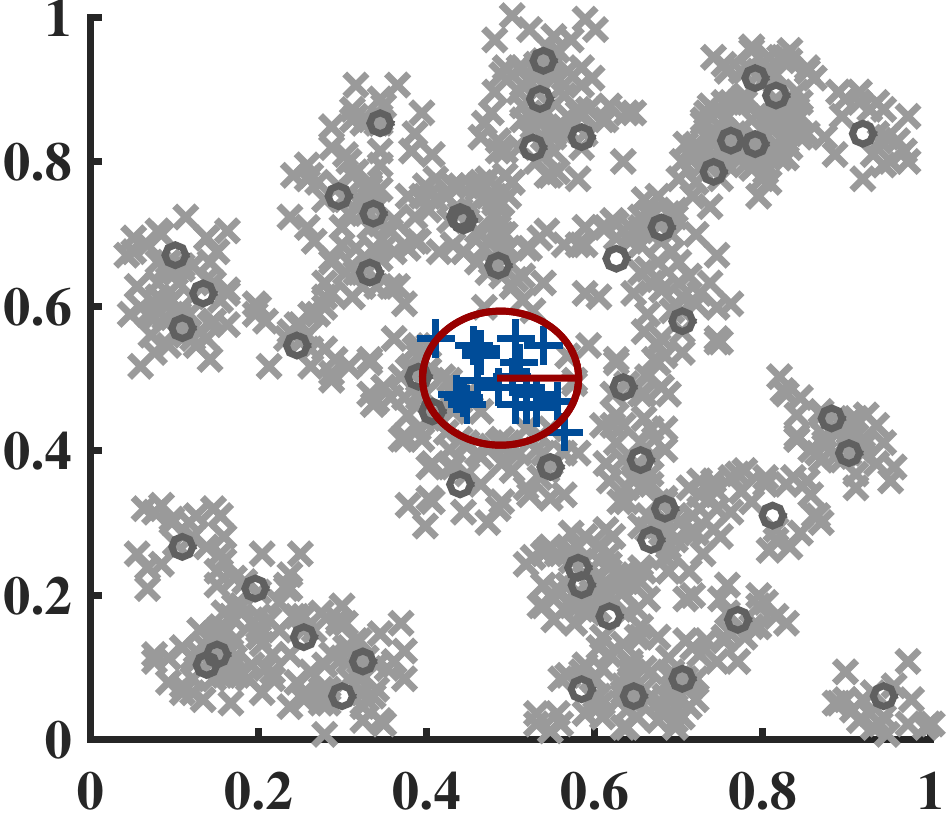}
                \caption{$D_2$.}
        \end{subfigure}
        \begin{subfigure}[b]{0.3\textwidth}
                \includegraphics[width=\textwidth]{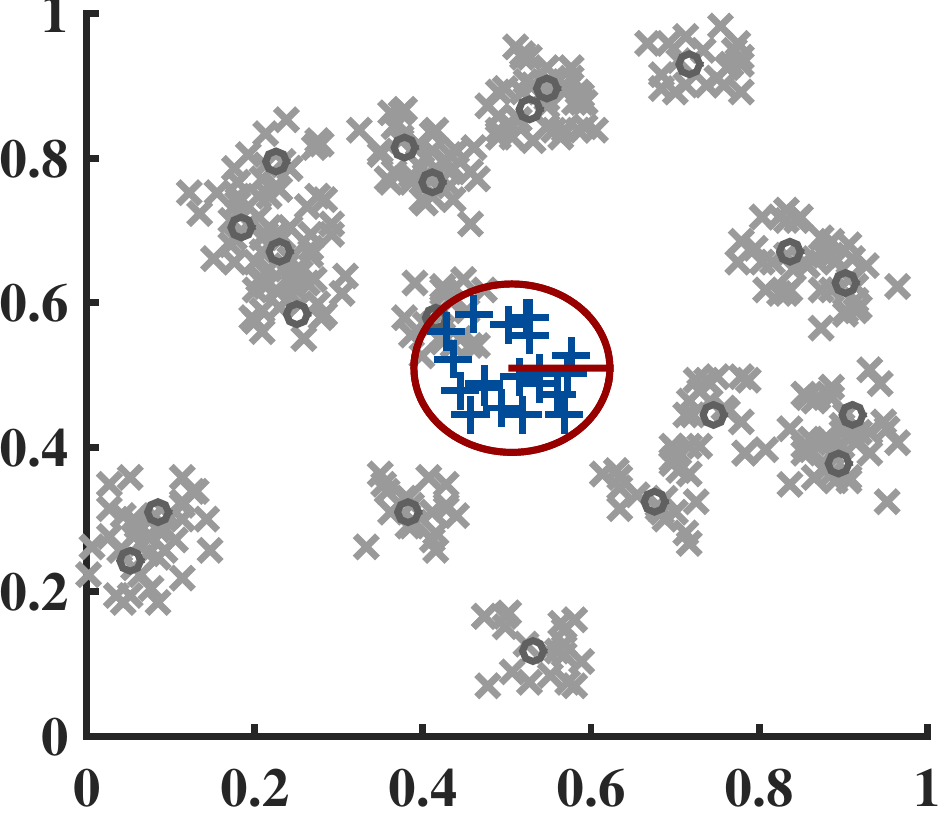}
                \caption{$D_3$.}
        \end{subfigure}
\end{center}
   \caption{Examples of synthetic training data generated for experiments.}
\label{synthOV}
\label{synthOV}
\end{figure}

\begin{figure}[!htb]
\begin{center}
        \begin{subfigure}[b]{0.3\textwidth}
                \includegraphics[width=\textwidth]{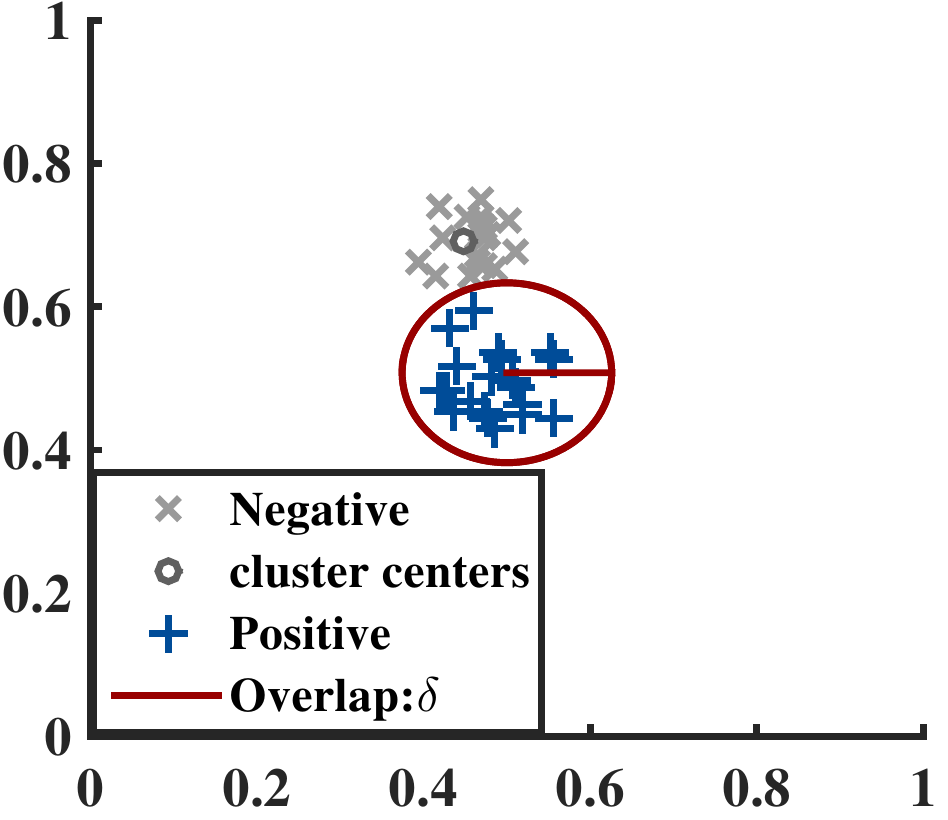}
                \caption{$\lambda_{\rm test}=1:1$.}
        \end{subfigure}
        \begin{subfigure}[b]{0.3\textwidth}
                \includegraphics[width=\textwidth]{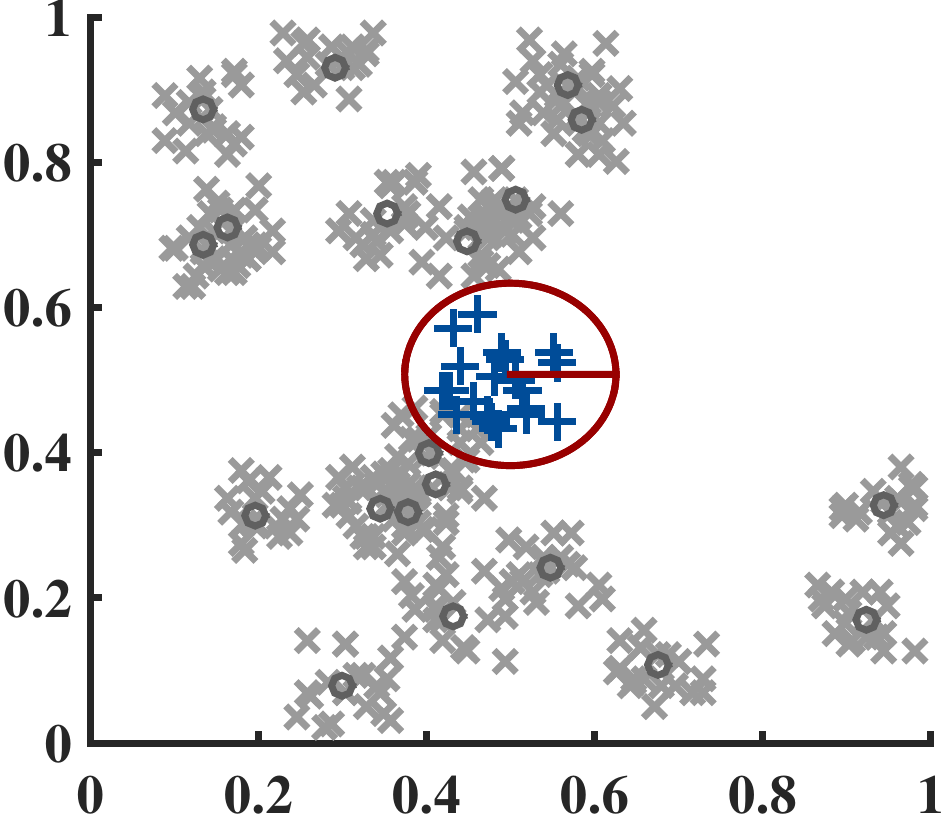}
                \caption{$\lambda_{\rm test}=1:20$.}
        \end{subfigure}        
        \begin{subfigure}[b]{0.3\textwidth}
                \includegraphics[width=\textwidth]{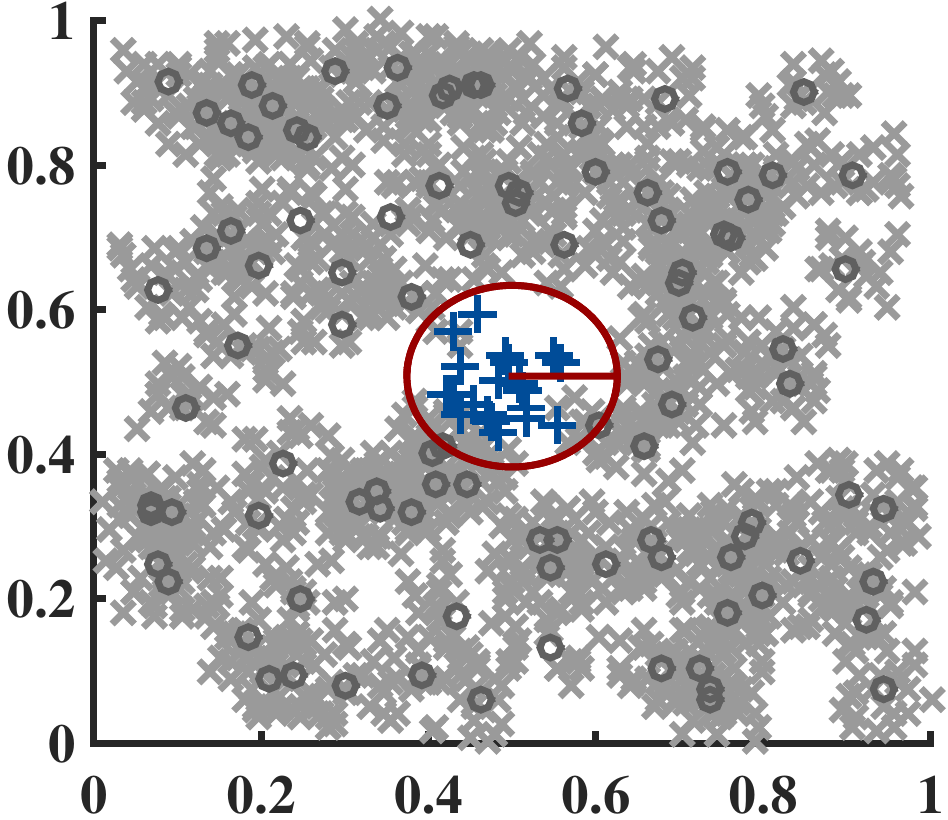}
                \caption{$\lambda_{\rm test}=1:100$.}
        \end{subfigure}
\end{center}
   \caption{Examples of synthetic test data generated with $\delta=0.2$ and different skew levels.}
\label{test}
\label{test}
\end{figure}

\subsubsection{Face Re-Identification Dataset}
Face re-identification is a video surveillance application where individuals in video streams are recognized at different time instants and/or locations over a network of distributed cameras.
Non-target faces captured in videos under various challenging conditions are compared to those of the target individual using a video-to-video face recognition system. One important challenge in this application is that the number of faces captured from the target individual (positive class) is typically limited and greatly outnumbered by non-target ones (negative class) \cite{de2015partially,pagano2014adaptive,icpram2016TUS}. 

In this classification problem, face captures from each individual may be grouped by the face tracker to trajectories.
Given a video stream, an efficient face tracking system, groups face captures from the same individual to a trajectory. A trajectory 
is Regions Of Interest (ROIs) of a same person regrouped with a face tracker according to a high quality tracking information collected by the tracker that follows the location of the ROIs over consecutive video frames.

FIA video database~\cite{goh2005cmu} contains video sequences that emulate a passport checking scenario. The video streams are collected from 221 participants under different capture conditions such as pose, illumination and expression in both indoor and outdoor environments. Videos were collected over three sessions where second and third sessions are three months later than the previous one. The participants are present before 6 cameras for about 5 seconds, resulting in total of 18 video sequences per person. 

\begin{figure}[!b]
\begin{center}
        \centering
\includegraphics[width=0.52\linewidth]{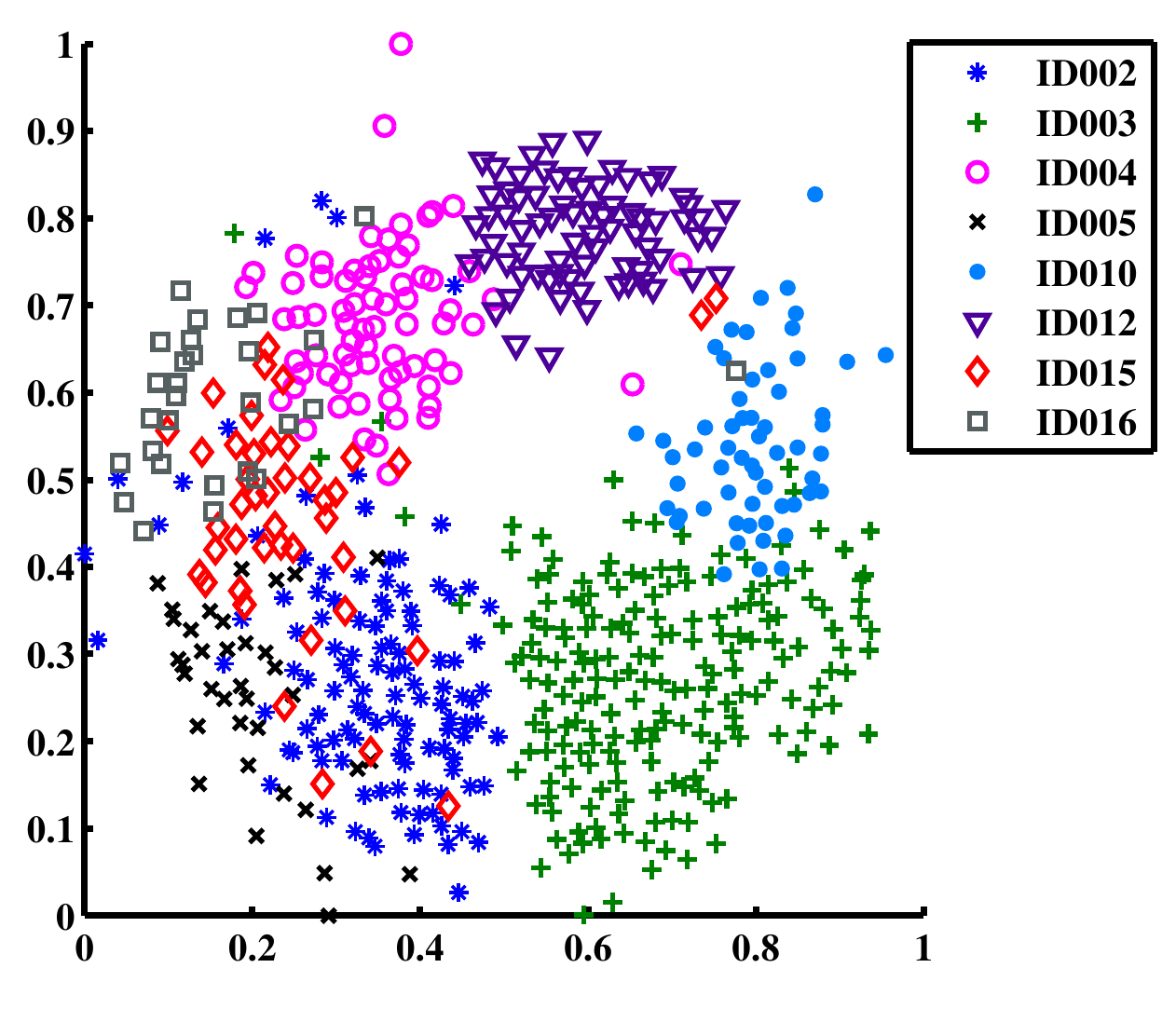}
\includegraphics[width=0.38\linewidth]{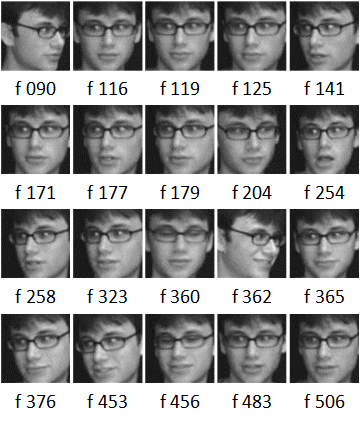}
\end{center}
   \caption{Examples of 2D mapping of LBP feature vectors belonging to 8 individuals using Sammon mapping \protect\cite{sammon1969nonlinear} on the left, and
   examples of 70 $\times$ 70 pixels ROIs in a trajectory captures with camera 3, during session one for ID010 with their frame numbers on the right.}
\label{ROIS}
\label{ROIS}
\end{figure}

For experiments in this paper using FIA dataset, only the faces captured with frontal camera in indoor environment is used for both design and testing.
ROIs are converted to gray-scale and rescaled to $70 \times 70$ pixels using Viola Jones algorithm~\cite{viola2001rapid} from this video. Then, multi-resolution gray-Scale and rotation invariant Local Binary Patterns (LBP)~\cite{ojala2002multiresolution} histograms have been extracted as features. The local image texture for LBP has been characterized with 8 neighbours on a 1 radius circle centred on each pixel. Finally, a feature vector with the length of 59 has been obtained for each ROI. Some examples of ROIs from this data set are presented in Figure \ref{ROIS}.

For experiments with video data, 10 individuals are randomly selected as targets and 90 individuals are randomly selected as non-targets. In each round of experiment, face patterns of one target individual (a trajectory) is considered as the positive class and 100 individuals (including 9 other target individuals and 90 non-target individuals) are selected as the negative class \footnote{The ROIs of each individual in FIA dataset are already grouped to trajectories.}.
ROI patterns from each trajectory are divided into 2 sets for design and testing. The design set is divided to 5 folds, and for each round one fold is considered for validation and remaining 4 folds are considered for training.
Then the roles of design and testing sets is reversed. 
Therefore, for each target individual, three independent sets are collected from these face patterns for training, validation and testing. Each set contains one group of samples from the target individual, 9 groups of samples from the remaining target individuals and 90 groups of samples from non-target individuals. 
Repeating this process for each target individual yields $10 \times 10 = 100$ overall experiments for this dataset.

Two imbalance levels ($\lambda_{\rm train}=1:50$ and 100) and four different imbalance levels $\lambda_{\rm test}=\{1:1, 1:20, 1:50, 1:100\}$ are considered for selecting the training and testing negative class for each positive individual, respectively. This is to evaluate the performance of different classification algorithms when they are trained on different imbalance levels, and to evaluate the robustness of the classification algorithms over varying skew levels during operations. 
When $\lambda_{\rm train}=1:50$, for each positive individual, only $T^-=50$ of 100
other individuals are used as the negative class from the training set that was collected for that positive individual. Therefore, when $\lambda_{\rm test}=1:100$, there are 50 negative individuals in the testing set that were not included in training the classification systems and the skew level of test data is higher than the skew level of training data.
When $\lambda_{\rm test}<1:50$, most of the negative individuals that were used for training do not appear in testing data.
 When $\lambda_{\rm train}=\lambda_{\rm test}=100$, the maximum imbalance level of testing data is the same as the imbalance level of training data. Therefore, all individuals are seen in both training and testing. However, in this case a high level of imbalance exists in both training and testing stages that makes both learning and classification more difficult. It is worth mentioning that in all settings, the skew level of the validation data is selected to be the same as testing data.
\subsubsection{KEEL Collection}

KEEL (Knowledge Extraction based on Evolutionary Learning) tool is an open source software that supports data management and a designer of experiments \cite{alcala2010keel}. The dataset collection in KEEL format contains several datasets for binary classification problem with different number of samples, attributes and imbalance levels. In this paper, the first group of this collection \footnote{(http://www.KEEL.es/dataset.php)} is used for experiments. In this group, the skew level of datasets ranges between 1:9 and 1:129.
The experiments on this collection is done using stratified 2 $\times$ 5 fold cross validation strategy. Therefore, the skew levels of training, validation and testing sets are equal.
\subsection{Experimental Protocol}
For validation, datasets are selected and generated to consider two possible cases.
Synthetic and video datasets are used to evaluate the algorithm when the ideal partitions (or clusters) of negative class are known a priori. The synthetic and video sets are also used for a binary classification problem where no information is available regarding the ideal clusters of data. The KEEL collection datasets are also a case when the data is not partitioned a priori.

We use SVM with RBF kernel~\cite{chang2011libsvm} as the base classifier where $K(\mathbf x', \mathbf x'') = \exp\{\rfrac{-\|\mathbf{x}'-\mathbf{x}''\|^2}{2 \kappa^2}\}$. The kernel parameter $\kappa$ is set as the average of the mean minimum distance between any two training samples and the scatter radius of the training samples in the input space~\cite{li2008adaboost}. The scatter radius is calculated by selecting the maximum distance between the training samples and a point corresponding to the mean of training samples. We used the LibSVM implementation of~\cite{chang2011libsvm}.

A brief description of the implemented ensembles, their variants and the abbreviations used for them are shown in Table \ref{tacr1}. The last column of the table shows the datasets that are used for experiments on these classification systems.
The abbreviations assigned to these ensembles are selected based on their sampling techniques and loss factor.

The baseline sampling techniques include Ada (resampling in AdaBoost), SMT (SMOTE in SMOTEBoost), RUS (random under-sampling in RUSBoost), RB (random balance in RB-Boost). For PBoost four partitioning techniques are used for under-sampling the negative class to evaluate the effect of the partitioning technique on the performance of PBoost ensemble: random under sampling without replacement (PRUS) and cluster under-sampling (PCUS) are used as general partitioning techniques for PBoost disregarding the data structure, whether or not the negative class is partitioned a priori. For PCUS, kernel $k$-means is used for clustering negative samples. To select $k$, it is varied over a range of possible values and the value of Dunn index~\cite{dunn1973fuzzy} is calculated for each case using a validation set. Finally, the optimal $k$, is selected when Dunn index takes its maximum value. Two cases are considered for PBoost in which the partitions of the negative class are known a priori. The ideal cluster under-sampling (PCUS$_i$) with synthetic datasets and trajectory under-sampling (PTUS) with video dataset.
 
The loss factor is calculated in 2 ways based on: the traditional technique i.e. weighted accuracy, and the F-measure. To indicate the use of proposed loss factor in the Boosting ensembles in Table \ref{tacr1}, the abbreviation is followed by -F. For the use of proposed loss factor calculation with the F-measure, $\beta$ is set as 2 in all experiments because $\beta \geq 1$ is more suitable for imbalanced data classification when the positive class is the minority class.
An experiment is done to evaluate the performance of Boosting ensembles with different values of $\beta$.

\begin{table}[!htb]
\small
\centering
\caption{Baseline Boosting ensembles and their variants.}
\label{tacr1}
\resizebox{1\textwidth}{!}{%
\begin{tabular}{|l|l|l|l|l|}\hline
\textbf{Abbreviation}  & \textbf{Sampling method} & \textbf{Boosting Ensemble}  & \textbf{Loss factor}& \textbf{Data}\\\hline\hline
   \textbf{Ada:}      & Resampling with replacement &  AdaBoost~\cite{freund1996experiments}   &  Weighted accuracy & Synthetic, Video, KEEL\\
   \textbf{Ada-F:}   & Resampling with replacement &  Modified AdaBoost~\cite{freund1996experiments}   &  Proposed F-measure& Synthetic, Video\\
    \textbf{SMT:}     & Synthetic minority over-sampling technique (SMOTE)  &  SMOTEBoost~\cite{chawla2003smoteboost}   &  Weighted accuracy& Synthetic, Video, KEEL\\
     \textbf{SMT-F}  & Synthetic minority over-sampling technique(SMOTE) &   Modified SMOTEBoost~\cite{chawla2003smoteboost}   &  Proposed F-measure& Synthetic, Video\\
     \textbf{RUS:}    & Random under-sampling with replacement (RUS) & RUSBoost~\cite{seiffert2010rusboost}  & Weighted accuracy& Synthetic, Video, KEEL\\
     \textbf{RUS-F:}  & Random under-sampling with replacement (RUS) &   Modified RUSBoost~\cite{seiffert2010rusboost}  & Proposed F-measure& Synthetic, Video\\
     \textbf{RB: }       & Combination of up-sampling (SMOTE) and under-sampling (RUS)&  RB-Boost~\cite{diez2015random} & Weighted accuracy& Synthetic, Video, KEEL\\
     \textbf{RB-F: }    & Combination of up-sampling (SMOTE) and under-sampling (RUS) &   Modified RB-Boost~\cite{diez2015random} & Proposed F-measure& Synthetic, Video\\
        \textbf{PRUS:}      & Random under-sampling without replacement (RUSwR)&  Progressive Boosting   &  Weighted accuracy& Synthetic, Video\\
   \textbf{PRUS-F:}   & Random under-sampling without replacement (RUSwR)&  Progressive Boosting   &  Proposed F-measure& Synthetic, Video, KEEL\\
     \textbf{PCUS:}    & Selecting clusters found by k-means & Progressive Boosting  & Weighted accuracy& Synthetic, Video\\
     \textbf{PCUS-F:}  & Selecting clusters found by k-means & Progressive Boosting  & Proposed F-measure& Synthetic, Video, KEEL\\
    \textbf{PCUS$_i$:}   & Selecting ideal clusters generated in synthetic dataset  &  Progressive Boosting   &  Weighted accuracy& Synthetic\\
     \textbf{PCUS$_i$-F:}   & Selecting ideal clusters generated in synthetic dataset   &  Progressive Boosting  &  Proposed F-measure& Synthetic\\
     \textbf{PTUS: }       & Selecting trajectories in video dataset &  Progressive Boosting & Weighted accuracy & Video\\
     \textbf{PTUS-F: }    & Selecting trajectories in video dataset  &  Progressive Boosting & Proposed F-measure & Video\\\hline
 \end{tabular}}
\vspace{-0.2cm}
\end{table} 

%\begin{table}[!htb]
%\small
%\centering
%\caption{Acronyms of variants of PBoost.}
%\label{tacr2}
%\resizebox{\textwidth}{!}{%
%\begin{tabular}{ll|ll|ll|ll }\hline\hline
%   \textbf{PRUS:}      &  PRUSBoost      &   \textbf{PCUS:}     &  PCUSBoost     & \textbf{PCUS$_i$:}      & PCUSBoost-ideal      & \textbf{PTUS: }    &  PTUSBoost\\
%   \textbf{PRUS-F:}  &  PRUSBoost -F  &   \textbf{PCUS-F:}  &  PCUSBoost-F  & \textbf{PCUS$_i$-F:}   & PCUSBoost-ideal-F   & \textbf{PTUS-F:} &  PTUSBoost-F\\\hline\hline
% \end{tabular}}
%\end{table} 

 In the experiments with synthetic and video data sets, two different imbalance levels are used for training and four different imbalance levels are used for testing. This is to evaluate the sensitivity of classification systems to the level of imbalance during training and their robustness to possible variations in skew level during operations. In experiments with synthetic data, the overlap level between positive and negative classes are also varied because the issue of imbalance is related to the level of overlap between classes~\cite{lopez2013insight}.
 
In experiments with synthetic and video datasets, the size of all Boosting ensembles is set equal to the maximum imbalance level of the data, except from PCUS. 
The reason for this setting is that the number of ideal clusters and the number of trajectories are both known and equal to the level of skew. In addition, based on a preliminary experiment with D$_2$ on baseline ensembles in Figure~\ref{E}, it is observed that the size of these ensembles does not have a significant impact on their performance. The performance of these ensembles vary in terms of F$_2$-measure as the ensemble size grows. However, their global performance in terms of AUPR do not change significantly. 
For PCUS, the size of ensemble is selected equal to the optimal \textit{k} obtained using Dunn index.

\begin{figure}[!tb]
\begin{center}
     \begin{subfigure}[b]{0.28\textwidth}
                \includegraphics[width=1\textwidth]{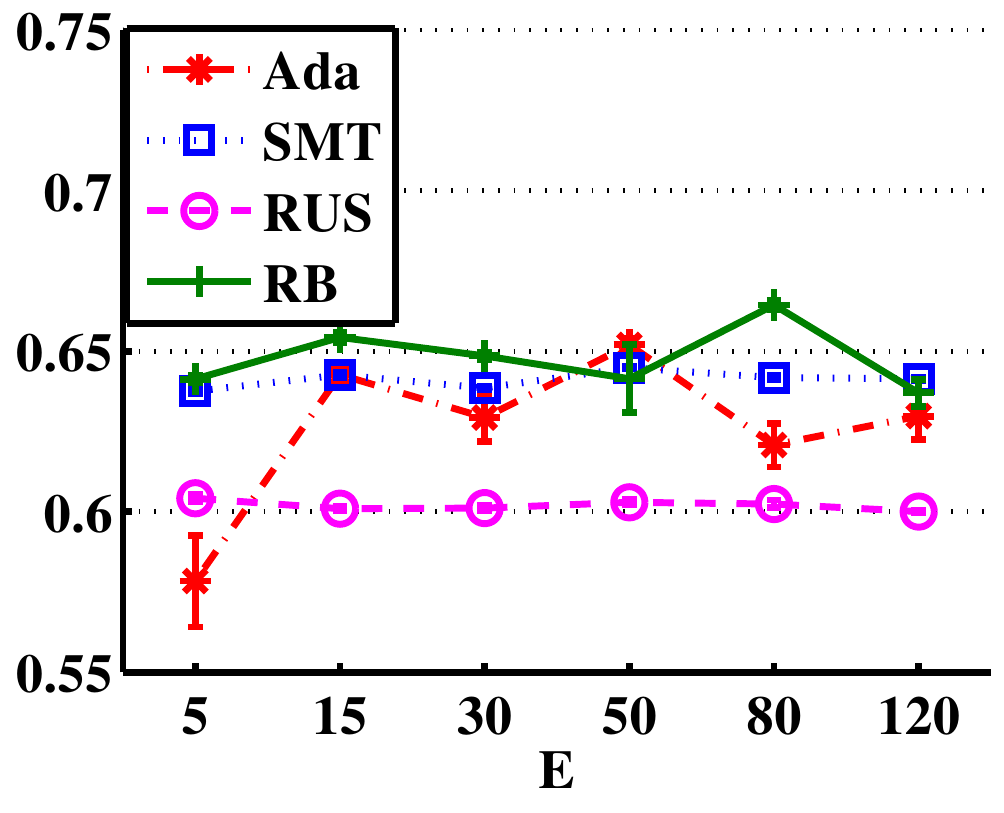}
                \caption{AUPR.}
        \end{subfigure}
     \begin{subfigure}[b]{0.28\textwidth}
                \includegraphics[width=1\textwidth]{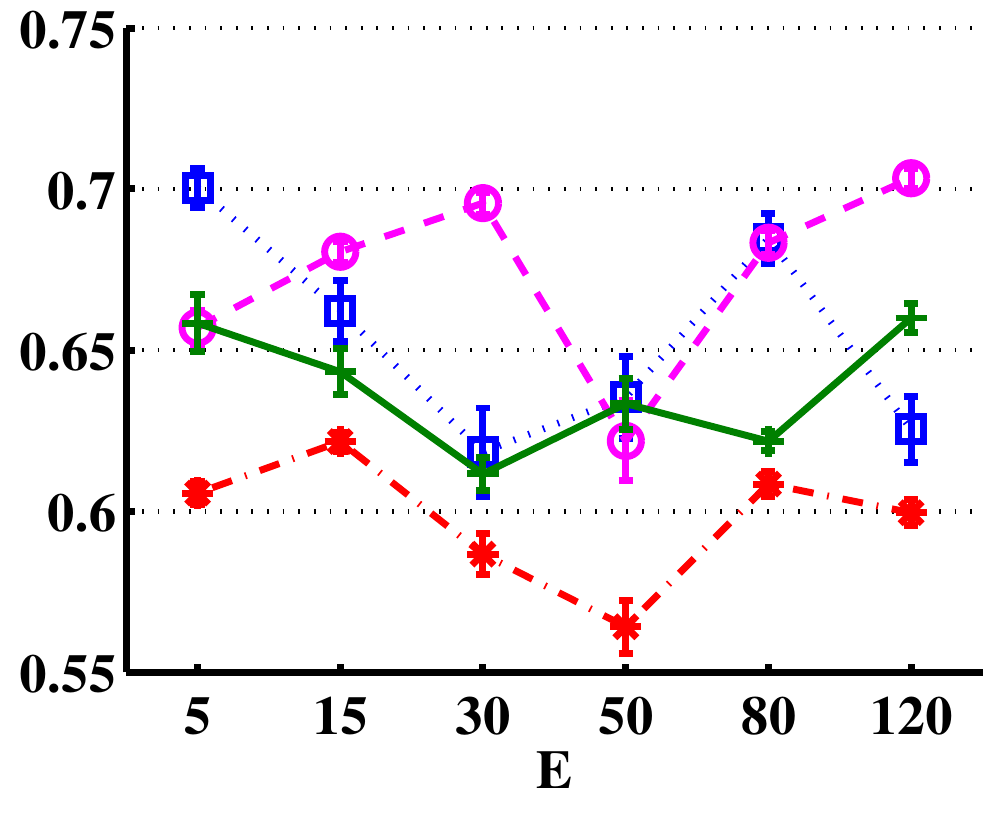}
                \caption{F$_2$-measure.}
        \end{subfigure}
\caption{Performance of baseline Boosting ensembles for different values of $E$ on $D_2$ with $\lambda_{\rm test}=1:100$.}
\label{E}
\label{E}
\end{center}
\end{figure}
 
\subsection{Evaluation Protocol}
Global performance evaluation curves such as ROC and PR, show the trade off between two metrics for different operational settings. For classifiers that output scores or probability estimates, this setting is usually the choice of decision threshold. Area under the curve, shows the global performance of the classifier over a range of possible decision thresholds, where local evaluation metric such as F-measure show the performance for a specific decision threshold. 

Therefore, when different classifiers are compared in terms of local metrics, the choice of the decision threshold becomes important. The decision threshold may be set to a fixed optimal value without considering the operating condition or based on operating conditions: the cost proportions or skew levels \cite{hernandez2012unified}. The performance metrics that can be maximized to set the decision threshold are accuracy, Brier score, AUC, expected cost and F-measure \cite{hernandez2012unified, lipton2014optimal}. To this aim, the classifiers are tested on a set of data called validation datasets that are independent from training and testing data.

As explained in Section 2.4, AUPR and F-measure are more suitable metrics to compare the performance of the classification systems when data is imbalanced. Therefore, in the experiments in this paper, AUPR is used to compare the performance of Boosting ensembles globally and F-measure with $\beta=2$ is used to compare the classifiers for a specific operating condition. AUPR shows the average value of precision for different values of recall (or TPR), and F$_2$-measure shows the harmonic mean of precision and recall when a higher importance is given to recall.

\begin{figure}[!t]
\begin{center}
        \centering
\includegraphics[width=0.24\linewidth]{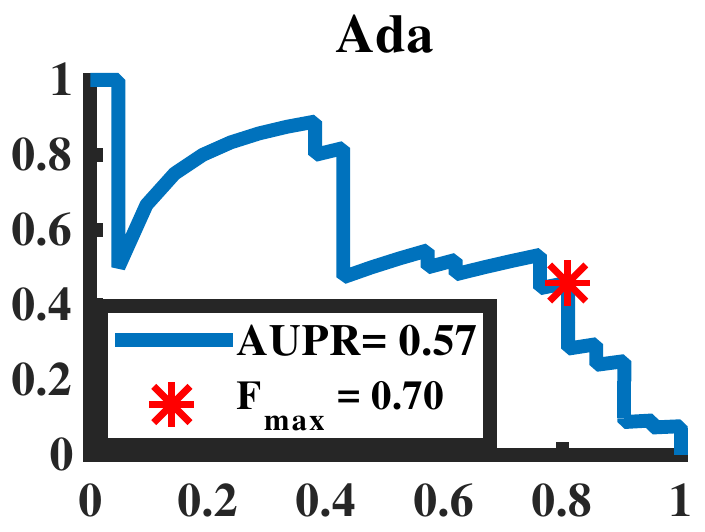}
\includegraphics[width=0.24\linewidth]{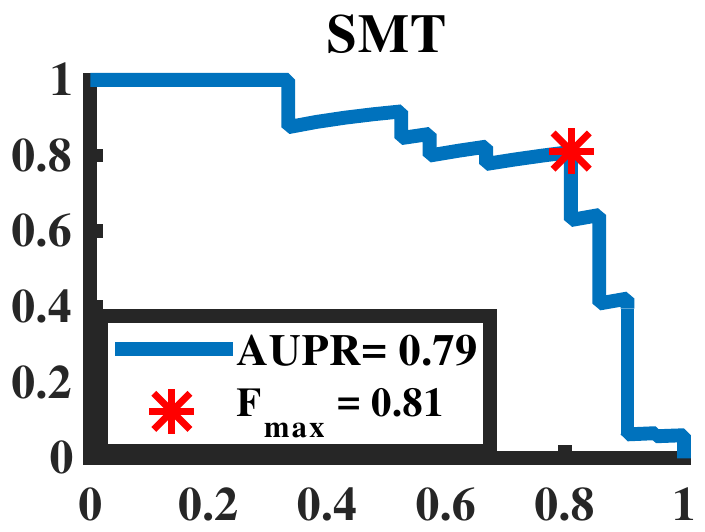}
\includegraphics[width=0.24\linewidth]{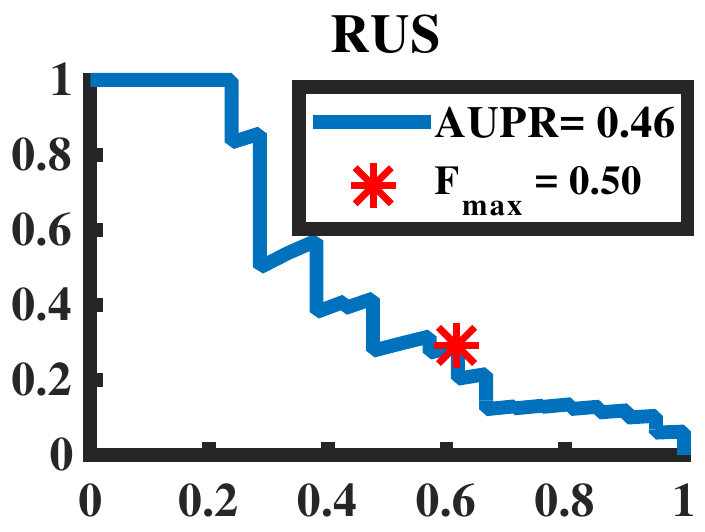}
\includegraphics[width=0.24\linewidth]{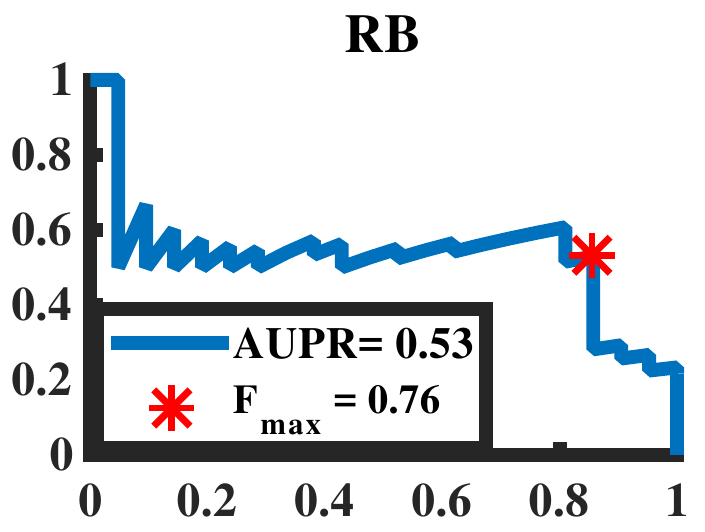}
\end{center}
   \caption{PR curve of baseline Boosting ensembles on validation data and finding the optimal threshold.}
\label{val}
\label{val}
\end{figure}
\begin{figure}[!t]
\begin{center}
        \centering
\includegraphics[width=0.24\linewidth]{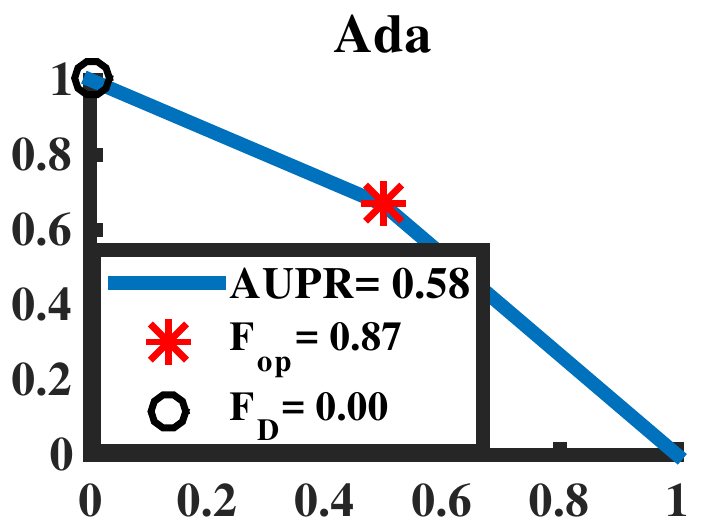}
\includegraphics[width=0.24\linewidth]{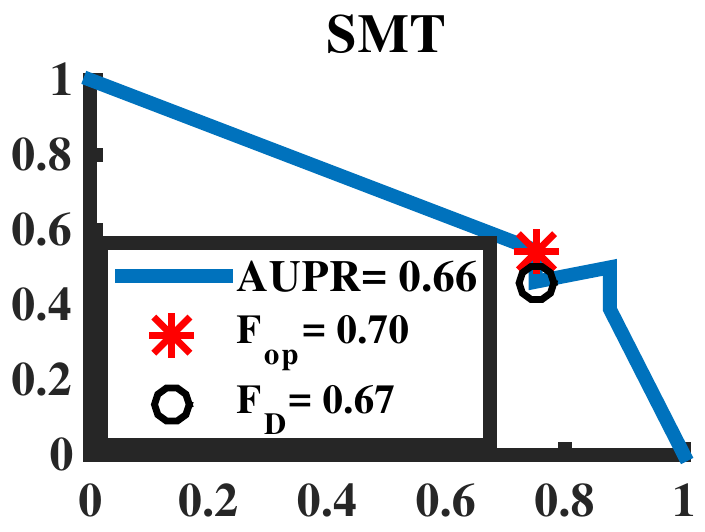}
\includegraphics[width=0.24\linewidth]{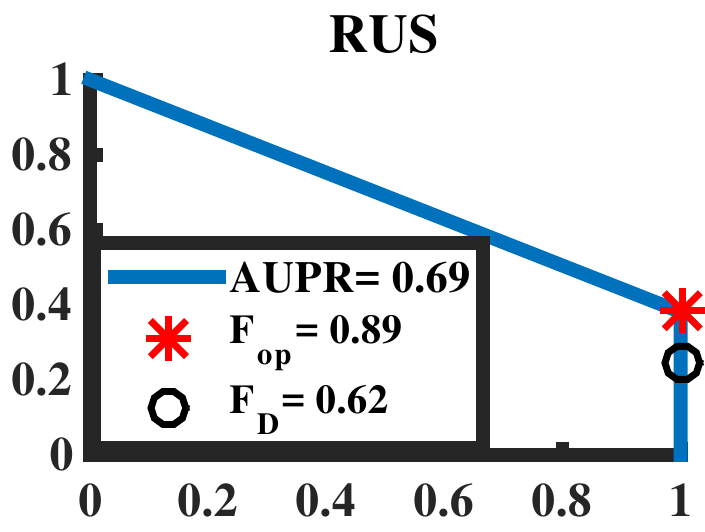}
\includegraphics[width=0.24\linewidth]{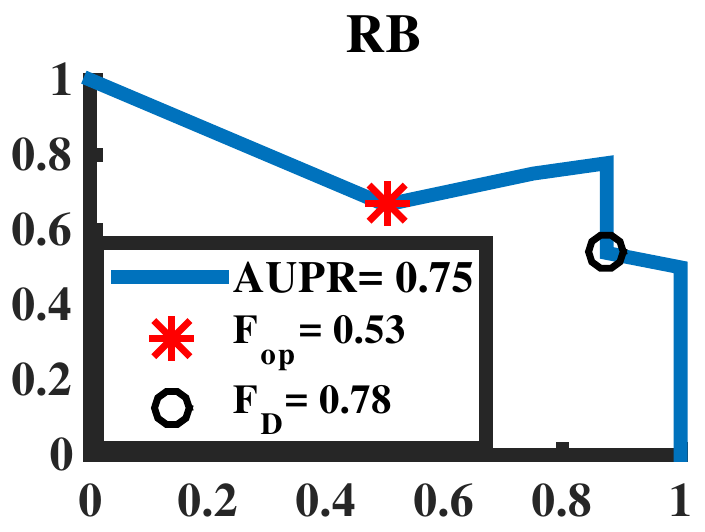}
\end{center}
   \caption{PR curve of baseline Boosting ensembles on test data using the optimal threshold obtained with validation data.}
\label{val_op}
\label{val_op}
\end{figure}

The value of AUPR and F$_2$-measure is averaged over 10 replications obtained by 2 $\times$ 5-fold cross validation. 
In our experiments, the decision threshold to obtain the F$_2$-measure is set to the value that maximizes the value of F$_2$-measure on the validation data for comparing the performance of different classification algorithms.
An example is shown in Figure \ref{val}, the PR curve of an experiment on abalone9-18 dataset of KEEL collection on the validation data. 
In Figure~\ref{val_op}, the ensembles are tested on a different test set and $F_{\rm op}$ shows the value of F-measure when the optimal threshold is selected using the validation step described.
 $F_{\rm D}$ is the value of F-measure when the combination function in Boosting ensembles is majority voting and the decisions of base classifiers are combined. It is observed that $F_{\rm op}$ and $F_{\rm D}$ may differ significantly and in most cases $F_{\rm op} > F_{\rm D}$. 

In our experiments, the performance of the proposed PBoost ensemble is also compared to state of the art Boosting ensembles in terms of computational complexity.
Time complexity for SVM training depends on several factors including the number of training samples, the learning (optimization) algorithm and the number of features. The computational complexity of SVM implemented in LibSVM is evaluated in~\cite{chang2011libsvm}, as $O(n_{\rm tr}d)$ per iteration $I$,
where $n_{\rm tr}$ is the training set size, and $d$ is the number of features.
The authors state that ``the number of iterations $p$ may be higher than linear to the number of training data". Therefore, the complexity is $O(n_{\rm tr}^p \cdot d)$ for some $p>2$. This means that, time complexity for SVM training is not proportional to, but increases more than linearly with respect to the training set size. 

In the proposed PRUS and baseline Boosting ensembles, $p$ is unknown and $d$ is identical in all algorithms.
Each iteration of Boosting ensembles includes a validation step that should be added to training complexity to obtain the overall time complexity of learning process. Time complexity of the validation step $O(n_{\rm SV} \cdot n_{\rm val})$, depends on the number of validation samples $n_{\rm val}$ and the number of support vectors $n_{\rm SV}$ obtained from training each SVM.
The reason is that, when an RBF SVM with $n_{\rm SV}$ support vectors is tested on a probe sample $\textbf{x}$, the value of $K(\mathbf x, \mathbf {\rm SV_j}) = \exp\{\rfrac{-\|\mathbf{x}-\mathbf{\rm SV_j}\|^2}{2 \kappa^2}\}$ is accumulated for all support vectors ($j=1,\dotsc,n_{\rm SV}$) and the sign of the resulting quantity determines the decision.

Table~\ref{tcmpx} shows the number of samples to train and validate these Boosting ensembles of the size $E$.
The number of validation samples in baseline Boosting ensembles is the same and equal to the overall number of training samples. However, the overall number of samples used for validation in PRUSBoost is calculated as:
\begin{eqnarray}
\sum^E_{e=1}(M^++\sum^e_{f=1}N_f)& = & EM^++\sum^E_{e=1}(E-(e-1))N_e, \\
& = & EM^++E^2-\sum^E_{e=1}eN_e+M^-, \\
& = & EM^++M^-+E^2-\sum^E_{e=1}eN_e.
\end{eqnarray}
This value is less than $E(M^++M^-)$ that is the total number of validation samples in the sate of the art Boosting ensembles.
Table~\ref{tcmpx} shows that the total number of training and validation samples in PRUS ensemble is the smallest one.

\begin{table}  [!t]
\small
%\setstretch{1.2}
\centering
\renewcommand\multirowsetup{\centering}
\caption{Number of training and validation samples.}
\label{tcmpx}
\resizebox{0.9\textwidth}{!}{%
\begin{tabular}{l| c c| c c }\hline \hline
    \textbf{Ensemble} & \textbf{$n_{\rm tr}$ in iteration $e$} & \textbf{Total $n_{\rm tr}$} &\textbf{$n_{\rm val}$ in iteration $e$} & \textbf{Total $n_{\rm val}$}\\\hline
    \textbf{Ada} &$M^++M^-$& $E(M^++M^-)$ & $M^++M^-$ & $E(M^++M^-)$ \\
    \textbf{SMT} & $2M^-$ & $2EM^-$ & $M^++M^-$ & $E(M^++M^-)$ \\
    \textbf{RUS} & $2M^+$ & $2EM^+$ & $M^++M^-$ & $E(M^++M^-)$\\
    \textbf{RB} & $M^++M^-$& $E(M^++M^-)$ &$M^++M^-$ &$E(M^++M^-)$ \\
    \textbf{PRUS} &$M^++N_e$& $EM^++M^-$ & $M^++\sum^e_{f=1}N_f$& $EM^++M^-+E^2-\sum^E_{e=1}eN_e$ \\\hline\hline
    \end{tabular}}
\end{table} 
%$\\$

\section{Results and Discussion}
The performance of the proposed and state of the art ensemble learning methods are analysed for synthetic and video data in 4 parts: (1) accuracy and robustness over different levels of overlap and imbalance between design and test data and of using the proposed loss factor; (2) the performance of RUSBoost with and without progressive partitioning; (3) the combined impact of progressive partitioning and proposed loss factor; (4) the computation complexity during design and testing.

\subsection{Results of Experiments with Synthetic Data}

\subsubsection{Impact of proposed loss factor} 
The performance of the baseline Boosting ensembles: AdaBoost
, SMOTEBoost, RUSBoost, and RB-Boost %\footnote{RBBoost failed a few times to be generated. Nevertheless, only successful attempts are considered.} 
are compared in Table~\ref{baseF} for different settings.
In addition, F$_\beta$ is used to optimize loss factor calculation in these ensembles.

\begin{table}  [!b]
 \centering
  \caption{Average of F$_2$-measure and AUPR performance of baseline techniques with and without proposed loss factor on synthetic data over different levels of skew and overlap in test data.}
\label{baseF}
  \resizebox{1\textwidth}{!}{
\begin{tabular}{l c|| c|c|c|c|| c|c|c|c|| c|c|c|c } \hline \hline
\multirow{2}[2]{*}{\textbf{Ensembles }}&\makecell{\textbf{Train} \\ \textbf{Data}}& \multicolumn{4}{c||}{\boldmath$D_1$  ($\boldsymbol{\lambda_{\rm train}}=$1:50, $\boldsymbol{\delta}=0.2$) } & \multicolumn{4}{c||}{\boldmath$D_2$  ($\boldsymbol{\lambda_{\rm train}}=$1:50, $\boldsymbol{\delta}=0.1$)} & \multicolumn{4}{c}{\boldmath\boldmath$D_3$  ($\boldsymbol{\lambda_{\rm train}}=$1:20, $\boldsymbol{\delta}=0.2$) } \\ \cline{2-14}
&$\boldsymbol{\lambda_{\rm test}}$& \textbf{1:1} & \textbf{1:20} & \textbf{1:50} & \textbf{1:100} & \textbf{1:1} & \textbf{1:20} & \textbf{1:50} & \textbf{1:100} &\textbf{1:1} & \textbf{1:20} & \textbf{1:50} & \textbf{1:100}  \\ \hline \hline
\multicolumn{14}{c}{\textbf{F$_2$-measure}}\\\hline \hline
\textbf{Ada} && \makecell{0.98 \\ {\small $\pm$ 0.00}} & \makecell{0.94 \\ {\small $\pm$ 0.00}} & \makecell{0.91 \\ {\small $\pm$ 0.00}} & \makecell{0.86 \\ {\small $\pm$ 0.00}} & \makecell{0.94 \\ {\small $\pm$ 0.01}} & \makecell{0.85 \\ {\small $\pm$ 0.01}} & \makecell{0.83 \\ {\small $\pm$ 0.01}} & \makecell{0.51 \\ {\small $\pm$ 0.01}} & \makecell{0.89 \\ {\small $\pm$ 0.01}} & \makecell{0.88 \\ {\small $\pm$ 0.01}} & \makecell{0.88 \\ {\small $\pm$ 0.01}} & \makecell{0.46 \\ {\small $\pm$ 0.00}}\\ \hline
\textbf{Ada-F} && \makecell{0.96 \\ {\small $\pm$ 0.00}} & \makecell{0.93 \\ {\small $\pm$ 0.00}} & \makecell{0.90 \\ {\small $\pm$ 0.00}} & \makecell{\textbf{0.87} \\ {\small $\pm$ 0.00}} & \makecell{0.94 \\ {\small $\pm$ 0.01}} & \makecell{\textbf{0.88} \\ {\small $\pm$ 0.01}} & \makecell{\textbf{0.85} \\ {\small $\pm$ 0.01}} & \makecell{\textbf{0.55} \\ {\small $\pm$ 0.01}} & \makecell{\textbf{0.92} \\ {\small $\pm$ 0.01}} & \makecell{\textbf{0.91} \\ {\small $\pm$ 0.01}} & \makecell{\textbf{0.91} \\ {\small $\pm$ 0.01}} & \makecell{\textbf{0.48} \\ {\small $\pm$ 0.00}}\\ \hline\hline
\textbf{SMT} && \makecell{0.96 \\ {\small $\pm$ 0.01}} & \makecell{0.90 \\ {\small $\pm$ 0.00}} & \makecell{0.84 \\ {\small $\pm$ 0.00}} & \makecell{0.81 \\ {\small $\pm$ 0.00}} & \makecell{0.93 \\ {\small $\pm$ 0.01}} & \makecell{0.86 \\ {\small $\pm$ 0.01}} & \makecell{0.84 \\ {\small $\pm$ 0.01}} & \makecell{0.58 \\ {\small $\pm$ 0.01}} & \makecell{0.92 \\ {\small $\pm$ 0.01}} & \makecell{0.92 \\ {\small $\pm$ 0.01}} & \makecell{0.92 \\ {\small $\pm$ 0.01}} & \makecell{0.56 \\ {\small $\pm$ 0.01}}\\ \hline
\textbf{SMT-F} && \makecell{0.95 \\ {\small $\pm$ 0.01}} & \makecell{0.89 \\ {\small $\pm$ 0.00}} & \makecell{0.84 \\ {\small $\pm$ 0.00}} & \makecell{0.81 \\ {\small $\pm$ 0.00}} & \makecell{0.95 \\ {\small $\pm$ 0.01}} & \makecell{0.85 \\ {\small $\pm$ 0.00}} & \makecell{0.82 \\ {\small $\pm$ 0.00}} & \makecell{0.55 \\ {\small $\pm$ 0.01}} & \makecell{\textbf{0.97} \\ {\small $\pm$ 0.00}} & \makecell{\textbf{0.96} \\ {\small $\pm$ 0.00}} & \makecell{\textbf{0.96} \\ {\small $\pm$ 0.00}} & \makecell{\textbf{0.57} \\ {\small $\pm$ 0.00}}\\ \hline\hline
\textbf{RUS} && \makecell{0.93 \\ {\small $\pm$ 0.00}} & \makecell{0.89 \\ {\small $\pm$ 0.00}} & \makecell{0.86 \\ {\small $\pm$ 0.00}} & \makecell{0.83 \\ {\small $\pm$ 0.00}} & \makecell{0.85 \\ {\small $\pm$ 0.01}} & \makecell{0.82 \\ {\small $\pm$ 0.01}} & \makecell{0.81 \\ {\small $\pm$ 0.01}} & \makecell{0.64 \\ {\small $\pm$ 0.00}} & \makecell{0.91 \\ {\small $\pm$ 0.01}} & \makecell{0.91 \\ {\small $\pm$ 0.01}} & \makecell{0.91 \\ {\small $\pm$ 0.01}} & \makecell{0.81 \\ {\small $\pm$ 0.01}}\\ \hline
\textbf{RUS-F} && \makecell{\textbf{0.95} \\ {\small $\pm$ 0.01}} & \makecell{\textbf{0.92} \\ {\small $\pm$ 0.01}} & \makecell{\textbf{0.88}\\ {\small $\pm$ 0.00}} & \makecell{\textbf{0.85} \\ {\small $\pm$ 0.00}} & \makecell{\textbf{0.90} \\ {\small $\pm$ 0.01}} & \makecell{\textbf{0.88} \\ {\small $\pm$ 0.01}} & \makecell{\textbf{0.84} \\ {\small $\pm$ 0.01}} & \makecell{\textbf{0.68} \\ {\small $\pm$ 0.00}} & \makecell{0.90 \\ {\small $\pm$ 0.01}} & \makecell{0.90 \\ {\small $\pm$ 0.01}} & \makecell{0.90 \\ {\small $\pm$ 0.01}} & \makecell{\textbf{0.82} \\ {\small $\pm$ 0.01}}\\ \hline\hline
\textbf{RB} && \makecell{0.99 \\ {\small $\pm$ 0.00}} & \makecell{0.94 \\ {\small $\pm$ 0.00}} & \makecell{0.89 \\ {\small $\pm$ 0.00}} & \makecell{0.86 \\ {\small $\pm$ 0.00}} & \makecell{0.99 \\ {\small $\pm$ 0.00}} & \makecell{0.91 \\ {\small $\pm$ 0.00}} & \makecell{0.91 \\ {\small $\pm$ 0.00}} & \makecell{0.61 \\ {\small $\pm$ 0.00}} & \makecell{0.93 \\ {\small $\pm$ 0.01}} & \makecell{0.93 \\ {\small $\pm$ 0.01}} & \makecell{0.93 \\ {\small $\pm$ 0.01}} & \makecell{0.50 \\ {\small $\pm$ 0.01}}\\ \hline
\textbf{RB-F} && \makecell{0.98 \\ {\small $\pm$ 0.00}} & \makecell{\textbf{0.95} \\ {\small $\pm$ 0.00}} & \makecell{\textbf{0.90} \\ {\small $\pm$ 0.00}} & \makecell{\textbf{0.87} \\ {\small $\pm$ 0.00}} & \makecell{0.97 \\ {\small $\pm$ 0.01}} & \makecell{0.90 \\ {\small $\pm$ 0.01}} & \makecell{0.88 \\ {\small $\pm$ 0.01}} & \makecell{0.59 \\ {\small $\pm$ 0.00}} & \makecell{\textbf{0.95} \\ {\small $\pm$ 0.01}} & \makecell{\textbf{0.94} \\ {\small $\pm$ 0.01}} & \makecell{\textbf{0.94} \\ {\small $\pm$ 0.01}} & \makecell{\textbf{0.57} \\ {\small $\pm$ 0.01}}\\ \hline
 \hline
%\end{tabular}}
%\end{table}
%
%
%\begin{table}  [htb]
% \centering
%  \caption{Average of AUPR performance of baseline techniques with and without proposed loss factor on synthetic data over different levels of skew and overlap in test data.}
%\label{baseAU}
%  \resizebox{1\textwidth}{!}{
%\begin{tabular}{l c|| c|c|c|c|| c|c|c|c|| c|c|c|c } \hline \hline
%\multirow{2}[2]{*}{\textbf{Ensembles }}&\makecell{\textbf{Train} \\ \textbf{Data}}& \multicolumn{4}{c||}{\boldmath$D_1$  ($\boldsymbol{\lambda_{\rm train}}=$1:50, $\boldsymbol{\delta}=0.2$) } & \multicolumn{4}{c||}{\boldmath$D_2$  ($\boldsymbol{\lambda_{\rm train}}=$1:50, $\boldsymbol{\delta}=0.1$)} & \multicolumn{4}{c}{\boldmath\boldmath$D_3$  ($\boldsymbol{\lambda_{\rm train}}=$1:20, $\boldsymbol{\delta}=0.2$) } \\ \cline{2-14}
%&$\boldsymbol{\lambda_{\rm test}}$& \textbf{1:1} & \textbf{1:20} & \textbf{1:50} & \textbf{1:100} & \textbf{1:1} & \textbf{1:20} & \textbf{1:50} & \textbf{1:100} &\textbf{1:1} & \textbf{1:20} & \textbf{1:50} & \textbf{1:100}  \\ \hline \hline
\multicolumn{14}{c}{\textbf{AUPR}}\\\hline \hline
\textbf{Ada} && \makecell{0.99 \\ {\small $\pm$ 0.00}} & \makecell{0.97 \\ {\small $\pm$ 0.00}} & \makecell{0.93 \\ {\small $\pm$ 0.00}} & \makecell{0.85 \\ {\small $\pm$ 0.00}} & \makecell{1.00 \\ {\small $\pm$ 0.00}} & \makecell{0.88 \\ {\small $\pm$ 0.00}} & \makecell{0.88 \\ {\small $\pm$ 0.00}} & \makecell{0.60 \\ {\small $\pm$ 0.00}} & \makecell{1.00 \\ {\small $\pm$ 0.00}} & \makecell{1.00 \\ {\small $\pm$ 0.00}} & \makecell{1.00 \\ {\small $\pm$ 0.00}} & \makecell{0.58 \\ {\small $\pm$ 0.00}}\\ \hline
\textbf{Ada-F} && \makecell{0.98 \\ {\small $\pm$ 0.00}} & \makecell{0.96 \\ {\small $\pm$ 0.00}} & \makecell{0.93 \\ {\small $\pm$ 0.00}} & \makecell{0.85 \\ {\small $\pm$ 0.00}} & \makecell{1.00 \\ {\small $\pm$ 0.00}} & \makecell{0.88 \\ {\small $\pm$ 0.00}} & \makecell{0.88 \\ {\small $\pm$ 0.00}} & \makecell{0.60 \\ {\small $\pm$ 0.00}} & \makecell{1.00 \\ {\small $\pm$ 0.00}} & \makecell{1.00 \\ {\small $\pm$ 0.00}} & \makecell{1.00 \\ {\small $\pm$ 0.00}} & \makecell{\textbf{0.59} \\ {\small $\pm$ 0.00}}\\ \hline\hline
\textbf{SMT} && \makecell{1.00 \\ {\small $\pm$ 0.00}} & \makecell{0.90 \\ {\small $\pm$ 0.00}} & \makecell{0.82 \\ {\small $\pm$ 0.00}} & \makecell{0.78 \\ {\small $\pm$ 0.00}} & \makecell{1.00 \\ {\small $\pm$ 0.00}} & \makecell{0.83 \\ {\small $\pm$ 0.00}} & \makecell{0.81 \\ {\small $\pm$ 0.00}} & \makecell{0.60 \\ {\small $\pm$ 0.00}} & \makecell{1.00 \\ {\small $\pm$ 0.00}} & \makecell{0.98 \\ {\small $\pm$ 0.00}} & \makecell{0.97 \\ {\small $\pm$ 0.00}} & \makecell{0.58 \\ {\small $\pm$ 0.00}}\\ \hline
\textbf{SMT-F} && \makecell{1.00 \\ {\small $\pm$ 0.00}} & \makecell{0.90 \\ {\small $\pm$ 0.00}} & \makecell{\textbf{0.83} \\ {\small $\pm$ 0.00}} & \makecell{0.78 \\ {\small $\pm$ 0.00}} & \makecell{1.00 \\ {\small $\pm$ 0.00}} & \makecell{0.83 \\ {\small $\pm$ 0.00}} & \makecell{0.81 \\ {\small $\pm$ 0.00}} & \makecell{0.60 \\ {\small $\pm$ 0.00}} & \makecell{1.00 \\ {\small $\pm$ 0.00}} & \makecell{0.98 \\ {\small $\pm$ 0.00}} & \makecell{0.97 \\ {\small $\pm$ 0.00}} & \makecell{0.58 \\ {\small $\pm$ 0.00}}\\ \hline\hline
\textbf{RUS} && \makecell{1.00 \\ {\small $\pm$ 0.00}} & \makecell{0.66 \\ {\small $\pm$ 0.00}} & \makecell{0.60 \\ {\small $\pm$ 0.00}} & \makecell{0.57 \\ {\small $\pm$ 0.00}} & \makecell{1.00 \\ {\small $\pm$ 0.00}} & \makecell{0.74 \\ {\small $\pm$ 0.00}} & \makecell{0.66 \\ {\small $\pm$ 0.00}} & \makecell{0.55 \\ {\small $\pm$ 0.00}} & \makecell{1.00 \\ {\small $\pm$ 0.00}} & \makecell{0.86 \\ {\small $\pm$ 0.01}} & \makecell{0.81 \\ {\small $\pm$ 0.01}} & \makecell{0.57 \\ {\small $\pm$ 0.00}}\\ \hline
\textbf{RUS-F} && \makecell{1.00 \\ {\small $\pm$ 0.00}} & \makecell{\textbf{0.77} \\ {\small $\pm$ 0.01}} & \makecell{\textbf{0.70} \\ {\small $\pm$ 0.01}} & \makecell{\textbf{0.66} \\ {\small $\pm$ 0.01}} & \makecell{1.00 \\ {\small $\pm$ 0.00}} & \makecell{0.74 \\ {\small $\pm$ 0.00}} & \makecell{\textbf{0.67} \\ {\small $\pm$ 0.00}} & \makecell{0.55 \\ {\small $\pm$ 0.00}} & \makecell{1.00 \\ {\small $\pm$ 0.00}} & \makecell{\textbf{0.88} \\ {\small $\pm$ 0.01}} & \makecell{\textbf{0.84} \\ {\small $\pm$ 0.01}} & \makecell{0.57 \\ {\small $\pm$ 0.00}}\\ \hline\hline
\textbf{RB} && \makecell{1.00 \\ {\small $\pm$ 0.00}} & \makecell{0.97 \\ {\small $\pm$ 0.00}} & \makecell{0.93 \\ {\small $\pm$ 0.00}} & \makecell{0.87 \\ {\small $\pm$ 0.00}} & \makecell{1.00 \\ {\small $\pm$ 0.00}} & \makecell{0.88 \\ {\small $\pm$ 0.00}} & \makecell{0.88 \\ {\small $\pm$ 0.00}} & \makecell{0.59 \\ {\small $\pm$ 0.01}} & \makecell{1.00 \\ {\small $\pm$ 0.00}} & \makecell{0.99 \\ {\small $\pm$ 0.00}} & \makecell{0.99 \\ {\small $\pm$ 0.00}} & \makecell{0.60 \\ {\small $\pm$ 0.00}}\\ \hline
\textbf{RB-F} && \makecell{1.00 \\ {\small $\pm$ 0.00}} & \makecell{\textbf{0.98} \\ {\small $\pm$ 0.00}} & \makecell{0.92 \\ {\small $\pm$ 0.00}} & \makecell{0.86 \\ {\small $\pm$ 0.00}} & \makecell{1.00 \\ {\small $\pm$ 0.00}} & \makecell{0.88 \\ {\small $\pm$ 0.00}} & \makecell{0.88 \\ {\small $\pm$ 0.00}} & \makecell{\textbf{0.63} \\ {\small $\pm$ 0.00}} & \makecell{1.00 \\ {\small $\pm$ 0.00}} & \makecell{0.99 \\ {\small $\pm$ 0.00}} & \makecell{0.99 \\ {\small $\pm$ 0.00}} & \makecell{0.59 \\ {\small $\pm$ 0.00}}\\ \hline
\hline
\end{tabular}}
\end{table} 
Given a fixed skew level of test data, the performance of all Boosting ensembles declines in terms of F-measure and AUPR as the overlap between positive and negative classes grows. This decline of performance is more significant when test data is imbalanced compared to the case where test data is balanced. 
In our experiments, changes in skew level of test data result in different number of misclassified negative samples and no change in the number of correctly classified positive samples. Therefore, even for the same level of overlap, the performance of all ensembles degrades, in terms of both F-measure and AUPR when testing on a more imbalanced data.

For the same level of overlap and different imbalance of training data ($D_1$ and $D_3$) the performance of all Boosting ensembles is lower when imbalance of training data is lower. The reason is that less information is provided for training and also the skew level of training and test data has a greater difference. Overall, Table~\ref{baseF} shows that RUS is the most robust to changes in overlap and imbalance.
Based on the Table~\ref{baseF}, the following results are obtained.
Using the proposed loss factor improves the performance of Ada with $D_2$ and $D_3$ for all $\lambda_{\rm test}$, and  $D_1$ with $\lambda_{\rm test}=1:100$. However, the performance drops slightly with $D_1$ and $\lambda_{\rm test}=1:1, 1:20$ and $1:50$.
Performance of SMT improves with $D_3$ for all $\lambda_{\rm test}$, and it declines with $D_2$ and $\lambda_{\rm test}=1:20, 1:50$ and $1:100$ as well as with $D_1$ and $\lambda_{\rm test}=1:1, 1:20$.
 Performance of RUS improves with $D_1$ and $D_2$ for all $\lambda_{\rm test}$, and with $D_3$ and $\lambda_{\rm test}=1:100$. However, performance of RUSBoost degrades slightly with $D_3$ and $\lambda_{\rm test}=1:20, 1:50$ and $1:100$.
 Performance of RB improves with $D_3$ for all $\lambda_{\rm test}$, and with $D_1$, but declines with $D_2$.
Based on the Table \ref{baseF}, the value of AUPR of Boosting ensembles after using the proposed loss factor does not change in most cases.

Using the proposed loss factor may improve the performance of the Boosting ensembles that rely on under-sampling of data in terms of F-measure, especially for more difficult problems with overlapping data. The performance of Boosting ensembles that involve up-sampling of positive samples does not improve significantly.
However, the use of proposed loss factor has no impact on the global performance of these Boosting ensembles in terms of AUPR. Therefore, the use of proposed loss factor performs similarly to adjusting the
decision threshold of the Boosting algorithms to better account for imbalance.

In Table~\ref{FvarB}, the performance of baseline ensembles and their variants for different values of $\beta$ is compared for $D2$. The goal is to evaluate the effect of the value of $\beta$ on improving the performance when the proposed loss factor is used. This Table shows the performance only when $\lambda_{\rm test}=1:100$ because the performance of baseline systems usually decline for higher skew levels of test data.

\begin{table}  [!b]
 \centering
  \caption{Average of F$_2$-measure performance of baseline techniques with and without proposed loss factor for different values of $\beta$ on $D_2$, $\lambda_{\rm test}=1:100$.}
\label{FvarB}
  \resizebox{0.8\textwidth}{!}{
\begin{tabular}{l||  c c c c c ||  c c c c c} \hline \hline
\multirow{2}{*}{\textbf{Ensembles }}&\multicolumn{5}{c||}{\textbf{$F_D$}} & \multicolumn{5}{c}{\textbf{$F_{\rm op}$}}\\
& $\beta=1$& $\beta=2$& $\beta=4$&$\beta=7$& $\beta=10$& $\beta=1$& $\beta=2$& $\beta=4$&$\beta=7$& $\beta=10$ \\ \hline \hline
	\textbf{Ada}   & \makecell{0.41 \\{\small $\pm$ 0.01}} & \makecell{0.36 \\ {\small $\pm$ 0.00}} & \makecell{0.44 \\ {\small $\pm$ 0.03}} & \makecell{0.82 \\ {\small $\pm$ 0.02}} & \makecell{0.86 \\ {\small $\pm$ 0.02}} & \makecell{0.49 \\ {\small $\pm$ 0.01}} & \makecell{0.61 \\ {\small $\pm$ 0.01}} & \makecell{0.77 \\ {\small $\pm$ 0.01}} & \makecell{0.86 \\ {\small $\pm$ 0.01}} & \makecell{0.88 \\ {\small $\pm$ 0.01}}\\ \hline
	\textbf{Ada-F} & \makecell{0.40 \\ {\small $\pm$ 0.01}} & \makecell{\textbf{0.50} \\ {\small $\pm$ 0.01}} & \makecell{\textbf{0.73} \\ {\small $\pm$ 0.02}} & \makecell{0.82 \\ {\small $\pm$ 0.02}} & \makecell{\textbf{0.89} \\ {\small $\pm$ 0.03}} & \makecell{\textbf{0.52} \\ {\small $\pm$ 0.01}} & \makecell{\textbf{0.64} \\ {\small $\pm$ 0.01}} & \makecell{\textbf{0.80} \\ {\small $\pm$ 0.01}} & \makecell{0.85 \\ {\small $\pm$ 0.01}} & \makecell{\textbf{0.89} \\ {\small $\pm$ 0.01}}\\ \hline\hline
	\textbf{SMT}   & \makecell{0.36 \\ {\small $\pm$ 0.00}} & \makecell{0.58 \\ {\small $\pm$ 0.00}} & \makecell{0.82 \\ {\small $\pm$ 0.00}} & \makecell{0.93 \\ {\small $\pm$ 0.00}} & \makecell{0.97 \\ {\small $\pm$ 0.00}} & \makecell{0.43 \\ {\small $\pm$ 0.01}} & \makecell{0.64 \\ {\small $\pm$ 0.01}} & \makecell{0.83 \\ {\small $\pm$ 0.00}} & \makecell{0.83 \\ {\small $\pm$ 0.01}} & \makecell{0.90 \\ {\small $\pm$ 0.01}}\\ \hline
	\textbf{SMT-F} & \makecell{0.36 \\ {\small $\pm$ 0.00}} & \makecell{0.57 \\ {\small $\pm$ 0.00}} & \makecell{\textbf{0.83} \\ {\small $\pm$ 0.00}} & \makecell{0.93 \\ {\small $\pm$ 0.00}} & \makecell{0.97 \\ {\small $\pm$ 0.00}} & \makecell{\textbf{0.47} \\ {\small $\pm$ 0.01}} & \makecell{0.60 \\ {\small $\pm$ 0.01}} & \makecell{0.81 \\ {\small $\pm$ 0.01}} & \makecell{\textbf{0.92} \\ {\small $\pm$ 0.00}} & \makecell{\textbf{0.92} \\ {\small $\pm$ 0.01}}\\ \hline\hline
	\textbf{RUS}   & \makecell{0.10 \\ {\small $\pm$ 0.00}} & \makecell{0.22 \\ {\small $\pm$ 0.00}} & \makecell{0.48 \\ {\small $\pm$ 0.00}} & \makecell{0.74 \\ {\small $\pm$ 0.00}} & \makecell{0.85 \\ {\small $\pm$ 0.00}} & \makecell{0.53 \\ {\small $\pm$ 0.01}} & \makecell{0.66 \\ {\small $\pm$ 0.01}} & \makecell{0.81 \\ {\small $\pm$ 0.01}} & \makecell{0.88 \\ {\small $\pm$ 0.01}} & \makecell{0.90 \\ {\small $\pm$ 0.00}}\\ \hline
	\textbf{RUS-F} & \makecell{\textbf{0.22} \\ {\small $\pm$ 0.00}} & \makecell{\textbf{0.26} \\ {\small $\pm$ 0.01}} & \makecell{0.48 \\ {\small $\pm$ 0.00}} & \makecell{0.73 \\ {\small $\pm$ 0.00}} & \makecell{0.85 \\ {\small $\pm$ 0.00}} & \makecell{\textbf{0.56} \\ {\small $\pm$ 0.00}} & \makecell{\textbf{0.68} \\ {\small $\pm$ 0.01}} & \makecell{\textbf{0.83} \\ {\small $\pm$ 0.00}} & \makecell{0.88 \\ {\small $\pm$ 0.01}} & \makecell{0.89 \\ {\small $\pm$ 0.01}}\\ \hline\hline
	\textbf{RB}    & \makecell{0.45 \\ {\small $\pm$ 0.00}} & \makecell{0.67 \\ {\small $\pm$ 0.00}} & \makecell{0.88 \\ {\small $\pm$ 0.00}} & \makecell{0.95 \\ {\small $\pm$ 0.00}} & \makecell{0.97 \\ {\small $\pm$ 0.00}} & \makecell{0.47 \\ {\small $\pm$ 0.01}} & \makecell{0.67 \\ {\small $\pm$ 0.00}} & \makecell{0.85 \\ {\small $\pm$ 0.00}} & \makecell{0.94 \\ {\small $\pm$ 0.00}} & \makecell{0.96 \\ {\small $\pm$ 0.00}}\\ \hline
    \textbf{RB-F} & \makecell{0.43 \\ {\small $\pm$ 0.00}} & \makecell{0.65 \\ {\small $\pm$ 0.00}} & \makecell{0.86 \\ {\small $\pm$ 0.00}} & \makecell{0.95 \\ {\small $\pm$ 0.00}} & \makecell{0.97 \\ {\small $\pm$ 0.00}} & \makecell{0.45 \\ {\small $\pm$ 0.01}} & \makecell{\textbf{0.70} \\ {\small $\pm$ 0.00}} & \makecell{\textbf{0.86} \\ {\small $\pm$ 0.00}} & \makecell{0.93 \\ {\small $\pm$ 0.00}} & \makecell{0.96 \\ {\small $\pm$ 0.00}}\\ \hline\hline
\end{tabular}}
\end{table}

Evaluation is done in terms of the same F$_\beta$-measure that is used in loss factor calculation. The results are shown in terms of both F$_D$ and F$_{\rm op}$. F$_D$ is the value of F-measure when the decisions of base classifiers are combined in Boosting ensembles and F$_{\rm op}$ is the value of F-measure when the scores of base classifiers are combined in Boosting ensembles and the optimal decision threshold of each ensemble is set to the point that maximizes F-measure when that ensemble is validated on an independent set of data (see section 4.2.1.).

The performance of Ada improves for most value of $\beta$ in terms of both F$_D$ and F$_{\rm op}$. Some improvements are seen for SMT and RB in terms of F$_{\rm op}$, but F$_D$ tend to stay the same in most cases and decrease in some cases.
The performance of RUS improves for $\beta=1$ and $2$ in terms of both F$_D$ and F$_{\rm op}$, and the improvement tends to decrease for higher $\beta$ values. This was expected, since using higher values of $\beta$ to calculate F-measure means giving more importance to recall than precision. Therefore, the impact of imbalance is masked when higher values of $\beta$ is used and the performance may not change when the loss factor is calculated based on F-measure. For each of the classification systems, the same reason result in higher values of F$_D$ and F$_{\rm op}$ with higher values of $\beta$.
Comparing F$_D$ and F$_{\rm op}$ of each ensemble for each value of $\beta$ shows that selecting the proper decision threshold can improve the performance in terms of accuracy and robustness, especially for lower values of $\beta$. 

The results are not shown in terms of AUPR because AUPR does not change with variations in the value of $\beta$, since the importance of recall and precision stays the same and equal in obtaining AUPR.

\subsubsection{Impact of progressive partitioning in RUSBoost}
In this section, progressive partitioning is integrated into RUS without the use of F-measure in loss factor calculation. It is observed in Table~\ref{PrF} that robustness of RUS improves significantly after using this method of sampling. Indeed, using all samples for training through partitioning avoids loss of information and may improve the classification accuracy. In addition, validating on different imbalance levels of data increases the robustness to variations in the imbalance level of test data.  

The performance of PRUS and PCUS$_i$ is significantly better than RUS in terms of both F-measure and AUPR, especially with $D_2$ and $D_3$.
Partitioning these datasets using PCUS$_i$, and random under-sampling without replacement is more effective in improving the performance of RUS.

\begin{table}  [!b]
 \centering
  \caption{Average of F$_2$-measure and AUPR performance of RUSBoost with and without integrating progressive Boosting on synthetic data over different levels of skew and overlap of test data.}
\label{PrF}
  \resizebox{1\textwidth}{!}{
\begin{tabular}{l c|| c|c|c|c|| c|c|c|c|| c|c|c|c } \hline \hline
\multirow{2}[2]{*}{\textbf{Ensembles }}&\makecell{\textbf{Train} \\ \textbf{Data}}& \multicolumn{4}{c||}{\boldmath$D_1$  ($\boldsymbol{\lambda_{\rm train}}=$1:50, $\boldsymbol{\delta}=0.2$) } & \multicolumn{4}{c||}{\boldmath$D_2$  ($\boldsymbol{\lambda_{\rm train}}=$1:50, $\boldsymbol{\delta}=0.1$)} & \multicolumn{4}{c}{\boldmath\boldmath$D_3$  ($\boldsymbol{\lambda_{\rm train}}=$1:20, $\boldsymbol{\delta}=0.2$) } \\ \cline{2-14}
&$\boldsymbol{\lambda_{\rm test}}$& \textbf{1:1} & \textbf{1:20} & \textbf{1:50} & \textbf{1:100} & \textbf{1:1} & \textbf{1:20} & \textbf{1:50} & \textbf{1:100} &\textbf{1:1} & \textbf{1:20} & \textbf{1:50} & \textbf{1:100}  \\ \hline \hline
\multicolumn{14}{c}{\textbf{F$_2$-measure}}\\\hline \hline
\textbf{RUS} && \makecell{0.93 \\ {\small $\pm$ 0.00}} & \makecell{0.89 \\ {\small $\pm$ 0.00}} & \makecell{0.86 \\ {\small $\pm$ 0.00}} & \makecell{0.83 \\ {\small $\pm$ 0.00}} & \makecell{0.85 \\ {\small $\pm$ 0.01}} & \makecell{0.82 \\ {\small $\pm$ 0.01}} & \makecell{0.81 \\ {\small $\pm$ 0.01}} & \makecell{0.64 \\ {\small $\pm$ 0.00}} & \makecell{0.91 \\ {\small $\pm$ 0.01}} & \makecell{0.91 \\ {\small $\pm$ 0.01}} & \makecell{0.91 \\ {\small $\pm$ 0.01}} & \makecell{0.81 \\ {\small $\pm$ 0.01}}\\ \hline
\textbf{PRUS} && \makecell{0.89 \\ {\small $\pm$ 0.01}} & \makecell{0.86 \\ {\small $\pm$ 0.00}} & \makecell{0.84 \\ {\small $\pm$ 0.00}} & \makecell{0.83 \\ {\small $\pm$ 0.00}} & \makecell{\textbf{0.86} \\ {\small $\pm$ 0.01}} & \makecell{\textbf{0.83} \\ {\small $\pm$ 0.01}} & \makecell{\textbf{0.82} \\ {\small $\pm$ 0.01}} & \makecell{\textbf{0.66} \\ {\small $\pm$ 0.01}} & \makecell{\textbf{0.96} \\ {\small $\pm$ 0.00}} & \makecell{\textbf{0.96} \\ {\small $\pm$ 0.00}} & \makecell{\textbf{0.96} \\ {\small $\pm$ 0.00}} & \makecell{\textbf{0.84} \\ {\small $\pm$ 0.00}}\\ \hline
\textbf{PCUS} && \makecell{0.83 \\ {\small $\pm$ 0.01}} & \makecell{0.81 \\ {\small $\pm$ 0.01}} & \makecell{0.79 \\ {\small $\pm$ 0.01}} & \makecell{0.77 \\ {\small $\pm$ 0.01}} & \makecell{0.78 \\ {\small $\pm$ 0.01}} & \makecell{0.74 \\ {\small $\pm$ 0.01}} & \makecell{0.74 \\ {\small $\pm$ 0.01}} & \makecell{0.59 \\ {\small $\pm$ 0.01}} & \makecell{0.91 \\ {\small $\pm$ 0.00}} & \makecell{0.91 \\ {\small $\pm$ 0.00}} & \makecell{0.91 \\ {\small $\pm$ 0.00}} & \makecell{0.79 \\ {\small $\pm$ 0.01}}\\ \hline
\textbf{PCUS$_i$} && \makecell{0.91 \\ {\small $\pm$ 0.00}} & \makecell{0.88 \\ {\small $\pm$ 0.00}} & \makecell{0.85 \\ {\small $\pm$ 0.00}} & \makecell{0.83 \\ {\small $\pm$ 0.00}} & \makecell{\textbf{0.87} \\ {\small $\pm$ 0.01}} & \makecell{\textbf{0.84} \\ {\small $\pm$ 0.01}} & \makecell{\textbf{0.83} \\ {\small $\pm$ 0.01}} & \makecell{\textbf{0.65} \\ {\small $\pm$ 0.01}} & \makecell{\textbf{0.93} \\ {\small $\pm$ 0.00}} & \makecell{\textbf{0.93} \\ {\small $\pm$ 0.00}} & \makecell{\textbf{0.93} \\ {\small $\pm$ 0.00}} & \makecell{\textbf{0.84} \\ {\small $\pm$ 0.00}}\\ \hline
 \hline 
%\end{tabular}}
%\end{table}
%\begin{table}  [!ht]
%\centering
%\caption{Average of AUPR performance of RUSBoost with and without integrating progressive Boosting on synthetic data over different levels of skew and overlap of test data.}
%\label{PrAU}
%\resizebox{1\textwidth}{!}{
%\begin{tabular}{l c|| c|c|c|c|| c|c|c|c|| c|c|c|c } \hline \hline
%\multirow{2}[2]{*}{\textbf{Ensembles }}&\makecell{\textbf{Train} \\ \textbf{Data}}& \multicolumn{4}{c||}{\boldmath$D_1$  ($\boldsymbol{\lambda_{\rm train}}=$1:50, $\boldsymbol{\delta}=0.2$) } & \multicolumn{4}{c||}{\boldmath$D_2$  ($\boldsymbol{\lambda_{\rm train}}=$1:50, $\boldsymbol{\delta}=0.1$)} & \multicolumn{4}{c}{\boldmath\boldmath$D_3$  ($\boldsymbol{\lambda_{\rm train}}=$1:20, $\boldsymbol{\delta}=0.2$) } \\ \cline{2-14}
%&$\boldsymbol{\lambda_{\rm test}}$& \textbf{1:1} & \textbf{1:20} & \textbf{1:50} & \textbf{1:100} & \textbf{1:1} & \textbf{1:20} & \textbf{1:50} & \textbf{1:100} &\textbf{1:1} & \textbf{1:20} & \textbf{1:50} & \textbf{1:100}  \\ \hline \hline
\multicolumn{14}{c}{\textbf{AUPR}}\\\hline \hline
\textbf{RUS} && \makecell{1.00 \\ {\small $\pm$ 0.00}} & \makecell{0.66 \\ {\small $\pm$ 0.00}} & \makecell{0.60 \\ {\small $\pm$ 0.00}} & \makecell{0.57 \\ {\small $\pm$ 0.00}} & \makecell{1.00 \\ {\small $\pm$ 0.00}} & \makecell{0.74 \\ {\small $\pm$ 0.00}} & \makecell{0.66 \\ {\small $\pm$ 0.00}} & \makecell{0.55 \\ {\small $\pm$ 0.00}} & \makecell{1.00 \\ {\small $\pm$ 0.00}} & \makecell{0.86 \\ {\small $\pm$ 0.01}} & \makecell{0.81 \\ {\small $\pm$ 0.01}} & \makecell{0.57 \\ {\small $\pm$ 0.00}}\\ \hline
\textbf{PRUS} && \makecell{1.00 \\ {\small $\pm$ 0.00}} & \makecell{\textbf{0.95} \\ {\small $\pm$ 0.00}} & \makecell{\textbf{0.89} \\ {\small $\pm$ 0.00}} & \makecell{\textbf{0.87} \\ {\small $\pm$ 0.00}} & \makecell{1.00 \\ {\small $\pm$ 0.00}} & \makecell{\textbf{0.90} \\ {\small $\pm$ 0.00}} & \makecell{\textbf{0.89} \\ {\small $\pm$ 0.00}} & \makecell{\textbf{0.61} \\ {\small $\pm$ 0.00}} & \makecell{1.00 \\ {\small $\pm$ 0.00}} & \makecell{\textbf{1.00} \\ {\small $\pm$ 0.00}} & \makecell{\textbf{1.00} \\ {\small $\pm$ 0.00}} & \makecell{\textbf{0.78} \\ {\small $\pm$ 0.01}}\\ \hline
\textbf{PCUS} && \makecell{0.99 \\ {\small $\pm$ 0.00}} & \makecell{\textbf{0.91} \\ {\small $\pm$ 0.00}} & \makecell{\textbf{0.85} \\ {\small $\pm$ 0.01}} & \makecell{\textbf{0.80} \\ {\small $\pm$ 0.01}} & \makecell{1.00 \\ {\small $\pm$ 0.00}} & \makecell{\textbf{0.85} \\ {\small $\pm$ 0.01}} & \makecell{\textbf{0.84} \\ {\small $\pm$ 0.01}} & \makecell{\textbf{0.56} \\ {\small $\pm$ 0.01}} & \makecell{1.00 \\ {\small $\pm$ 0.00}} & \makecell{\textbf{1.00} \\ {\small $\pm$ 0.00}} & \makecell{\textbf{0.99} \\ {\small $\pm$ 0.00}} & \makecell{\textbf{0.77} \\ {\small $\pm$ 0.01}}\\ \hline
\textbf{PCUS$_i$} && \makecell{1.00 \\ {\small $\pm$ 0.00}} & \makecell{\textbf{0.93} \\ {\small $\pm$ 0.00}} & \makecell{\textbf{0.87} \\ {\small $\pm$ 0.00}} & \makecell{\textbf{0.82} \\ {\small $\pm$ 0.00}} & \makecell{1.00 \\ {\small $\pm$ 0.00}} & \makecell{\textbf{0.88} \\ {\small $\pm$ 0.00}} & \makecell{\textbf{0.87} \\ {\small $\pm$ 0.00}} & \makecell{\textbf{0.65} \\ {\small $\pm$ 0.00}} & \makecell{1.00 \\ {\small $\pm$ 0.00}} & \makecell{\textbf{1.00} \\ {\small $\pm$ 0.00}} & \makecell{\textbf{0.99} \\ {\small $\pm$ 0.00}} & \makecell{\textbf{0.82} \\ {\small $\pm$ 0.01}}\\ \hline
\hline 
 \end{tabular}}
\end{table}

\begin{table}  [!t]
 \centering
  \caption{Average of F$_2$-measure and AUPR performance of proposed and baseline techniques on synthetic data over different levels of skew and overlap of test data.}
\label{PrFF}
  \resizebox{1\textwidth}{!}{
\begin{tabular}{l c|| c|c|c|c|| c|c|c|c|| c|c|c|c } \hline \hline
\multirow{2}[2]{*}{\textbf{Ensembles }}&\makecell{\textbf{Train} \\ \textbf{Data}}& \multicolumn{4}{c||}{\boldmath$D_1$  ($\boldsymbol{\lambda_{\rm train}}=$1:50, $\boldsymbol{\delta}=0.2$) } & \multicolumn{4}{c||}{\boldmath$D_2$  ($\boldsymbol{\lambda_{\rm train}}=$1:50, $\boldsymbol{\delta}=0.1$)} & \multicolumn{4}{c}{\boldmath$D_3$  ($\boldsymbol{\lambda_{\rm train}}=$1:20, $\boldsymbol{\delta}=0.2$) } \\ \cline{2-14}
&$\boldsymbol{\lambda_{\rm test}}$& \textbf{1:1} & \textbf{1:20} & \textbf{1:50} & \textbf{1:100} & \textbf{1:1} & \textbf{1:20} & \textbf{1:50} & \textbf{1:100} &\textbf{1:1} & \textbf{1:20} & \textbf{1:50} & \textbf{1:100}  \\ \hline \hline
\multicolumn{14}{c}{\textbf{F$_2$-measure}}\\\hline \hline
\textbf{Ada} && \makecell{0.98 \\ {\small $\pm$ 0.00}} & \makecell{0.94 \\ {\small $\pm$ 0.00}} & \makecell{\textbf{0.91} \\ {\small $\pm$ 0.00}} & \makecell{0.86 \\ {\small $\pm$ 0.00}} & \makecell{0.94 \\ {\small $\pm$ 0.01}} & \makecell{0.85 \\ {\small $\pm$ 0.01}} & \makecell{0.83 \\ {\small $\pm$ 0.01}} & \makecell{0.51 \\ {\small $\pm$ 0.01}} & \makecell{0.89 \\ {\small $\pm$ 0.01}} & \makecell{0.88 \\ {\small $\pm$ 0.01}} & \makecell{0.88 \\ {\small $\pm$ 0.01}} & \makecell{0.46 \\ {\small $\pm$ 0.00}}\\ \hline
\textbf{SMT} && \makecell{0.96 \\ {\small $\pm$ 0.01}} & \makecell{0.90 \\ {\small $\pm$ 0.00}} & \makecell{0.84 \\ {\small $\pm$ 0.00}} & \makecell{0.81 \\ {\small $\pm$ 0.00}} & \makecell{0.93 \\ {\small $\pm$ 0.01}} & \makecell{0.86 \\ {\small $\pm$ 0.01}} & \makecell{0.84 \\ {\small $\pm$ 0.01}} & \makecell{0.58 \\ {\small $\pm$ 0.01}} & \makecell{0.92 \\ {\small $\pm$ 0.01}} & \makecell{0.92 \\ {\small $\pm$ 0.01}} & \makecell{0.92 \\ {\small $\pm$ 0.01}} & \makecell{0.56 \\ {\small $\pm$ 0.01}}\\ \hline
\textbf{RB} && \makecell{\textbf{0.99} \\ {\small $\pm$ 0.00}} & \makecell{0.94 \\ {\small $\pm$ 0.00}} & \makecell{0.89 \\ {\small $\pm$ 0.00}} & \makecell{\textbf{0.86} \\ {\small $\pm$ 0.00}} & \makecell{\textbf{0.99} \\ {\small $\pm$ 0.00}} & \makecell{0.91 \\ {\small $\pm$ 0.00}} & \makecell{0.91 \\ {\small $\pm$ 0.00}} & \makecell{0.61 \\ {\small $\pm$ 0.00}} & \makecell{0.93 \\ {\small $\pm$ 0.01}} & \makecell{0.93 \\ {\small $\pm$ 0.01}} & \makecell{0.93 \\ {\small $\pm$ 0.01}} & \makecell{0.50 \\ {\small $\pm$ 0.01}}\\ \hline
\textbf{RUS} && \makecell{0.93 \\ {\small $\pm$ 0.00}} & \makecell{0.89 \\ {\small $\pm$ 0.00}} & \makecell{0.86 \\ {\small $\pm$ 0.00}} & \makecell{0.83 \\ {\small $\pm$ 0.00}} & \makecell{0.85 \\ {\small $\pm$ 0.01}} & \makecell{0.82 \\ {\small $\pm$ 0.01}} & \makecell{0.81 \\ {\small $\pm$ 0.01}} & \makecell{0.64 \\ {\small $\pm$ 0.00}} & \makecell{0.91 \\ {\small $\pm$ 0.01}} & \makecell{0.91 \\ {\small $\pm$ 0.01}} & \makecell{0.91 \\ {\small $\pm$ 0.01}} & \makecell{0.81 \\ {\small $\pm$ 0.01}}\\ \hline\hline
\textbf{PRUS-F} && \makecell{0.89 \\ {\small $\pm$ 0.00}} & \makecell{0.86 \\ {\small $\pm$ 0.00}} & \makecell{0.84 \\ {\small $\pm$ 0.00}} & \makecell{0.82 \\ {\small $\pm$ 0.00}} & \makecell{0.88 \\ {\small $\pm$ 0.01}} & \makecell{0.84 \\ {\small $\pm$ 0.01}} & \makecell{0.84 \\ {\small $\pm$ 0.01}} & \makecell{\textbf{0.66} \\ {\small $\pm$ 0.00}} & \makecell{\textbf{0.94} \\ {\small $\pm$ 0.01}} & \makecell{\textbf{0.94} \\ {\small $\pm$ 0.01}} & \makecell{\textbf{0.94} \\ {\small $\pm$ 0.01}} & \makecell{\textbf{0.81} \\ {\small $\pm$ 0.01}}\\ \hline
\textbf{PCUS-F} && \makecell{0.76 \\ {\small $\pm$ 0.01}} & \makecell{0.74 \\ {\small $\pm$ 0.01}} & \makecell{0.72 \\ {\small $\pm$ 0.01}} & \makecell{0.71 \\ {\small $\pm$ 0.01}} & \makecell{0.77 \\ {\small $\pm$ 0.01}} & \makecell{0.73 \\ {\small $\pm$ 0.01}} & \makecell{0.73 \\ {\small $\pm$ 0.01}} & \makecell{0.60 \\ {\small $\pm$ 0.01}} & \makecell{0.92 \\ {\small $\pm$ 0.01}} & \makecell{0.92 \\ {\small $\pm$ 0.01}} & \makecell{0.92 \\ {\small $\pm$ 0.01}} & \makecell{0.79 \\ {\small $\pm$ 0.01}}\\ \hline
\textbf{PCUS$_i$-F} && \makecell{\textbf{0.99}\\ {\small $\pm$ 0.00}} & \makecell{\textbf{0.95} \\ {\small $\pm$ 0.01}} & \makecell{0.89 \\ {\small $\pm$ 0.00}} & \makecell{\textbf{0.86} \\ {\small $\pm$ 0.00}} & \makecell{\textbf{0.99} \\ {\small $\pm$ 0.00}} & \makecell{\textbf{0.92} \\ {\small $\pm$ 0.01}} & \makecell{\textbf{0.91} \\ {\small $\pm$ 0.01}} & \makecell{\textbf{0.67} \\ {\small $\pm$ 0.00}} & \makecell{\textbf{0.95} \\ {\small $\pm$ 0.01}} & \makecell{\textbf{0.95} \\ {\small $\pm$ 0.01}} & \makecell{\textbf{0.95} \\ {\small $\pm$ 0.01}} & \makecell{\textbf{0.84} \\ {\small $\pm$ 0.01}}\\ \hline
 \hline
%\end{tabular}}
%\end{table}
%
%
%\begin{table} [!htb]
% \centering
%  \caption{Average of AUPR performance of proposed and baseline techniques on synthetic data over different levels of skew and overlap of test data.}
%\label{PrFAU}
%  \resizebox{1\textwidth}{!}{
%\begin{tabular}{l c|| c|c|c|c|| c|c|c|c|| c|c|c|c } \hline \hline
%\multirow{2}[2]{*}{\textbf{Ensembles }}&\makecell{\textbf{Train} \\ \textbf{Data}}& \multicolumn{4}{c||}{\boldmath$D_1$  ($\boldsymbol{\lambda_{\rm train}}=$1:50, $\boldsymbol{\delta}=0.2$) } & \multicolumn{4}{c||}{\boldmath$D_2$  ($\boldsymbol{\lambda_{\rm train}}=$1:50, $\boldsymbol{\delta}=0.1$)} & \multicolumn{4}{c}{\boldmath$D_3$  ($\boldsymbol{\lambda_{\rm train}}=$1:20, $\boldsymbol{\delta}=0.2$) } \\ \cline{2-14}
%&$\boldsymbol{\lambda_{\rm test}}$& \textbf{1:1} & \textbf{1:20} & \textbf{1:50} & \textbf{1:100} & \textbf{1:1} & \textbf{1:20} & \textbf{1:50} & \textbf{1:100} &\textbf{1:1} & \textbf{1:20} & \textbf{1:50} & \textbf{1:100}  \\ \hline \hline
\multicolumn{14}{c}{\textbf{AUPR}}\\\hline \hline
\textbf{Ada} && \makecell{0.99 \\ {\small $\pm$ 0.00}} & \makecell{\textbf{0.97} \\ {\small $\pm$ 0.00}} & \makecell{\textbf{0.93} \\ {\small $\pm$ 0.00}} & \makecell{0.85 \\ {\small $\pm$ 0.00}} & \makecell{1.00 \\ {\small $\pm$ 0.00}} & \makecell{0.88 \\ {\small $\pm$ 0.00}} & \makecell{\textbf{0.88} \\ {\small $\pm$ 0.00}} & \makecell{0.60 \\ {\small $\pm$ 0.00}} & \makecell{1.00 \\ {\small $\pm$ 0.00}} & \makecell{\textbf{1.00} \\ {\small $\pm$ 0.00}} & \makecell{\textbf{1.00} \\ {\small $\pm$ 0.00}} & \makecell{0.58 \\ {\small $\pm$ 0.00}}\\ \hline
\textbf{SMT} && \makecell{1.00 \\ {\small $\pm$ 0.00}} & \makecell{0.90 \\ {\small $\pm$ 0.00}} & \makecell{0.82 \\ {\small $\pm$ 0.00}} & \makecell{0.78 \\ {\small $\pm$ 0.00}} & \makecell{1.00 \\ {\small $\pm$ 0.00}} & \makecell{0.83 \\ {\small $\pm$ 0.00}} & \makecell{0.81 \\ {\small $\pm$ 0.00}} & \makecell{0.60 \\ {\small $\pm$ 0.00}} & \makecell{1.00 \\ {\small $\pm$ 0.00}} & \makecell{0.98 \\ {\small $\pm$ 0.00}} & \makecell{0.97 \\ {\small $\pm$ 0.00}} & \makecell{0.58 \\ {\small $\pm$ 0.00}}\\ \hline
\textbf{RB} && \makecell{1.00 \\ {\small $\pm$ 0.00}} & \makecell{\textbf{0.97} \\ {\small $\pm$ 0.00}} & \makecell{\textbf{0.93} \\ {\small $\pm$ 0.00}} & \makecell{\textbf{0.87} \\ {\small $\pm$ 0.00}} & \makecell{1.00 \\ {\small $\pm$ 0.00}} & \makecell{0.88 \\ {\small $\pm$ 0.00}} & \makecell{\textbf{0.88} \\ {\small $\pm$ 0.00}} & \makecell{0.59 \\ {\small $\pm$ 0.01}} & \makecell{1.00 \\ {\small $\pm$ 0.00}} & \makecell{0.99 \\ {\small $\pm$ 0.00}} & \makecell{0.99 \\ {\small $\pm$ 0.00}} & \makecell{0.60 \\ {\small $\pm$ 0.00}}\\ \hline
\textbf{RUS} && \makecell{1.00 \\ {\small $\pm$ 0.00}} & \makecell{0.66 \\ {\small $\pm$ 0.00}} & \makecell{0.60 \\ {\small $\pm$ 0.00}} & \makecell{0.57 \\ {\small $\pm$ 0.00}} & \makecell{1.00 \\ {\small $\pm$ 0.00}} & \makecell{0.74 \\ {\small $\pm$ 0.00}} & \makecell{0.66 \\ {\small $\pm$ 0.00}} & \makecell{0.55 \\ {\small $\pm$ 0.00}} & \makecell{1.00 \\ {\small $\pm$ 0.00}} & \makecell{0.86 \\ {\small $\pm$ 0.01}} & \makecell{0.81 \\ {\small $\pm$ 0.01}} & \makecell{0.57 \\ {\small $\pm$ 0.00}}\\ \hline\hline
\textbf{PRUS-F} && \makecell{1.00 \\ {\small $\pm$ 0.00}} & \makecell{0.94 \\ {\small $\pm$ 0.00}} & \makecell{0.90 \\ {\small $\pm$ 0.00}} & \makecell{\textbf{0.87} \\ {\small $\pm$ 0.00}} & \makecell{1.00 \\ {\small $\pm$ 0.00}} & \makecell{\textbf{0.90} \\ {\small $\pm$ 0.00}} & \makecell{\textbf{0.90} \\ {\small $\pm$ 0.00}} & \makecell{\textbf{0.61} \\ {\small $\pm$ 0.00}} & \makecell{1.00 \\ {\small $\pm$ 0.00}} & \makecell{\textbf{1.00} \\ {\small $\pm$ 0.00}} & \makecell{\textbf{1.00} \\ {\small $\pm$ 0.00}} & \makecell{\textbf{0.76} \\ {\small $\pm$ 0.01}}\\ \hline
\textbf{PCUS-F} && \makecell{0.99 \\ {\small $\pm$ 0.00}} & \makecell{0.89 \\ {\small $\pm$ 0.01}} & \makecell{0.83 \\ {\small $\pm$ 0.01}} & \makecell{0.79 \\ {\small $\pm$ 0.01}} & \makecell{1.00 \\ {\small $\pm$ 0.00}} & \makecell{0.86 \\ {\small $\pm$ 0.01}} & \makecell{0.85 \\ {\small $\pm$ 0.01}} & \makecell{0.56 \\ {\small $\pm$ 0.01}} & \makecell{1.00 \\ {\small $\pm$ 0.00}} & \makecell{0.99 \\ {\small $\pm$ 0.00}} & \makecell{0.99 \\ {\small $\pm$ 0.00}} & \makecell{\textbf{0.78} \\ {\small $\pm$ 0.01}}\\ \hline
\textbf{PCUS$_i$-F} && \makecell{1.00 \\ {\small $\pm$ 0.00}} & \makecell{\textbf{0.97 }\\ {\small $\pm$ 0.00}} & \makecell{0.91 \\ {\small $\pm$ 0.00}} & \makecell{\textbf{0.87} \\ {\small $\pm$ 0.00}} & \makecell{1.00 \\ {\small $\pm$ 0.00}} & \makecell{\textbf{0.89} \\ {\small $\pm$ 0.00}} & \makecell{\textbf{0.88} \\ {\small $\pm$ 0.00}} & \makecell{\textbf{0.64} \\ {\small $\pm$ 0.00}} & \makecell{1.00 \\ {\small $\pm$ 0.00}} & \makecell{\textbf{1.00} \\ {\small $\pm$ 0.00}} & \makecell{\textbf{1.00} \\ {\small $\pm$ 0.00}} & \makecell{\textbf{0.82} \\ {\small $\pm$ 0.00}}\\ \hline
\hline 
\end{tabular}}
\end{table}

\subsubsection{Impact of progressive partitioning and loss factor combined}
In this section progressive partitioning and the proposed loss factor are integrated into RUS algorithm, resulting in PRUS-F, PCUS-F, and PCUS$_i$-F.
Over all ranges of skew and overlap in Table~\ref{PrFF}, PCUS$_i$-F outperforms other classification systems and PRUS-F takes the second place, especially for higher levels of imbalance in test data. Combining the use of F-measure and progressive partitioning is more effective in increasing performance and robustness compared to using each of them independently because accuracy and robustness to imbalance improve at the same time, not separately, during learning process. If the negative class is partitioned a priori (CUS$_i$), PBoost performs significantly better than the case when general partitioning techniques (RUS and CUS) are used.

\subsection{Results of Experiments with Video Data}
Similarly to the synthetic data sets, the results of experiments on video dataset are shown in three parts, assessing the impact of: (1) using the proposed loss factor on the performance of baseline Boosting ensembles, (2) integrating progressive partitioning into RUS, and (3) using the proposed loss factor and progressive partitioning compared with the baseline and state of the art Boosting ensembles.

From Table~\ref{baseF_v}, the performance level of all ensembles is lower when the skew level of training data is higher. This is despite the fact that when the imbalance of training data is lower, the data that is used to test classifiers contain samples from some individuals that are not in the training data.
Using the proposed loss factor improves the performance of Ada, RUS and SMT in terms of F-measure, and has no impact on the performance of RB in most cases of skew between classes in training and testing data.
The performance of these ensembles after using the proposed loss factor does not change in terms of AUPR and therefore they are not shown here. In fact, the use of proposed loss factor performs similarly to adjusting the
decision threshold of the Boosting algorithms to better account for imbalance and therefore may improve the performance only in terms of local performance metrics like F-measure.

\begin{table}  [!t]
 \centering
  \caption{Average of F$_2$-measure performance of baseline techniques before and after using the proposed loss factor on video data over different levels of skew in training and test data.}
\label{baseF_v}
  \resizebox{0.78\textwidth}{!}{
\begin{tabular}{l c|| c|c|c|c|| c|c|c|c} \hline \hline
\multirow{2}[2]{*}{\textbf{Ensembles }}&\makecell{\textbf{Train} \\ \textbf{Data}}& \multicolumn{4}{c||}{$\boldsymbol{\lambda_{\rm train}}=$1:50 } & \multicolumn{4}{c}{$\boldsymbol{\lambda_{\rm train}}=$1:100} \\ \cline{2-10}
&$\boldsymbol{\lambda_{\rm test}}$& \textbf{1:1} & \textbf{1:20} & \textbf{1:50} & \textbf{1:100} & \textbf{1:1} & \textbf{1:20} & \textbf{1:50} & \textbf{1:100}   \\ \hline \hline
%\multicolumn{10}{c}{\textbf{F$_2$-measure}}\\\hline \hline
Ada && \makecell{0.80 \\ {\small $\pm$ 0.01}} & \makecell{0.86 \\ {\small $\pm$ 0.01}} & \makecell{0.78 \\ {\small $\pm$ 0.01}} & \makecell{0.85 \\ {\small $\pm$ 0.01}} & \makecell{0.75 \\ {\small $\pm$ 0.01}} & \makecell{0.83 \\ {\small $\pm$ 0.00}} & \makecell{0.72 \\ {\small $\pm$ 0.01}} & \makecell{0.82 \\ {\small $\pm$ 0.00}}\\ \hline
Ada-F && \makecell{\textbf{0.81} \\ {\small $\pm$ 0.01}} & \makecell{\textbf{0.89} \\ {\small $\pm$ 0.00}} & \makecell{\textbf{0.80} \\ {\small $\pm$ 0.01}} & \makecell{\textbf{0.87} \\ {\small $\pm$ 0.00}} & \makecell{0.74 \\ {\small $\pm$ 0.01}} & \makecell{\textbf{0.84} \\ {\small $\pm$ 0.00}} & \makecell{0.72 \\ {\small $\pm$ 0.01}} & \makecell{0.82 \\ {\small $\pm$ 0.00}}\\ \hline\hline
SMT && \makecell{0.93 \\ {\small $\pm$ 0.00}} & \makecell{0.96 \\ {\small $\pm$ 0.00}} & \makecell{0.92 \\ {\small $\pm$ 0.00}} & \makecell{0.96 \\ {\small $\pm$ 0.00}} & \makecell{0.91 \\ {\small $\pm$ 0.00}} & \makecell{0.95 \\ {\small $\pm$ 0.00}} & \makecell{0.89 \\ {\small $\pm$ 0.00}} & \makecell{0.94 \\ {\small $\pm$ 0.00}}\\ \hline
SMT-F && \makecell{0.93 \\ {\small $\pm$ 0.00}} & \makecell{\textbf{0.98} \\ {\small $\pm$ 0.00}} & \makecell{0.92 \\ {\small $\pm$ 0.00}} & \makecell{\textbf{0.97} \\ {\small $\pm$ 0.00}} & \makecell{0.91 \\ {\small $\pm$ 0.00}} & \makecell{\textbf{0.96} \\ {\small $\pm$ 0.00}} & \makecell{\textbf{0.90} \\ {\small $\pm$ 0.00}} & \makecell{\textbf{0.95} \\ {\small $\pm$ 0.00}}\\ \hline\hline
RUS && \makecell{0.94 \\ {\small $\pm$ 0.00}} & \makecell{0.97 \\ {\small $\pm$ 0.00}} & \makecell{0.93 \\ {\small $\pm$ 0.00}} & \makecell{0.96 \\ {\small $\pm$ 0.00}} & \makecell{0.88 \\ {\small $\pm$ 0.00}} & \makecell{0.94 \\ {\small $\pm$ 0.00}} & \makecell{0.87 \\ {\small $\pm$ 0.00}} & \makecell{0.93 \\ {\small $\pm$ 0.00}}\\ \hline
RUS-F && \makecell{\textbf{0.96} \\ {\small $\pm$ 0.00}} & \makecell{\textbf{0.98} \\ {\small $\pm$ 0.00}} & \makecell{\textbf{0.94} \\ {\small $\pm$ 0.00}} & \makecell{\textbf{0.98} \\ {\small $\pm$ 0.00}} & \makecell{\textbf{0.90} \\ {\small $\pm$ 0.00}} & \makecell{\textbf{0.96} \\ {\small $\pm$ 0.00}} & \makecell{\textbf{0.88} \\ {\small $\pm$ 0.00}} & \makecell{\textbf{0.94} \\ {\small $\pm$ 0.00}}\\ \hline\hline
RB && \makecell{0.95 \\ {\small $\pm$ 0.00}} & \makecell{0.99 \\ {\small $\pm$ 0.00}} & \makecell{0.94 \\ {\small $\pm$ 0.00}} & \makecell{0.98 \\ {\small $\pm$ 0.00}} & \makecell{0.91 \\ {\small $\pm$ 0.00}} & \makecell{0.97 \\ {\small $\pm$ 0.00}} & \makecell{0.89 \\ {\small $\pm$ 0.00}} & \makecell{0.96 \\ {\small $\pm$ 0.00}}\\ \hline
RB-F && \makecell{0.95 \\ {\small $\pm$ 0.00}} & \makecell{0.98 \\ {\small $\pm$ 0.00}} & \makecell{0.93 \\ {\small $\pm$ 0.00}} & \makecell{0.98 \\ {\small $\pm$ 0.00}} & \makecell{0.90 \\ {\small $\pm$ 0.00}} & \makecell{0.97 \\ {\small $\pm$ 0.00}} & \makecell{0.88 \\ {\small $\pm$ 0.00}} & \makecell{0.96 \\ {\small $\pm$ 0.00}}\\ \hline
\end{tabular}}
\end{table}

\begin{table}  [!t]
 \centering
  \caption{Average of F$_2$-measure and AUPR performance of RUSBoost with and without integrating progressive Boosting on video data over different levels of skew in training and test data.}
\label{PrF_v}
  \resizebox{0.78\textwidth}{!}{
\begin{tabular}{l c|| c|c|c|c|| c|c|c|c} \hline \hline
\multirow{2}[2]{*}{\textbf{Ensembles }}&\makecell{\textbf{Train} \\ \textbf{Data}}& \multicolumn{4}{c||}{$\boldsymbol{\lambda_{\rm train}}=$1:50 } & \multicolumn{4}{c}{$\boldsymbol{\lambda_{\rm train}}=$1:100} \\ \cline{2-10}
&$\boldsymbol{\lambda_{\rm test}}$& \textbf{1:1} & \textbf{1:20} & \textbf{1:50} & \textbf{1:100} & \textbf{1:1} & \textbf{1:20} & \textbf{1:50} & \textbf{1:100}   \\ \hline \hline
\multicolumn{10}{c}{\textbf{F$_2$-measure}}\\\hline \hline
RUS && \makecell{0.94 \\ {\small $\pm$ 0.00}} & \makecell{0.97 \\ {\small $\pm$ 0.00}} & \makecell{0.93 \\ {\small $\pm$ 0.00}} & \makecell{0.96 \\ {\small $\pm$ 0.00}} & \makecell{0.88 \\ {\small $\pm$ 0.00}} & \makecell{0.94 \\ {\small $\pm$ 0.00}} & \makecell{0.87 \\ {\small $\pm$ 0.00}} & \makecell{0.93 \\ {\small $\pm$ 0.00}}\\ \hline
PRUS && \makecell{\textbf{0.95} \\ {\small $\pm$ 0.00}} & \makecell{0.95 \\ {\small $\pm$ 0.00}} & \makecell{0.92 \\ {\small $\pm$ 0.00}} & \makecell{\textbf{0.97} \\ {\small $\pm$ 0.00}} & \makecell{\textbf{0.91} \\ {\small $\pm$ 0.00}} & \makecell{\textbf{0.95} \\ {\small $\pm$ 0.00}} & \makecell{\textbf{0.94} \\ {\small $\pm$ 0.00}} & \makecell{\textbf{0.94} \\ {\small $\pm$ 0.00}}\\ \hline
PCUS && \makecell{0.94 \\ {\small $\pm$ 0.00}} & \makecell{0.95 \\ {\small $\pm$ 0.00}} & \makecell{0.93 \\ {\small $\pm$ 0.00}} & \makecell{0.96 \\ {\small $\pm$ 0.00}} & \makecell{\textbf{0.90} \\ {\small $\pm$ 0.00}} & \makecell{0.94 \\ {\small $\pm$ 0.00}} & \makecell{0.92 \\ {\small $\pm$ 0.00}} & \makecell{0.92 \\ {\small $\pm$ 0.00}}\\ \hline
PTUS && \makecell{\textbf{0.96} \\ {\small $\pm$ 0.00}} & \makecell{0.96 \\ {\small $\pm$ 0.00}} & \makecell{0.93 \\ {\small $\pm$ 0.00}} & \makecell{0.95 \\ {\small $\pm$ 0.00}} & \makecell{\textbf{0.92} \\ {\small $\pm$ 0.00}} & \makecell{\textbf{0.97} \\ {\small $\pm$ 0.00}} & \makecell{\textbf{0.98} \\ {\small $\pm$ 0.00}} & \makecell{\textbf{0.95} \\ {\small $\pm$ 0.00}}\\ \hline
 \hline 
%\end{tabular}}
%\end{table}
%
%
%\begin{table}  [!htb]
% \centering
%  \caption{Average of AUPR performance of RUSBoost with and without integrating progressive Boosting on video data over different levels of skew in training and test data.}
%\label{PrAU_v}
%  \resizebox{0.78\textwidth}{!}{
%\begin{tabular}{l c|| c|c|c|c|| c|c|c|c} \hline \hline
%\multirow{2}[2]{*}{\textbf{Ensembles }}&\makecell{\textbf{Train} \\ \textbf{Data}}& \multicolumn{4}{c||}{$\boldsymbol{\lambda_{\rm train}}=$1:50 } & \multicolumn{4}{c}{$\boldsymbol{\lambda_{\rm train}}=$1:100} \\ \cline{2-10}
%&$\boldsymbol{\lambda_{\rm test}}$& \textbf{1:1} & \textbf{1:20} & \textbf{1:50} & \textbf{1:100} & \textbf{1:1} & \textbf{1:20} & \textbf{1:50} & \textbf{1:100}   \\ \hline \hline
\multicolumn{10}{c}{\textbf{AUPR}}\\\hline \hline
RUS && \makecell{1.00 \\ {\small $\pm$ 0.00}} & \makecell{0.99 \\ {\small $\pm$ 0.00}} & \makecell{0.94 \\ {\small $\pm$ 0.00}} & \makecell{0.97 \\ {\small $\pm$ 0.00}} & \makecell{0.84 \\ {\small $\pm$ 0.00}} & \makecell{0.94 \\ {\small $\pm$ 0.00}} & \makecell{0.80 \\ {\small $\pm$ 0.00}} & \makecell{0.92 \\ {\small $\pm$ 0.00}}\\ \hline
PRUS && \makecell{1.00 \\ {\small $\pm$ 0.00}} & \makecell{0.99 \\ {\small $\pm$ 0.00}} & \makecell{0.92 \\ {\small $\pm$ 0.00}} & \makecell{0.96 \\ {\small $\pm$ 0.00}} & \makecell{0.89 \\ {\small $\pm$ 0.00}} & \makecell{0.94 \\ {\small $\pm$ 0.00}} & \makecell{\textbf{0.85} \\ {\small $\pm$ 0.00}} & \makecell{\textbf{0.94} \\ {\small $\pm$ 0.00}}\\ \hline
PCUS && \makecell{1.00 \\ {\small $\pm$ 0.00}} & \makecell{0.99 \\ {\small $\pm$ 0.00}} & \makecell{0.93 \\ {\small $\pm$ 0.00}} & \makecell{0.95 \\ {\small $\pm$ 0.00}} & \makecell{\textbf{0.88} \\ {\small $\pm$ 0.00}} & \makecell{0.93 \\ {\small $\pm$ 0.00}} & \makecell{\textbf{0.89} \\ {\small $\pm$ 0.00}} & \makecell{\textbf{0.93} \\ {\small $\pm$ 0.00}}\\ \hline
PTUS && \makecell{1.00 \\ {\small $\pm$ 0.00}} & \makecell{\textbf{1.00} \\ {\small $\pm$ 0.00}} & \makecell{\textbf{0.95} \\ {\small $\pm$ 0.00}} & \makecell{0.97 \\ {\small $\pm$ 0.00}} & \makecell{\textbf{0.90} \\ {\small $\pm$ 0.00}} & \makecell{\textbf{0.96} \\ {\small $\pm$ 0.00}} & \makecell{\textbf{0.89} \\ {\small $\pm$ 0.00}} & \makecell{\textbf{0.97} \\ {\small $\pm$ 0.00}}\\ \hline
\hline 
 \end{tabular}}
\end{table}

After integrating the progressive partitioning in RUS using PRUS and PTUS, the performance of RUS increases and becomes more robust in terms of both F-measure and AUPR (see Table~\ref{PrF_v}), especially when TUS is used for partitioning because validating base classifiers on different imbalance levels of imbalance result in more robust classification systems and using all samples for training through partitioning avoids loss of information and may improve the classification accuracy.

Comparing the performance of final PBoost variants with baseline ensembles in Table~\ref{PrFF_v}, PTUS-F outperforms all other approaches in terms of F-measure and AUPR. In some cases, RB performs the same as PTUS-F. From these results, it is observed that combining the use of F-measure and integration of progressive partitioning, is more effective in increasing performance and robustness compared to using each of them independently.
In the experiments on the video data, trajectory under-sampling is more effective when used in PBoost compared to random under-sampling without replacement and cluster under-sampling. This is the case when partitions of negative class are known a priori.

\begin{table}  [!t]
 \centering
  \caption{Average of F$_2$-measure and AUPR performance of proposed and baseline techniques on video data over different levels of skew in training and test data.}
\label{PrFF_v}
  \resizebox{0.78\textwidth}{!}{
\begin{tabular}{l c|| c|c|c|c|| c|c|c|c} \hline \hline
\multirow{2}[2]{*}{\textbf{Ensembles }}&\makecell{\textbf{Train} \\ \textbf{Data}}& \multicolumn{4}{c||}{$\boldsymbol{\lambda_{\rm train}}=$1:50 } & \multicolumn{4}{c}{$\boldsymbol{\lambda_{\rm train}}=$1:100} \\ \cline{2-10}
&$\boldsymbol{\lambda_{\rm test}}$& \textbf{1:1} & \textbf{1:20} & \textbf{1:50} & \textbf{1:100} & \textbf{1:1} & \textbf{1:20} & \textbf{1:50} & \textbf{1:100}   \\ \hline \hline
\multicolumn{10}{c}{\textbf{F$_2$-measure}}\\\hline \hline
Ada && \makecell{0.80 \\ {\small $\pm$ 0.01}} & \makecell{0.86 \\ {\small $\pm$ 0.01}} & \makecell{0.78 \\ {\small $\pm$ 0.01}} & \makecell{0.85 \\ {\small $\pm$ 0.01}} & \makecell{0.75 \\ {\small $\pm$ 0.01}} & \makecell{0.83 \\ {\small $\pm$ 0.00}} & \makecell{0.72 \\ {\small $\pm$ 0.01}} & \makecell{0.82 \\ {\small $\pm$ 0.00}}\\ \hline
SMT && \makecell{0.93 \\ {\small $\pm$ 0.00}} & \makecell{0.96 \\ {\small $\pm$ 0.00}} & \makecell{0.92 \\ {\small $\pm$ 0.00}} & \makecell{0.96 \\ {\small $\pm$ 0.00}} & \makecell{0.91 \\ {\small $\pm$ 0.00}} & \makecell{0.95 \\ {\small $\pm$ 0.00}} & \makecell{0.89 \\ {\small $\pm$ 0.00}} & \makecell{0.94 \\ {\small $\pm$ 0.00}}\\ \hline
RUS && \makecell{0.94 \\ {\small $\pm$ 0.00}} & \makecell{0.97 \\ {\small $\pm$ 0.00}} & \makecell{0.93 \\ {\small $\pm$ 0.00}} & \makecell{0.96 \\ {\small $\pm$ 0.00}} & \makecell{0.88 \\ {\small $\pm$ 0.00}} & \makecell{0.94 \\ {\small $\pm$ 0.00}} & \makecell{0.87 \\ {\small $\pm$ 0.00}} & \makecell{0.93 \\ {\small $\pm$ 0.00}}\\ \hline
RB && \makecell{0.95 \\ {\small $\pm$ 0.00}} & \makecell{\textbf{0.99} \\ {\small $\pm$ 0.00}} & \makecell{0.94 \\ {\small $\pm$ 0.00}} & \makecell{0.98 \\ {\small $\pm$ 0.00}} & \makecell{0.91 \\ {\small $\pm$ 0.00}} & \makecell{0.97 \\ {\small $\pm$ 0.00}} & \makecell{0.89 \\ {\small $\pm$ 0.00}} & \makecell{\textbf{0.96} \\ {\small $\pm$ 0.00}}\\ \hline
PRUS-F && \makecell{0.95 \\ {\small $\pm$ 0.00}} & \makecell{\textbf{0.99} \\ {\small $\pm$ 0.00}} & \makecell{0.94 \\ {\small $\pm$ 0.00}} & \makecell{0.98 \\ {\small $\pm$ 0.00}} & \makecell{0.90 \\ {\small $\pm$ 0.00}} & \makecell{0.97 \\ {\small $\pm$ 0.00}} & \makecell{0.90 \\ {\small $\pm$ 0.00}} & \makecell{0.94 \\ {\small $\pm$ 0.00}}\\ \hline
PCUS-F && \makecell{0.94 \\ {\small $\pm$ 0.00}} & \makecell{0.98 \\ {\small $\pm$ 0.00}} & \makecell{0.93 \\ {\small $\pm$ 0.00}} & \makecell{0.96 \\ {\small $\pm$ 0.00}} & \makecell{0.90 \\ {\small $\pm$ 0.00}} & \makecell{0.96 \\ {\small $\pm$ 0.00}} & \makecell{0.89 \\ {\small $\pm$ 0.00}} & \makecell{0.93 \\ {\small $\pm$ 0.00}}\\ \hline
PTUS-F && \makecell{\textbf{0.96} \\ {\small $\pm$ 0.00}} & \makecell{\textbf{0.99} \\ {\small $\pm$ 0.00}} & \makecell{\textbf{0.95} \\ {\small $\pm$ 0.00}} & \makecell{\textbf{0.99} \\ {\small $\pm$ 0.00}} & \makecell{\textbf{0.92} \\ {\small $\pm$ 0.00}} & \makecell{\textbf{0.98} \\ {\small $\pm$ 0.00}} & \makecell{\textbf{0.91} \\ {\small $\pm$ 0.00}} & \makecell{\textbf{0.96} \\ {\small $\pm$ 0.00}}\\ \hline
 \hline
%\end{tabular}}
%\end{table}
%
%
%\begin{table} [!htb]
% \centering
%  \caption{Average of AUPR performance of proposed and baseline techniques on video data over different levels of skew in training and test data.}
%\label{PrFAU_v}
%  \resizebox{0.78\textwidth}{!}{
%\begin{tabular}{l c|| c|c|c|c|| c|c|c|c} \hline \hline
%\multirow{2}[2]{*}{\textbf{Ensembles }}&\makecell{\textbf{Train} \\ \textbf{Data}}& \multicolumn{4}{c||}{$\boldsymbol{\lambda_{\rm train}}=$1:50 } & \multicolumn{4}{c}{$\boldsymbol{\lambda_{\rm train}}=$1:100} \\ \cline{2-10}
%&$\boldsymbol{\lambda_{\rm test}}$& \textbf{1:1} & \textbf{1:20} & \textbf{1:50} & \textbf{1:100} & \textbf{1:1} & \textbf{1:20} & \textbf{1:50} & \textbf{1:100}   \\ \hline \hline
\multicolumn{10}{c}{\textbf{AUPR}}\\\hline \hline
Ada && \makecell{0.96 \\ {\small $\pm$ 0.00}} & \makecell{0.97 \\ {\small $\pm$ 0.00}} & \makecell{0.94 \\ {\small $\pm$ 0.00}} & \makecell{0.97 \\ {\small $\pm$ 0.00}} & \makecell{0.91 \\ {\small $\pm$ 0.00}} & \makecell{0.97 \\ {\small $\pm$ 0.00}} & \makecell{0.89 \\ {\small $\pm$ 0.00}} & \makecell{0.96 \\ {\small $\pm$ 0.00}}\\ \hline
SMT && \makecell{0.96 \\ {\small $\pm$ 0.00}} & \makecell{0.97 \\ {\small $\pm$ 0.00}} & \makecell{0.95 \\ {\small $\pm$ 0.00}} & \makecell{0.97 \\ {\small $\pm$ 0.00}} & \makecell{0.94 \\ {\small $\pm$ 0.00}} & \makecell{0.97 \\ {\small $\pm$ 0.00}} & \makecell{0.94 \\ {\small $\pm$ 0.00}} & \makecell{0.97 \\ {\small $\pm$ 0.00}}\\ \hline
RUS && \makecell{1.00 \\ {\small $\pm$ 0.00}} & \makecell{0.99 \\ {\small $\pm$ 0.00}} & \makecell{0.94 \\ {\small $\pm$ 0.00}} & \makecell{0.97 \\ {\small $\pm$ 0.00}} & \makecell{0.84 \\ {\small $\pm$ 0.00}} & \makecell{0.94 \\ {\small $\pm$ 0.00}} & \makecell{0.80 \\ {\small $\pm$ 0.00}} & \makecell{0.92 \\ {\small $\pm$ 0.00}}\\ \hline
RB && \makecell{\textbf{1.00} \\ {\small $\pm$ 0.00}} & \makecell{\textbf{1.00} \\ {\small $\pm$ 0.00}} & \makecell{\textbf{0.98} \\ {\small $\pm$ 0.00}} & \makecell{\textbf{1.00} \\ {\small $\pm$ 0.00}} & \makecell{0.96 \\ {\small $\pm$ 0.00}} & \makecell{\textbf{0.99} \\ {\small $\pm$ 0.00}} & \makecell{0.96 \\ {\small $\pm$ 0.00}} & \makecell{\textbf{0.99} \\ {\small $\pm$ 0.00}}\\ \hline
PRUS-F && \makecell{1.00 \\ {\small $\pm$ 0.00}} & \makecell{0.98 \\ {\small $\pm$ 0.00}} & \makecell{0.96 \\ {\small $\pm$ 0.00}} & \makecell{0.98 \\ {\small $\pm$ 0.00}} & \makecell{\textbf{0.98} \\ {\small $\pm$ 0.00}} & \makecell{0.97 \\ {\small $\pm$ 0.00}} & \makecell{0.94 \\ {\small $\pm$ 0.00}} & \makecell{0.95 \\ {\small $\pm$ 0.00}}\\ \hline
PCUS-F && \makecell{1.00 \\ {\small $\pm$ 0.00}} & \makecell{0.96 \\ {\small $\pm$ 0.01}} & \makecell{0.95 \\ {\small $\pm$ 0.00}} & \makecell{0.97 \\ {\small $\pm$ 0.02}} & \makecell{0.89 \\ {\small $\pm$ 0.01}} & \makecell{0.94 \\ {\small $\pm$ 0.02}} & \makecell{0.85 \\ {\small $\pm$ 0.01}} & \makecell{0.93 \\ {\small $\pm$ 0.02}}\\ \hline
PTUS-F && \makecell{\textbf{1.00} \\ {\small $\pm$ 0.00}} & \makecell{\textbf{1.00} \\ {\small $\pm$ 0.00}} & \makecell{\textbf{0.98} \\ {\small $\pm$ 0.00}} & \makecell{\textbf{1.00} \\ {\small $\pm$ 0.00}} & \makecell{\textbf{0.98} \\ {\small $\pm$ 0.00}} & \makecell{\textbf{0.99} \\ {\small $\pm$ 0.00}} & \makecell{\textbf{0.97} \\ {\small $\pm$ 0.00}} & \makecell{\textbf{0.99 }\\ {\small $\pm$ 0.00}}\\ \hline
\hline 
\end{tabular}}
\end{table}

\subsection{Results of Experiments with KEEL Collection}
In Table \ref{Keel}, the performance of baseline and proposed Boosting ensembles is compared in terms of F$_2$-measure and AUPR for experiments with 21 KEEL datasets. The second column of the table shows the imbalance level of training and test data in each dataset that ranges between 1:9  and 1:29. In this table, for each dataset, the best values are bold and second-best values are italic-bold to show the first and second best classifiers, respectively. In terms of F-measure, PCUS has one of the two highest values for 17 datasets, RB has one of the two highest values for 15 datasets and PRUS has one of the two highest values for 13 datasets. RB is the best classifier for 14 datasets while PCUS and PRUS are the best for 5 and 7 datasets, respectively.
In terms of AUPR, PCUS and RB have one of the two highest values for 17 datasets,  and PRUS has one of the two highest values for 8 datasets. RB is the best classifier for 15 datasets while PCUS and PRUS are the best for 5 datasets.
In the cases when no natural data partitioning is known a priori, clustering with k-means is more effective than random under-sampling because base classifiers are trained on different parts of feature space and therefore the ensemble have more diversity compared to the case when the base classifiers are trained on samples from all parts of feature space. More sophisticated clustering methods like kernel k-means and spectral clustering may be more suitable.

\begin{table} [!b]
 \centering
  \caption{Average of F$_2$-measure and AUPR performance of proposed and baseline techniques on 21 real-world datasets from Keel collection.}
\label{Keel}
  \resizebox{\textwidth}{!}{
\begin{tabular}{l l || c c c c c c || c c c c c c } \hline \hline
&& \multicolumn{6}{c||}{\textbf{F$_2$-measure}} & \multicolumn{6}{c}{\textbf{AUPR}}\\   
\textbf{Data} &  $\boldsymbol{\lambda_{\rm train}}$, $\boldsymbol{\lambda_{\rm test}}$& \textbf{Ada} & \textbf{SMT} & \textbf{RUS} & \textbf{RB} & \textbf{PRUS-F} & \textbf{PCUS-F} & \textbf{Ada} & \textbf{SMT} & \textbf{RUS} & \textbf{RB} & \textbf{PRUS-F} & \textbf{PCUS-F} \\  \hline \hline
abalone19                  & 129.44    &                                  -                                & \makecell{0.16 \\ {\small $\pm$ 0.00}} & \makecell{0.06 \\ {\small $\pm$ 0.00}} & \makecell{\textbf{\textit{0.24}} \\ {\small $\pm$ 0.00}} & \makecell{\textbf{0.31} \\ {\small $\pm$ 0.01}} & \makecell{0.14 \\ {\small $\pm$ 0.01}} &                                 -                                & \makecell{\textbf{0.45} \\ {\small $\pm$ 0.00}} & \makecell{\textbf{\textit{0.43}} \\ {\small $\pm$ 0.00}} & \makecell{0.21 \\ {\small $\pm$ 0.00}} & \makecell{0.21 \\ {\small $\pm$ 0.00}} & \makecell{0.25 \\ {\small $\pm$ 0.01}}\\ \hline
abalone9-18               & 16.40     & \makecell{0.19 \\ {\small $\pm$ 0.01}} & \makecell{\textbf{\textit{0.63} }\\ {\small $\pm$ 0.01}} & \makecell{0.35 \\ {\small $\pm$ 0.01}} & \makecell{\textbf{0.67} \\ {\small $\pm$ 0.01}} & \makecell{0.62 \\ {\small $\pm$ 0.01}} & \makecell{0.49 \\ {\small $\pm$ 0.02}} & \makecell{0.56 \\ {\small $\pm$ 0.01}} & \makecell{0.62 \\ {\small $\pm$ 0.01}} & \makecell{0.45 \\ {\small $\pm$ 0.01}} & \makecell{\textbf{0.68} \\ {\small $\pm$ 0.02}} & \makecell{\textbf{\textit{0.65}} \\ {\small $\pm$ 0.01}} & \makecell{0.53 \\ {\small $\pm$ 0.01}}\\ \hline
ecoli-0-1-3-7-vs-2-6   & 39.14   & \makecell{0.23 \\ {\small $\pm$ 0.03}} & \makecell{0.32 \\ {\small $\pm$ 0.04}} &  \makecell{0.28 \\ {\small $\pm$ 0.00}}& \makecell{\textbf{0.93} \\ {\small $\pm$ 0.02}} & \makecell{\textbf{\textit{0.85}} \\ {\small $\pm$ 0.00}} & \makecell{0.82 \\ {\small $\pm$ 0.04}} & \makecell{0.70 \\ {\small $\pm$ 0.02}} & \makecell{\textbf{\textit{0.95}} \\ {\small $\pm$ 0.01}} &  \makecell{0.75 \\ {\small $\pm$ 0.00}} & \makecell{\textbf{1.00} \\ {\small $\pm$ 0.00}} & \makecell{0.85 \\ {\small $\pm$ 0.00}} & \makecell{\textbf{\textit{0.95}} \\ {\small $\pm$ 0.01}}\\ \hline
ecoli4                         & 15.80     & \makecell{0.76 \\ {\small $\pm$ 0.01}} & \makecell{0.91 \\ {\small $\pm$ 0.00}} & \makecell{0.76 \\ {\small $\pm$ 0.00}} & \makecell{\textbf{0.95} \\ {\small $\pm$ 0.01}} & \makecell{\textbf{0.95} \\ {\small $\pm$ 0.00}} & \makecell{\textbf{0.95} \\ {\small $\pm$ 0.00}} & \makecell{0.94 \\ {\small $\pm$ 0.00}} & \makecell{0.90 \\ {\small $\pm$ 0.01}} & \makecell{0.80 \\ {\small $\pm$ 0.01}} & \makecell{\textbf{0.98} \\ {\small $\pm$ 0.00}} & \makecell{0.95 \\ {\small $\pm$ 0.00}} & \makecell{\textbf{\textit{0.97}} \\ {\small $\pm$ 0.00}}\\ \hline
glass-0-1-6-vs-2        & 10.29    & \makecell{0.18 \\ {\small $\pm$ 0.01}} & \makecell{0.79 \\ {\small $\pm$ 0.01}} & \makecell{0.60 \\ {\small $\pm$ 0.01}} & \makecell{0.89 \\ {\small $\pm$ 0.00}} & \makecell{\textbf{\textit{0.90}} \\ {\small $\pm$ 0.01}} & \makecell{\textbf{0.93} \\ {\small $\pm$ 0.00}} & \makecell{0.59 \\ {\small $\pm$ 0.01}} & \makecell{0.78 \\ {\small $\pm$ 0.01}} & \makecell{0.67 \\ {\small $\pm$ 0.00}} & \makecell{0.86 \\ {\small $\pm$ 0.00}} & \makecell{\textbf{0.96} \\ {\small $\pm$ 0.00}} & \makecell{\textbf{\textit{0.92}} \\ {\small $\pm$ 0.01}}\\ \hline
glass-0-1-6-vs-5      & 19.44    & \makecell{0.30 \\ {\small $\pm$ 0.04}} & \makecell{0.97 \\ {\small $\pm$ 0.01}} & \makecell{0.17 \\ {\small $\pm$ 0.02}} & \makecell{\textbf{1.00} \\ {\small $\pm$ 0.00}} & \makecell{\textbf{\textit{0.96}} \\ {\small $\pm$ 0.00}} & \makecell{\textbf{\textit{0.96}} \\ {\small $\pm$ 0.00}} & \makecell{0.86 \\ {\small $\pm$ 0.01}} & \makecell{\textbf{1.00} \\ {\small $\pm$ 0.00}} & \makecell{0.54 \\ {\small $\pm$ 0.00}} & \makecell{\textbf{1.00} \\ {\small $\pm$ 0.00}} & \makecell{0.93 \\ {\small $\pm$ 0.00}} & \makecell{\textbf{1.00} \\ {\small $\pm$ 0.00}}\\ \hline
glass2                       & 11.59    & \makecell{0.12 \\ {\small $\pm$ 0.01}} & \makecell{0.75 \\ {\small $\pm$ 0.01}} & \makecell{0.58 \\ {\small $\pm$ 0.00}} & \makecell{0.83 \\ {\small $\pm$ 0.00}} & \makecell{\textbf{0.91} \\ {\small $\pm$ 0.00}} & \makecell{\textbf{\textit{0.89}} \\ {\small $\pm$ 0.01}} & \makecell{0.62 \\ {\small $\pm$ 0.01}} & \makecell{0.72 \\ {\small $\pm$ 0.01}} & \makecell{0.67 \\ {\small $\pm$ 0.00}} & \makecell{0.75 \\ {\small $\pm$ 0.01}} & \makecell{\textbf{0.92} \\ {\small $\pm$ 0.01}} & \makecell{\textbf{\textit{0.88}} \\ {\small $\pm$ 0.01}}\\ \hline
glass4                        & 15.46    & \makecell{0.42 \\ {\small $\pm$ 0.03}} & \makecell{0.87 \\ {\small $\pm$ 0.01}} & \makecell{0.21 \\ {\small $\pm$ 0.02}} & \makecell{\textbf{1.00} \\ {\small $\pm$ 0.00}} & \makecell{\textbf{\textit{0.97}} \\ {\small $\pm$ 0.00}} & \makecell{\textbf{\textit{0.97}} \\ {\small $\pm$ 0.00}} & \makecell{0.88 \\ {\small $\pm$ 0.01}} & \makecell{0.97 \\ {\small $\pm$ 0.00}} & \makecell{0.57 \\ {\small $\pm$ 0.01}} & \makecell{\textbf{1.00} \\ {\small $\pm$ 0.00}} & \makecell{\textbf{\textit{0.99}} \\ {\small $\pm$ 0.00}} & \makecell{\textbf{1.00} \\ {\small $\pm$ 0.00}}\\ \hline
glass5                        & 22.78    & \makecell{0.39 \\ {\small $\pm$ 0.03}} & \makecell{0.96 \\ {\small $\pm$ 0.01}} & \makecell{0.08 \\ {\small $\pm$ 0.01}} & \makecell{\textbf{1.00} \\ {\small $\pm$ 0.00}} & \makecell{0.96 \\ {\small $\pm$ 0.00}} & \makecell{\textbf{\textit{0.97}} \\ {\small $\pm$ 0.00}} & \makecell{0.86 \\ {\small $\pm$ 0.01}} & \makecell{0.99 \\ {\small $\pm$ 0.00}} & \makecell{0.51 \\ {\small $\pm$ 0.00}} & \makecell{\textbf{1.00} \\ {\small $\pm$ 0.00}} & \makecell{0.98 \\ {\small $\pm$ 0.00}} & \makecell{\textbf{1.00} \\ {\small $\pm$ 0.00}}\\ \hline
page-blocks-1-3-vs-4 & 15.86   & \makecell{0.16 \\ {\small $\pm$ 0.01}} & \makecell{0.22 \\ {\small $\pm$ 0.01}} &  \makecell{0.23 \\ {\small $\pm$ 0.00}} & \makecell{\textbf{0.76} \\ {\small $\pm$ 0.01}} & \makecell{0.54 \\ {\small $\pm$ 0.00}} & \makecell{\textbf{\textit{0.58}} \\ {\small $\pm$ 0.03}} & \makecell{0.81 \\ {\small $\pm$ 0.00}} & \makecell{0.81 \\ {\small $\pm$ 0.01}} &\makecell{0.52 \\ {\small $\pm$ 0.00}}   & \makecell{\textbf{1.00} \\ {\small $\pm$ 0.00}} & \makecell{0.95 \\ {\small $\pm$ 0.00}} & \makecell{\textbf{0\textit{.98}} \\ {\small $\pm$ 0.01}}\\ \hline
shuttle-c0-vs-c4           & 13.87    & \makecell{0.12 \\ {\small $\pm$ 0.00}} & \makecell{\textbf{\textit{0.71}} \\ {\small $\pm$ 0.02}} &  \makecell{0.24 \\ {\small $\pm$ 0.00}} & \makecell{\textbf{0.95} \\ {\small $\pm$ 0.00}} & \makecell{0.57 \\ {\small $\pm$ 0.00}} & \makecell{\textbf{\textit{0.71}} \\ {\small $\pm$ 0.03}} & \makecell{0.82 \\ {\small $\pm$ 0.00}} & \makecell{\textbf{\textit{0.95}} \\ {\small $\pm$ 0.00}} & \makecell{0.85 \\ {\small $\pm$ 0.00}} & \makecell{\textbf{1.00} \\ {\small $\pm$ 0.00}} & \makecell{0.87 \\ {\small $\pm$ 0.02}} & \makecell{0.87 \\ {\small $\pm$ 0.02}}\\ \hline
shuttle-c2-vs-c4            & 20.50    & \makecell{0.32 \\ {\small $\pm$ 0.04}} & \makecell{0.09 \\ {\small $\pm$ 0.01}} & \makecell{0.18 \\ {\small $\pm$ 0.00}}& \makecell{\textbf{1.00} \\ {\small $\pm$ 0.00}} & \makecell{0.80 \\ {\small $\pm$ 0.00}} & \makecell{\textbf{\textit{0.87}} \\ {\small $\pm$ 0.02}} & \makecell{0.77 \\ {\small $\pm$ 0.01}} & \makecell{\textbf{\textit{0.81}} \\ {\small $\pm$ 0.02}} &\makecell{0.55 \\ {\small $\pm$ 0.00}}   & \makecell{\textbf{1.00} \\ {\small $\pm$ 0.00}} & \makecell{0.72 \\ {\small $\pm$ 0.00}} & \makecell{0.78 \\ {\small $\pm$ 0.01}}\\ \hline
vowel0                         & 9.98     & \makecell{0.65 \\ {\small $\pm$ 0.01}} & \makecell{0.98 \\ {\small $\pm$ 0.00}} & \makecell{\textbf{1.00} \\ {\small $\pm$ 0.00}} & \makecell{\textbf{1.00} \\ {\small $\pm$ 0.00}} & \makecell{\textbf{1.00} \\ {\small $\pm$ 0.00}} & \makecell{\textbf{1.00} \\ {\small $\pm$ 0.00}} & \makecell{0.94 \\ {\small $\pm$ 0.00}} & \makecell{\textbf{1.00} \\ {\small $\pm$ 0.00}} & \makecell{\textbf{1.00} \\ {\small $\pm$ 0.00}} & \makecell{\textbf{1.00} \\ {\small $\pm$ 0.00}} & \makecell{\textbf{1.00} \\ {\small $\pm$ 0.00}} & \makecell{\textbf{1.00} \\ {\small $\pm$ 0.00}}\\ \hline
yeast-0-5-6-7-9-vs-4   & 9.35    & \makecell{0.41 \\ {\small $\pm$ 0.02}} & \makecell{\textbf{\textit{0.67}} \\ {\small $\pm$ 0.00}} & \makecell{0.60 \\ {\small $\pm$ 0.00}} & \makecell{0.66 \\ {\small $\pm$ 0.00}} & \makecell{\textbf{0.69 }\\ {\small $\pm$ 0.00}} & \makecell{\textbf{\textit{0.67}} \\ {\small $\pm$ 0.01}} & \makecell{0.64 \\ {\small $\pm$ 0.01}} & \makecell{0.64 \\ {\small $\pm$ 0.00}} & \makecell{0.58 \\ {\small $\pm$ 0.00}} & \makecell{\textbf{0.72} \\ {\small $\pm$ 0.01}} & \makecell{0.62 \\ {\small $\pm$ 0.00}} & \makecell{\textbf{\textit{0.67}} \\ {\small $\pm$ 0.01}}\\ \hline
yeast-1-2-8-9-vs-7     & 30.57     & \makecell{0.06 \\ {\small $\pm$ 0.01}} & \makecell{0.38 \\ {\small $\pm$ 0.01}} & \makecell{0.21 \\ {\small $\pm$ 0.00}} & \makecell{0.39 \\ {\small $\pm$ 0.00}} & \makecell{\textbf{\textit{0.45}} \\ {\small $\pm$ 0.00}} & \makecell{\textbf{0.46} \\ {\small $\pm$ 0.01}} & \makecell{\textbf{0.54} \\ {\small $\pm$ 0.01}} & \makecell{\textbf{\textit{0.49}} \\ {\small $\pm$ 0.01}} & \makecell{0.39 \\ {\small $\pm$ 0.01}} & \makecell{0.46 \\ {\small $\pm$ 0.01}} & \makecell{0.42 \\ {\small $\pm$ 0.01}} & \makecell{0.46 \\ {\small $\pm$ 0.01}}\\ \hline
yeast-1-4-5-8-vs-7     & 22.10    & \makecell{0.02 \\ {\small $\pm$ 0.00}} & \makecell{\textbf{\textit{0.40}} \\ {\small $\pm$ 0.00}} & \makecell{0.24 \\ {\small $\pm$ 0.01}} & \makecell{\textbf{\textit{0.40}} \\ {\small $\pm$ 0.00}} & \makecell{\textbf{\textit{0.40}} \\ {\small $\pm$ 0.00}} & \makecell{\textbf{0.46} \\ {\small $\pm$ 0.01}} & \makecell{\textbf{0.53} \\ {\small $\pm$ 0.00}} & \makecell{\textbf{0.47} \\ {\small $\pm$ 0.00}} & \makecell{0.34 \\ {\small $\pm$ 0.02}} & \makecell{0.36 \\ {\small $\pm$ 0.00}} & \makecell{0.37 \\ {\small $\pm$ 0.00}} & \makecell{\textbf{\textit{0.47}} \\ {\small $\pm$ 0.01}}\\ \hline
yeast-1-vs-7             & 14.30      & \makecell{0.24 \\ {\small $\pm$ 0.02}} & \makecell{\textbf{\textit{0.57}} \\ {\small $\pm$ 0.00}} & \makecell{0.37 \\ {\small $\pm$ 0.01}} & \makecell{0.56 \\ {\small $\pm$ 0.00}} & \makecell{\textbf{0.61} \\ {\small $\pm$ 0.01}} & \makecell{\textbf{\textit{0.57}} \\ {\small $\pm$ 0.01}} & \makecell{0.58 \\ {\small $\pm$ 0.01}} & \makecell{0.54 \\ {\small $\pm$ 0.00}} & \makecell{0.44 \\ {\small $\pm$ 0.02}} & \makecell{\textbf{\textit{0.57}} \\ {\small $\pm$ 0.01}} & \makecell{\textbf{0.59} \\ {\small $\pm$ 0.01}} & \makecell{\textbf{\textit{0.57}} \\ {\small $\pm$ 0.01}}\\ \hline
yeast-2-vs-4               & 9.08    & \makecell{0.67 \\ {\small $\pm$ 0.02}} & \makecell{\textbf{\textit{0.85}} \\ {\small $\pm$ 0.00}} & \makecell{0.73 \\ {\small $\pm$ 0.00}} & \makecell{\textbf{0.86} \\ {\small $\pm$ 0.00}} & \makecell{0.84 \\ {\small $\pm$ 0.00}} & \makecell{0.78 \\ {\small $\pm$ 0.01}} & \makecell{0.84 \\ {\small $\pm$ 0.01}} & \makecell{0.82 \\ {\small $\pm$ 0.01}} & \makecell{0.73 \\ {\small $\pm$ 0.00}} & \makecell{\textbf{0.91} \\ {\small $\pm$ 0.01}} & \makecell{0.87 \\ {\small $\pm$ 0.00}} & \makecell{\textbf{\textit{0.88}} \\ {\small $\pm$ 0.01}}\\ \hline
yeast-2-vs-8                & 23.10   & \makecell{0.49 \\ {\small $\pm$ 0.02}} & \makecell{0.50 \\ {\small $\pm$ 0.00}} & \makecell{0.43 \\ {\small $\pm$ 0.00}} & \makecell{0.63 \\ {\small $\pm$ 0.01}} & \makecell{\textbf{0.71} \\ {\small $\pm$ 0.00}} & \makecell{\textbf{\textit{0.69}} \\ {\small $\pm$ 0.00}} & \makecell{0.72 \\ {\small $\pm$ 0.01}} & \makecell{0.54 \\ {\small $\pm$ 0.00}} & \makecell{0.53 \\ {\small $\pm$ 0.00}} & \makecell{\textbf{0.76} \\ {\small $\pm$ 0.00}} & \makecell{\textbf{\textit{0.75}} \\ {\small $\pm$ 0.00}} & \makecell{\textbf{\textit{0.75}} \\ {\small $\pm$ 0.00}}\\ \hline
yeast4                         & 28.10    & \makecell{0.10 \\ {\small $\pm$ 0.00}} & \makecell{0.56 \\ {\small $\pm$ 0.00}} & \makecell{0.38 \\ {\small $\pm$ 0.00}} & \makecell{\textbf{0.60} \\ {\small $\pm$ 0.00}} & \makecell{\textbf{\textit{0.58}} \\ {\small $\pm$ 0.00}} & \makecell{0.57 \\ {\small $\pm$ 0.00}} & \makecell{0.44 \\ {\small $\pm$ 0.01}} & \makecell{\textbf{0.57} \\ {\small $\pm$ 0.00}} & \makecell{0.52 \\ {\small $\pm$ 0.00}} & \makecell{\textbf{\textit{0.55}} \\ {\small $\pm$ 0.00}} & \makecell{0.52 \\ {\small $\pm$ 0.00}} & \makecell{\textbf{\textit{0.55}} \\ {\small $\pm$ 0.01}}\\ \hline
yeast5                           & 32.73    & \makecell{0.42 \\ {\small $\pm$ 0.02}} & \makecell{0.81 \\ {\small $\pm$ 0.00}} & \makecell{0.64 \\ {\small $\pm$ 0.00}} & \makecell{\textbf{0.86} \\ {\small $\pm$ 0.00}} & \makecell{\textbf{\textit{0.83}} \\ {\small $\pm$ 0.00}} & \makecell{\textbf{\textit{0.83}} \\ {\small $\pm$ 0.00}} & \makecell{0.63 \\ {\small $\pm$ 0.00}} & \makecell{0.79 \\ {\small $\pm$ 0.00}} & \makecell{0.68 \\ {\small $\pm$ 0.00}} & \makecell{\textbf{0.80} \\ {\small $\pm$ 0.00}} & \makecell{\textbf{0.80} \\ {\small $\pm$ 0.00}} & \makecell{\textbf{0.80} \\ {\small $\pm$ 0.00}}\\ \hline
yeast6                           & 41.40    & \makecell{0.29 \\ {\small $\pm$ 0.02}} & \makecell{0.59 \\ {\small $\pm$ 0.01}} & \makecell{0.38 \\ {\small $\pm$ 0.01}} & \makecell{\textbf{0.67} \\ {\small $\pm$ 0.01}} & \makecell{0.64 \\ {\small $\pm$ 0.01}} & \makecell{\textbf{\textit{0.66}} \\ {\small $\pm$ 0.01}} & \makecell{0.61 \\ {\small $\pm$ 0.01}} & \makecell{0.63 \\ {\small $\pm$ 0.01}} & \makecell{0.54 \\ {\small $\pm$ 0.01}} & \makecell{\textbf{0.70} \\ {\small $\pm$ 0.01}} & \makecell{0.68 \\ {\small $\pm$ 0.01}} & \makecell{\textbf{\textit{0.69}} \\ {\small $\pm$ 0.01}}\\ \hline
\hline 
\end{tabular}}
\end{table}

\subsection{Computational Complexity}
In this section the time complexity needed to design and test the proposed and baseline Boosting ensembles are compared. To compare the training time and memory cost of these ensembles, the number of training samples is counted and to compare their validation time and memory cost the number of validation samples and the number of support vectors of base classifiers are considered. 

Figure~\ref{cmpx1} show the results obtained with data set $D_2$ in our experiments. The number of training and validation samples, the average number of support vectors, and overall number of evaluations of the kernel function ($n_{\rm SV} \cdot n_{\rm val}$) is presented in Figure~\ref{cmpx1}(a)-(d) to estimate and compare design time of the proposed and baseline Boosting ensembles. To compare the complexity of these classification systems during testing $O(n_{\rm SV})$ with a probe sample $\textbf{x}$, we compared the overall number of support vectors in these ensembles in Figure~\ref{cmpx1}(e) because computing each SVM output requires $n_{\rm SV}$ evaluations of the kernel function.  

Given $n_{\rm tr}^e$ as the number of samples to train the $e^{\rm th}$ classifier in the ensemble, Figure~\ref{cmpx1}(a) shows $\sum_{\rm e=1}^{\rm E}n_{\rm tr}^e$. In Figure~\ref{cmpx1}(b), $\sum_{\rm e=1}^{\rm E}n_{\rm val}^e$ is presented, where $n_{\rm val}^e$ is the number of samples that the $e^{\rm th}$ classifier in the ensemble is validated with. Average number of support vectors in $E$ classifiers of the ensembles are shown in Figure~\ref{cmpx1}(c). Given $n_{\rm SV}^e$ as the number of support vectors obtained after training the $e^{\rm th}$ classifier in the ensemble, Figure~\ref{cmpx1}(d) shows $\sum_{\rm e=1}^{\rm E}n_{\rm val}^e \cdot n_{\rm SV}^e$ for each ensemble.

\begin{figure}[!htb]
\begin{center}
        \begin{subfigure}[b]{0.3\textwidth}
            \includegraphics[width=1\linewidth]{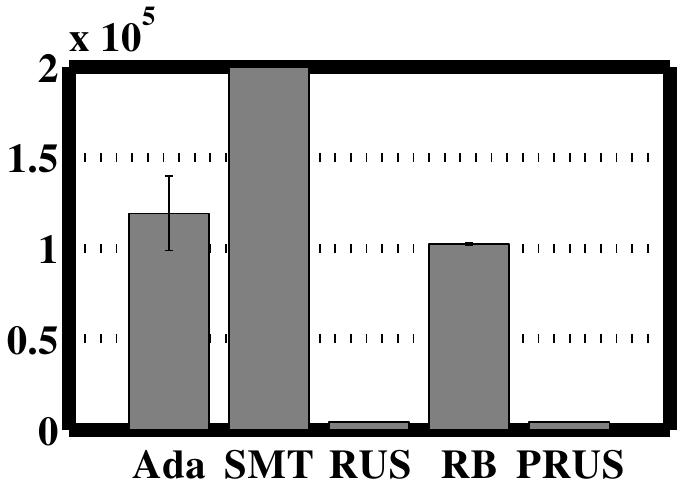} 
         \caption{}
         \end{subfigure} 
        \begin{subfigure}[b]{0.3\textwidth}
        \includegraphics[width=1\linewidth]{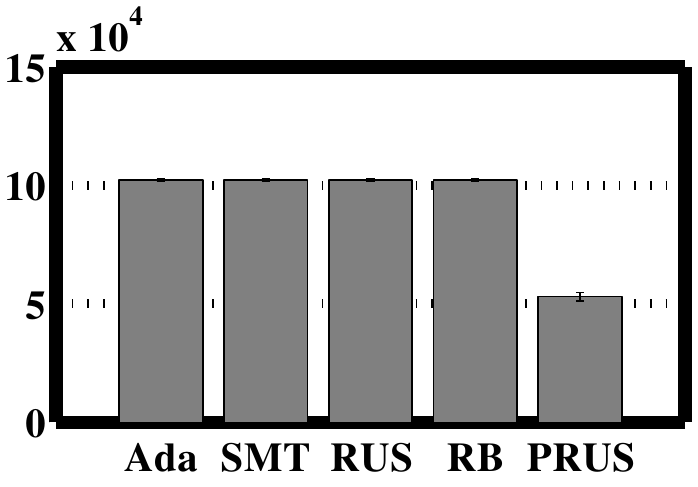} 
        \caption{}
         \end{subfigure}
         \begin{subfigure}[b]{0.3\textwidth}
        \includegraphics[width=1\linewidth]{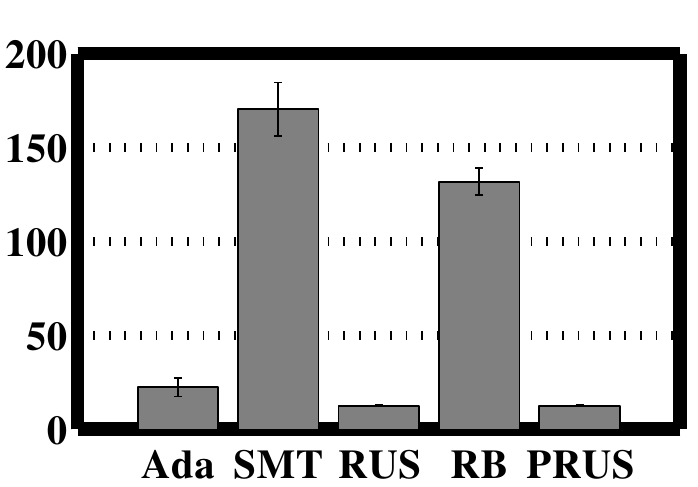} 
        \caption{}
         \end{subfigure}
         \begin{subfigure}[b]{0.3\textwidth}
        \includegraphics[width=1\linewidth]{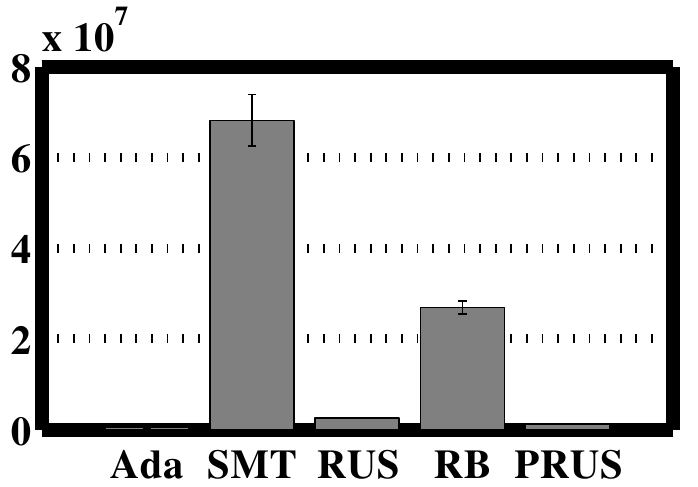} 
        \caption{}
         \end{subfigure}
         \hspace{+0.5cm}
         \begin{subfigure}[b]{0.3\textwidth}
        \includegraphics[width=1\linewidth]{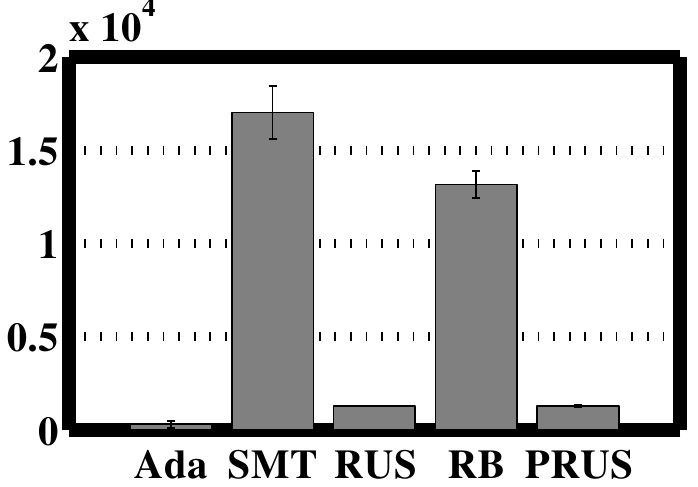}
        \caption{}
         \end{subfigure}
\caption{Complexity related to the design and testing process of ensembles: (a) total number of training samples, (b) total number of validation samples, (c) Number of $n_{\rm SV} \cdot n_{\rm val}$ during validation, (d) total number of $n_{\rm SV} \cdot n_{\rm val}$ during validation, (e) total number of evaluations of the kernel function per probe sample during testing.}
\label{cmpx1}
\label{cmpx1}
\end{center}
\end{figure}

In terms of training (see Figure~\ref{cmpx1}(a)), PRUS and RUS are under-sampling ensembles and have the lowest computational cost, while SMT and RB-Boost include up-sampling and are significantly more costly. Total number of validation samples is equal for Ada, SMT, RUS and RB, and total number of validation samples is less with PRUS (see Figure~\ref{cmpx1}(b)). The average number of support vectors is higher for SMT (see Figure~\ref{cmpx1}(c)) because the base classifiers in this ensemble are trained on higher number of samples.
 Therefore, SMT is the most costly method, in terms of validation (see Figure~\ref{cmpx1}(d)). Note that time and memory required for partitioning in PRUS, and generating synthetic samples in SMT and RB-Boost is neglected here. Nevertheless, PRUS is the most efficient ensemble technique in terms of designing memory and time complexity. 
 
 The number of training and validation samples as well as the average number of support vectors is smaller with PRUS and therefore, PRUS is less costly in terms of design time and memory complexity.

In terms of testing time complexity (see Figure~\ref{cmpx1}(e)) PRUS and RUS have the lowest number of evaluations of the kernel function per probe sample. RB-Boost and SMT have the highest number of evaluations of the kernel function per probe sample. 

 Although Ada is given the same ensemble size, it fails to generate enough classifiers \footnote{AdaBoost failed many times to be generated because the training subset in step 2.i of Algo. \ref{Boosting} may contain only negative class samples during sampling or weight update in step 2.vii may lead to that due to unsuitable loss factor calculation in step 2.ivv. Nevertheless, only successful attempts are considered.} and consequently result in smaller number of support vectors. Therefore, the total number of validation and testing processes of Ada is lower than expected.

%, 
\subsection{Summary of Results}
As a summary of results on synthetic, video and KEEL datasets, we observed that: 
\begin{enumerate}
\item Using the proposed loss factor calculation may reduce the bias of performance in Boosting ensembles and increase the accuracy.
\item Partitioning improves the performance of RUS in all cases in terms of both accuracy and robustness to imbalance.
\item Integrating both partitioning and the proposed loss factor outperforms state of the art Boosting ensembles, relying on the choice of partitioning technique for each dataset such that:
\begin{enumerate}
\item With synthetic data, PCUS$_i$-F outperforms all systems in terms of both F-measure and AUPR, while PRUS and PCUS outperform RUS in most cases of skew and overlap between classes.
\item With the video data, PTUS is more accurate than the state of the art ensembles, and PRUS as well as PCUS.
\item PCUS is one of the best two classifiers in most KEEL problems, and performs very closely to RB in terms of both F-measure and AUPR. 
\end{enumerate}
\item PBoost is computationally less costly than the state of the art Boosting ensembles in terms of computational complexity.
\end{enumerate}
Therefore, PBoost is an effective approach in correct classification of data when data is imbalanced in comparison to the state of the art Boosting ensembles. This method relies on the choice of partitioning technique for each dataset and performs significantly better when a more suitable partitioning technique is used. In problems that the natural clusters are known, %like as conversion of multi-class classification problems to binary classification,
the performance is better than using the general partitioning methods such as random under-sampling without replacement or k-means clustering. 
Therefore, PBoost can be more efficient than baseline Boosting ensembles considering both accuracy and complexity factors.

\section{Conclusion}
In this paper, a new Boosting ensemble algorithm named as PBoost is proposed to address imbalance based on the idea of modifying RUSBoost by (1) under-sampling the majority class using partitional techniques, (2) validating classifiers on a growing validation subset, and (3) using a more suitable loss factor calculation.
 The partitions enter the Boosting process progressively for designing classifiers over iterations to avoid information loss and to maintain diversity among them. Validating base classifiers on a growing number of negative samples makes the PBoost ensembles more robust to possible skew levels of data during operations in addition to lowering the computational complexity.
 The new loss factor defined in this Boosting ensemble handles bias of performance towards negative class, and guides the Boosting process in a more effective direction with the purpose of correctly classifying both classes.  
Experiments show that PBoost may perform differently with different techniques of partitioning for each dataset such that more suitable clustering result in better performance. Nevertheless, regardless of the partitioning method, the PBoost ensembles perform comparably to state of the art Boosting ensembles in terms of accuracy. In addition, PBoost has significantly lower computational complexity in both designing and testing stages compared to state of the art Boosting ensembles. 
The applicability of PBoost to multi-class classification problems and deployment of more sophisticated  clustering methods in PBoost can be further investigated in a future work.

\section*{Acknowledgement}
This work was partially supported by the Natural Sciences and Engineering Research Council of Canada and Mitacs.

\section*{References}
{\small
\bibliographystyle{elsarticle-num}
\bibliography{refs}
}

\end{document}